\documentclass[sn-mathphys-num, iicol]{sn-jnl}

\usepackage{graphicx}%
\usepackage{multirow}%
\usepackage{amsmath,amssymb,amsfonts}%
\usepackage{amsthm}%
\usepackage[title]{appendix}%
\usepackage[dvipsnames]{xcolor}%
\usepackage{textcomp}%
\usepackage{manyfoot}%
\usepackage{booktabs}%
\usepackage{algorithm}%
\usepackage{algorithmicx}%
\usepackage{algpseudocode}%
\usepackage{listings}%

\usepackage{amsfonts}

\definecolor{colorbaseline}{RGB}{235,235,235}
\definecolor{colorpsp}{RGB}{224,238,238}
\definecolor{colorist}{RGB}{225, 240, 213}
\definecolor{colorA2M2}{RGB}{255,214,165}
\definecolor{colorA2M2lt}{RGB}{255,245,205}
\definecolor{colormtfan}{RGB}{242,216,137}
\definecolor{grey}{HTML}{808080}
\definecolor{colorR}{HTML}{3370BD}
\definecolor{colortau}{HTML}{4CAF50}
\definecolor{colortauorg}{HTML}{FF82AB}

\definecolor{colororange}{HTML}{658533}

\definecolor{revisioncolor}{HTML}{3370BD}

\newcommand{\revision}[1]{#1}
\newcommand{\mnrevision}[1]{#1}

\PassOptionsToPackage{pagebackref,breaklinks,colorlinks}{hyperref}

\usepackage{bookmark}

\usepackage{multirow}
\usepackage{subcaption} 
\newcommand{\tablestyle}[2]{\setlength{\tabcolsep}{#1}\renewcommand{\arraystretch}{#2}\centering\footnotesize}

\usepackage{makecell}
\usepackage{tabu}
 \usepackage{amssymb}
 \usepackage{mathtools}  
 \usepackage{float}
 \usepackage{booktabs}
 \usepackage{amsmath}
\usepackage{colortbl}
\usepackage{arydshln} 
\usepackage{lmodern}
\usepackage{microtype}

\usepackage{silence}
\usepackage{hyperref}

\hypersetup{
    colorlinks=true,
    linkcolor=red,
    citecolor=red,
    urlcolor=blue
}

\usepackage{bibunits} 
\defaultbibliographystyle{sn-mathphys-num} 
\defaultbibliography{ref_fsl}              

\begin{document}

\begin{bibunit}

\title[Article Title]{A$^2$M$^2$-Net: Adaptively Aligned Multi-Scale Moment for Few-Shot Action Recognition}


\author[1]{\fnm{Zilin} \sur{Gao}}\email{gzl@mail.dlut.com}\equalcont{Equal contribution.}

\author[2,3]{\fnm{Qilong} \sur{Wang}}\email{qlwang@tju.com}\equalcont{Equal contribution.}

\author[4]{\fnm{Bingbing} \sur{Zhang}}\email{icyzhang@dlnu.edu.cn}

\author[2,3]{\fnm{Qinghua} \sur{Hu}}\email{huqinghua@tju.edu.cn}

\author*[1]{\fnm{Peihua} \sur{Li}}\email{peihuali@dlut.edu.cn}

\affil[1]{\centering{\orgdiv{School of Information and Communication Engineering},} \\
\orgname{Dalian University of Technology}, \city{Dalian}, \country{China}}

\affil[2]{\orgdiv{College of Intelligence and Computing}, \orgname{Tianjin University}, \city{Tianjin}, \country{China}}

\affil[3]{\orgdiv{Haihe Laboratory of Information Technology Application Innovation}, \city{Tianjin}, \country{China}}


\affil[4]{\orgdiv{School of Computer Science and Engineering}, \orgname{Dalian Minzu University}, \city{Dalian}, \country{China}}


\abstract{Thanks to capability to alleviate  the cost of large-scale annotation, few-shot action recognition (FSAR) has attracted increased attention of researchers in recent years. Existing FSAR approaches typically neglect the role of individual motion pattern in comparison, and under-explore the feature statistics for video dynamics. 
\mnrevision{Thereby, they struggle to handle the challenging temporal misalignment in video dynamics, particularly by using 2D backbones.}
To overcome these limitations, this work proposes an adaptively aligned multi-scale second-order moment network, namely A$^2$M$^2$-Net, to describe the latent video dynamics with a collection of powerful representation candidates and adaptively align them in an instance-guided manner. To this end, our A$^2$M$^2$-Net involves two core components, namely, adaptive alignment (A$^2$ module) for matching, and multi-scale second-order moment (M$^2$ block) for strong representation. Specifically, M$^2$ block develops a collection of semantic second-order descriptors at multiple spatio-temporal scales. Furthermore, A$^2$ module aims to adaptively select informative candidate descriptors while considering the individual motion pattern. By such means, our A$^2$M$^2$-Net is able to handle the challenging temporal misalignment problem by establishing an adaptive alignment protocol for strong representation. Notably, our proposed method generalizes well to various few-shot settings and diverse metrics. The experiments are conducted on five widely used FSAR benchmarks, and the results show our A$^2$M$^2$-Net achieves very competitive performance compared to state-of-the-arts, demonstrating its effectiveness and generalization.}

\keywords{Multi-scale spatio-temporal feature, Second-order statistics, Few-shot action recognition, Video understanding}

\maketitle

\section{Introduction}\label{sec:introduction}
  The few-shot action recognition (FSAR) problem aims to recognize the unknown instance video (query) with limited annotated samples (support set). Due to its role in alleviating the dependence on the costly large-scale annotation required for deep neural network, FSAR has attracted increasing attention in recent years. Research on FSAR can be roughly summarized into two categories: metric learning and representation learning. Among them, metric learning~\cite{OTAM_2020, Hybrid_2022, MTFAN_2022, CMOT} aims to compare the support-query instance pairs. Due to the sequential characteristics of video instance, FSAR can be specially formulated as a temporal alignment problem~\cite{OTAM_2020, ITA_2021, MTFAN_2022, CMOT}. Apart from the advanced metric, the discriminability of approach greatly benefits from an informative spatio-temporal representations~\cite{AMFAR_cvpr23, RVN_CVIU, TBSN, MoLo_cvpr23, TARN_bishay2019tarn, ZHANG_PR, TRX_2021, MGCSM_MM23}. Combined with a stronger representation, the metric can consistently exhibit more competitive performance.

\begin{figure}
		\begin{minipage}[t]{0.45\linewidth}
			\centering
   			\includegraphics[width=3.in]{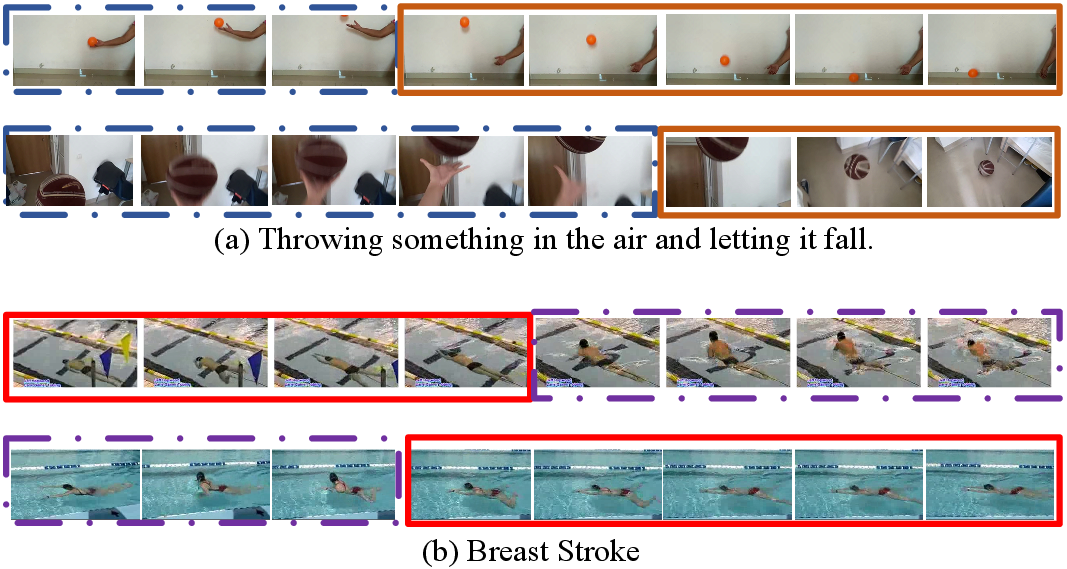}\\
			\vspace{0.02cm}
		\end{minipage}%
	\caption{Some examples of challenging scenarios. The same subactions across instances are marked with boxes in the same color. In the upper video pair, which is labeled \textit{Throwing something in the air and letting it fall}, the durations of two key subactions: \textit{throwing in the air} (highlighted in dashed \textcolor[RGB]{55,92,146}{blue} box) and \textit{falling down} (highlighted in solid \textcolor{brown}{brown} box) vary significantly. The lower video pair is labeled \textit{Breast stroke}. The critical subactions, \textit{stroking} \& \textit{inhaling} (indicated by dashed box in \textcolor[RGB]{101,52,154}{purple}) and \textit{leg kicking} (indicated by the solid  \textcolor[RGB]{255,0,0}{red} box), occur in reverse order across the two instances. These variations in duration and order lead to temporal misalignment, posing significant challenges to temporal alignment.}
	\vspace{-0.2cm}
	\label{fig:example}
\end{figure}

Despite significance advance, \mnrevision{existing methods frequently built with 2D networks struggle to handle the challenging temporal misalignment lying in video dynamics}. To be specific, due to diverse durations and different order of subactions, FSAR is challenged by the temporal misalignment problem as shown in Figure~\ref{fig:example}. Some metric-based approaches rely on a strict temporal order~\cite{OTAM_2020} or a set-to-set fashion (namely, temporally unordered)~\cite{Hybrid_2022}, making it difficult to flexibly align the unconventional temporal instance pairs as mentioned above. 
In representation learning, several promising works~\cite{TRX_2021, MGCSM_MM23} attempt to tackle this challenge by utilizing a multi-scale mechanism to capture temporal dynamics. However, they often equally utilize the representations from various scales for recognition, neglecting the individual scale preference. In addition to overlooking instance preferences, the discriminability of features is also underexplored. Specifically, leading studies typically perform matching based on global average pooling outputs, which represent first-order statistics~\cite{PAMI20}. Thus, they fail to fully exploit the benefits of rich feature statistics, which could potentially impede accurate comparisons.

\mnrevision{To address these limitations, we propose an adaptively aligned multi-scale moment network (A$^2$M$^2$-Net), aiming to achieve individual temporal alignment with powerful representations based on 2D networks.} A$^2$M$^2$-Net introduces an adaptive temporal alignment scheme (A$^2$ module) and elaborates on a discriminative multi-scale spatio-temporal moment (M$^2$ block). Specifically, M$^2$ block formulates semantic second-order temporal sequences across various observation scales to fully utilize feature statistics. This results in a collection of discriminative multi-scale spatio-temporal features, which are desired to comprehensively cover the complex temporal dynamics in videos. Furthermore, considering individual motion preferences of a given sample, A$^2$ module matches above multi-scale representation candidates in an interactive manner which aligns with an optimal transportation problem and realized by the Earth Mover's Distance (EMD) metric~\cite{EMD_2000}.
By promoting informative representations and suppressing unserviceable ones, A$^2$ module achieves adaptive temporal alignment guided by instance motion preferences. Ultimately, by integrating M$^2$ block and A$^2$ module into the top of network, A$^2$M$^2$-Net provides a data-driven alternative for temporal alignment approach on powerful representations.
To evaluate the effectiveness of A$^2$M$^2$-Net, we conduct experiments on five popular FSAR benchmarks: Something-Something V2 Full~\cite{OTAM_2020} (SSV2-Full), Something-Something V2 Small~\cite{CMN_zhu2018} (SSV2-Small), Kinetics-100~\cite{CMN_zhu2018} (K-100), UCF-101~\cite{UCF101_arxiv} and HMDB-51~\cite{HMDB_2011_ICCV}. The contributions of this work can be summarized as follows:
\begin{itemize}                                 
\item [1)]
We propose an adaptively aligned multi-scale moment (A$^2$M$^2$-Net) approach for FSAR. To be specific, A$^2$M$^2$-Net develops a paradigm  consisting of adaptive temporal alignment scheme and powerful multi-scale representations. By jointly considering the feature modeling and alignment procedures, A$^2$M$^2$-Net employs selective representative features for recognition, and thus effectively alleviates the temporal alignment challenge caused by video dynamics.
\item [2)]
Furthermore, A$^2$M$^2$-Net proposes a spatio-temporal multi-scale second-order moment (M$^2$ block), describing the complex temporal dynamics with discriminative representation. To achieve this, M$^2$ block develops a simple-yet-effective multi-scale mechanism to cover the latent diverse instance motion patterns. In each scale, M$^2$ block leverages rich feature statistics to effectively characterize the spatio-temporal attributes. 
\item [3)]
\mnrevision{Extensive experiments on various FSAR benchmarks show our A$^2$M$^2$-Net is very competitive to and performs better than state-of-the-arts by using 2D networks.}
\end{itemize}

\section{Related work}
\subsection{Few-Shot Image Recognition (FSIR)}
As a fundamental topic, FSIR drives other downstream few-shot learning studies. The majority of literatures can be coarsely summarized into the following lines: metric learning, representation learning and optimization algorithm. 
The metric-based approach aims to develop an advanced metric for comparing support and query images, typically in a prototypical~\cite{proto_2017} manner. The metric can be learnable~\cite{relationnet} or a predefined formulation~\cite{DC,SAML,DeepBDC_22,DeepEMD_2020, DeepEMD_PAMI}. In addition, some representation learning approaches develop task-specific representation with a cross-attention block~\cite{CTX, FSL_Transf_TCSVT} or a feature reconstruction mechanism~\cite{feat_2020}. Moreover, some works~\cite{
ModelAgnostic, Wang2022GlobalCO, Antoniou2018HowTT, Andrychowicz2016LearningTL, Ravi2016OptimizationAA} introduce new optimization algorithms with the goal of speeding up the optimization process or estimating an accurate gradient descent direction for the limited training samples.
In contrast to these FSIR studies, our A$^2$M$^2$-Net discuss the FSAR task, tailored for the video instance, which is more challenging due to temporal diversity.

\subsection{Few-Shot Action Recognition (FSAR)}
FSAR represents a more realistic extension of action recognition~\cite{i3d_Carreira_2017_CVPR, R2+1D_tran_2018_CVPR, SlowFast_Feichtenhofer_2019_ICCV, X3D_Feichtenhofer_2020_CVPR, AMSNet, VLG_IJCV, TDO_IJCV, EAN_IJCV, C2F_IJCV} in the few-shot learning scenario. Overall, the studies are typically towards metric learning and spatio-temporal representation learning. 
Along the metric learning route, OTAM~\cite{OTAM_2020} emphasizes the temporal order by employing a soft-DTW metric, thereby imposing strict temporal constraints. In contrast, HyRSM~\cite{Hybrid_2022} introduces a favorable Bidirectional Mean Hausdorff Metric (Bi-MHM), in which the sequence is regarded as a disordered set; SA-TAP~\cite{TAP_2021} leverages a learnable subnetwork for adaptive transformation of frame-to-frame similarity matrices, allowing dynamic adjustment. In addition, EMD~\cite{EMD_2000}, as investigated in MTFAN~\cite{MTFAN_2022} and CMOT~\cite{CMOT}, can be used for sequence matching. Among these works, MTFAN and CMOT are the most closely related to our A$^2$M$^2$-Net. However, both of them neither consider the diverse, fine-grained spatio-temporal features for latent video dynamics nor fully utilize the merits of feature statistics to enhance EMD alignment.

Capturing the spatio-temporal characteristics from multiple  levels is a simple-yet-effective solution for handling the video dynamics, particularly in the few-shot scenario. \mnrevision{Most existing approaches are built with 2D networks.} Specifically, CML~\cite{CML} creates cascade interactions~\cite{TPN_CVPR20} across multiple layers; HCL~\cite{HCL_eccv22} engages in patch-level matching for the optimal frame pair; M$^3$Net~\cite{M3Net_MM23}, MTFAN~\cite{MTFAN_2022} and CLIP-M$^2$DF~\cite{CLIP_M2DF_arxiv24} provide perspectives at both global video level and local frame level. Focusing on the temporal dimension, TRAPN~\cite{TRAPN}, TRX~\cite{TRX_2021} and STRM~\cite{STRM_2022} explore the multi-scale temporal relations through frame tuples~\cite{Zhou_2018_ECCV}, offering a finer-grained temporal description. However, these approaches often treat the various spatio-temporal views equally, overlooking the distinct roles in recognition due to individual motion preferences. 
\mnrevision{In addition, TARN~\cite{TARN_bishay2019tarn}, MASTAF~\cite{MASTAF} and SAFSAR~\cite{SAFSAR} are developed on 3D networks, benefiting from the spatio-temporal features learned by backbone themselves. Meanwhile}, some researchers~\cite{CLIP_FSAR_arXiv23, MACLIP_arXiv23} have delved into comparing the visual embedding features with text ones in a vision-language (VL) contrastive paradigm~\cite{CLIP_ICML, Actionclip_TNNLS}, \revision{by devising temporal modeling module~\cite{CLIP_FSAR_arXiv23} at the top of network or exploring the multi-modality interaction manner~\cite{MACLIP}. Inspired by these works, we also evaluate our A$^2$M$^2$-Net on CLIP framework.}

\begin{figure*}
		\begin{minipage}[t]{0.99\linewidth}
			\centering
			\includegraphics[width=6.25in]{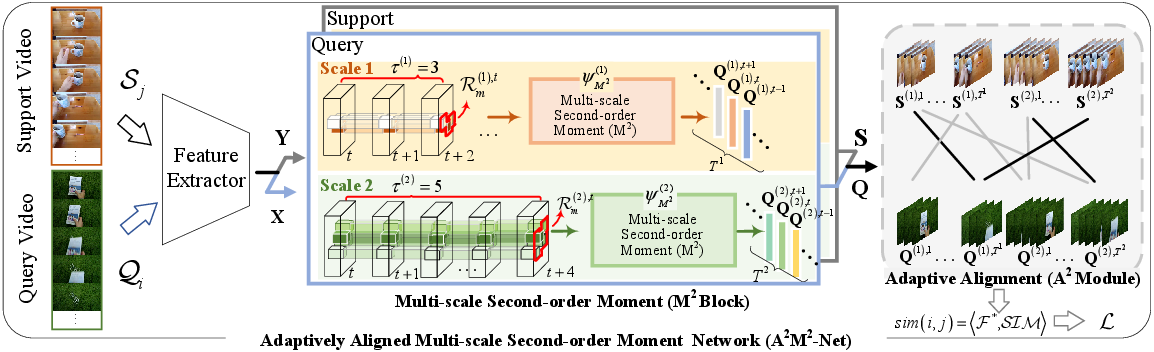}\\
		\end{minipage}%
	\caption{Overview of A$^2$M$^2$-Net. The extracted features of query video ($\mathbf{X}$) and support video ($\mathbf{Y}$) are fed into multi-scale blocks independently, yielding multi-scale spatio-temporal features (i.e., $\mathbf{Q}^{(1)}, \mathbf{Q}^{(2)}$ and $\mathbf{S}^{(1)}, \mathbf{S}^{(2)}$ respectively). In scale $b$, M$^2$ block develops a scale-related semantic second-order representation. By varying the spatio-temporal scales $\tau^{(b)}\times\mathcal{R}^{(b)}$, $\psi_{M^2}$ outputs a collection of representative motion descriptors, aiming to cover the latent video dynamics. Ultimately, A$^2$ module performs an instance-guided temporal alignment for these feature candidates, selecting favorable features for temporal alignment.}
	\vspace{-0.2cm}
	\label{fig:overview}
\end{figure*}

\section{Method}\label{sec:method}

In this section, we first give a problem formulation for the FSAR task. Then, we demonstrate an overview of A$^2$M$^2$-Net framework. Afterwards, we provide a detailed introduction to the proposed A$^2$M$^2$-Net, which consists of a multi-scale moment (M$^2$ block) for generating rich spatio-temporal descriptors and an adaptive alignment (A$^2$ module) for automatically choosing descriptor candidates in an implicit manner.


\subsection{Problem Formulation}~\label{sec:fsl_task}
The goal of FSAR is to develop a discriminative model with robust generalization capability, so that it can classify unseen videos (queries) with a limited number of labeled videos (support set). Specifically, the basic unit of inference, termed an episode, typically comprises query videos $\{\mathcal{Q}_1, \cdots, \mathcal{Q}_Z\}$ and a set of support videos $\mathcal{S} = \{\mathcal{S}_1,\cdots, \mathcal{S}_{NK}\}$ involving $N$ classes (ways) $\times$ $K$ samples (shots). In FSL settings, the number of available annotated samples $\mathcal{S}$ is generally quite small, (e.g., $N=5$, $K\in\{1,2,3,4,5\}$). In particular, to fully evaluate the generalization of the algorithm, the categories for training and test do not overlap.


\newcommand\fr[2]{\mathbf{#1}^{#2}}

\subsection{Overview of \texorpdfstring{A$^2$M$^2$-Net}{A2M2-Net} Framework}~\label{sec:pipeline}
The overview of A$^2$M$^2$-Net framework is illustrated in Figure~\ref{fig:overview}. 
Both the query video $\mathcal{Q}_i$ and support video $\mathcal{S}_j$ are processed through a shared feature extractor and a subsequent multi-scale moment (M$^2$ module). The output multi-scale spatio-temporal descriptors are then sent to an adaptive alignment (A$^2$) block for individualized selective matching across scales. The alignment similarity is subsequently used to compute the loss.

Given a query video $\mathcal{Q}_i$, its deep sequential features, with a temporal length $T$, are represented by $\mathbf{X}=\{\fr{X}{t}\}_{t=1}^{T}$. These features are then input into a multi-scale (for instance, $B$ scales) spatio-temporal moment (M$^2$) module, $\psi_{M^2}$. For each scale-$b$ of M$^2$ module, it produces a sequence of second-order moments with length $T^b$: $\fr{Q}{(b)}=\{\fr{Q}{(b),t}\}_{t=1}^{T^b}$. Ultimately, the outputs of all $B$ scales for $\mathcal{Q}_i$, are respectively collected ($\mathbf{Q}=\{\mathbf{Q}^{\left(b\right)}\}_{b=1}^{B}$) and the outputs of $\mathcal{Q}_i$ and $\mathcal{S}_j$ are adaptively aligned (A$^2$) by an EMD metric, which can be regarded as an implicitly task-specific scale feature selection protocol.

One query video $\mathcal{Q}_{i}$ is individually compared with various support instances $\mathcal{S}_{*}$. Their output similarity scores, denoted as $sim\left(i,*\right)$, serve as logits for cross-entropy loss and inference. 
In particular, when the number of shots ($K$) exceeds one,  
it learns prototype for each category by leveraging $NK$ labeled support instances, following the suggestion in~\cite{DeepEMD_2020}. Each prototype is also organized as a multi-scale representation, initialized by averaging $K$ support samples within one class.
And for sake of clarity, we take $K$=1 as an example in subsequent demonstration.

\begin{figure}
    \centering
    \includegraphics[width=0.48\textwidth]{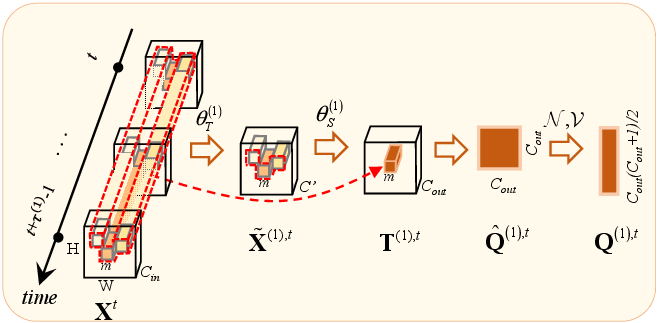}
    \caption{Illustration of \revision{scale $(b)$} in multi-scale second-order moment, i.e., $\revision{\psi_{M^2}}$.}
    \label{fig:M2}
\end{figure}


\newcommand\wfr[2]{\widetilde{\mathbf{#1}}^{#2}}
\newcommand\hfr[2]{\hat{\mathbf{#1}}^{#2}}
\newcommand\gray[1]{\textcolor[RGB]{120,120,120}{#1}}

\subsection{Multi-Scale Spatio-temporal Second-order Moment (\texorpdfstring{$\text{M}^2$}{M2} Block)}~\label{sec:pyramid}
To address the challenges of spatio-temporal diversity, it demands a more potent representation that effectively captures the essence of spatio-temporal dynamics. Drawing inspiration from the significant promise shown by high-order video statistics in prior research~\cite{TCP_neurips_2021}, we introduce a multi-scale temporal moment approach, whose goal is to take fully advantage of high-order statistical information in videos, particularly in few-shot setting.

To be specific, let the extracted feature be $\mathbf{X}\in \mathbb{R}^{T\times C_{in} \times M }$, where $C_{in}$ represents the number of channels and $M$ is the reshaped size of height ($H$) and width ($W$), i.e., $M=H\times W$. M$^2$ module initially develops a multi-scale scheme with scale-specific function $\{\psi_{M^2}^{(b)}\}_{b=1}^B$, which can be formulated by:

\begin{equation}\label{eq:ms_Q}
\left\{ 
    \begin{aligned}
    &\hat{\mathbf{Q}}^{\left(1\right)} = \psi^{(1)}_{M^2}\left(\mathbf{X}\right), \cr 
    & ~~~~~~~~~~~\vdots  \cr
    &\hat{\mathbf{Q}}^{\left(B\right)} = \psi^{(B)}_{M^2}\left(\mathbf{X}\right).
    \end{aligned}
\right.
\end{equation}
The notation $\hat{\mathbf{Q}}^{\left(b\right)}$ represents the $b$-th ($b=1,\cdots,B$) scale sequential output with length $T^b$: $\hat{\mathbf{Q}}^{(b)} = {\{\hat{\mathbf{Q}}^{(b), t}}\}_{t=1}^{T^b}$. Here  $\hat{\mathbf{Q}}^{(b), t}$ represents the $t$-th element in sequence $\hat{\mathbf{Q}}^{(b)}$, where $t$ is the temporal index. Particularly, $\hat{\mathbf{Q}}^{(b), t}$ is expected to portray scale-related characters by aggregating spatio-temporal features, utilizing a specialized second-order moment layer $\psi_{M^2}^{(b)}$ as follows:

\begin{equation}\label{eq:M2_input_x_new}
    \revision{\fr{\hat{Q}}{\left(b\right),t} = \frac{1}{M} \sum_{m=1}^M \fr{T}{\left(b\right),t}_m \left(\fr{T}{\left(b\right),t}_m\right)^{\top}.}
\end{equation}
\revision{As shown in Equation~\ref{eq:M2_input_x_new}, $\fr{\hat{Q}}{\left(b\right),t}\in\mathbb{R}^{C_{out}\times C_{out}}$ is obtained by summarizing second-order statistics of semantic feature $\mathbf{T}^{(b),t}\in\mathbb{R}^{C_{out}\times M}$, where $C_{out}$ denotes output channel and the notation $\top$ refers to matrix transpose operation. }

Furthermore, due to the latent appearance variation along temporal dimension, we consider a long-term deformable~\cite{deform} scheme on $\mathbf{T}^{(b),t}$ to simulate the temporal trajectory~\cite{Traj_nips18}, implemented as following:
\begin{equation}
   \revision{ \mathbf{T}^{(b)} = \theta_S^{(b)}(\theta_T^{(b)}(\mathbf{X})).}
\end{equation}

\revision{Concretely, $\theta_T^{(b)}$ refers to a temporal convolution with kernel size $\tau^{(b)}$, producing the output feature $\widetilde{\mathbf{X}}^{(b)}$. $\theta_S^{(b)}$ denotes a deformable convolution layer~\cite{deform} with a predefined gridded kernel $\mathcal{R}^{*(b)}$ (e.g., $3^2$\footnote{Here, the notation $3^2$ refers to a simplification of spatial size $3\times 3$.}) and offset $\Delta^{(b)}$. To enhance temporal awareness, $\Delta^{(b)}$ is generated by applying an MLP layer $\mathcal{K}^{(b)}$ to the temporal difference of $\widetilde{\mathbf{X}}^{(b)}$, specifically $\widetilde{\mathbf{X}}^{(b),t}_{\Delta} = \widetilde{\mathbf{X}}^{(b),t} - \widetilde{\mathbf{X}}^{(b),t-1}$. Notably, the gridded kernel $\mathcal{R}^{*(b)}$ combined with $\Delta^{(b)}$ results in our flexible spatial neighborhood $\mathcal{R}^{(b)}$ for scale $b$, as the highlighted red region in Figure~\ref{fig:overview}. Furthermore}, $\mathcal{R}^{(b)}$ and $\tau^{(b)}$ here respectively reflect the scale-exclusive spatial and temporal receptive field for output $\mathbf{T}^{(b)}$ in scale $b$, and thus their combination spanning various scales constructs a multi-scale spatio-temporal observation pattern as the formulation in Equation~\ref{eq:ms_Q}.

The output $\{\hat{\mathbf{Q}}^{(b),t}\}_{b=1,t=1}^{B,T^b}$ is covariance matrix, whose space is a Riemannian manifold~\cite{pennec2006riemannian, arsigny2005fast}, and is beneficial from a robust estimation~\cite{MPNCOV_Li_2017_ICCV} for its geometrical structure.
Therefore, we adopt a square-root normalization~\cite{iSQRT_2018_CVPR} for $\mathbf{\hat{Q}}$ (dubbed $\mathcal{N}\left(\cdot\right)$). Subsequently, as a covariance matrix is symmetric, we conserve its  upper triangular entries (including the diagonals) and produce a vectorial representation denoted by $\mathcal{V}\left(\cdot\right)$. The ultimate output $\mathbf{Q}^{(b),t}=\mathcal{V}\left(\mathcal{N}\left(\fr{\hat{Q}}{(b),t}\right)\right)\in\mathbb{R}^{C_{out}(C_{out}+1)/2}$ is employed for following adaptive alignment. The process of generating $\mathbf{Q}^{(b),t}$ is depicted in Figure~\ref{fig:M2}.

\noindent{\textbf{Discussion}} To summarize, the multi-scale spatio-temporal second-order moment provides sufficiently powerful motion descriptor representations. Specifically, from the aspect of feature diversity, it describes the video dynamics with a spatio-temporal multi-scale protocol $\psi_{M^2}$. Furthermore, from the feature distribution view, it  elaborates a semantic second-order moment for each scale by scale-specialized layer $\psi_{M^2}^{(b)}$. 
In contrast to the simple spatio-temporal-isolated second-order moment (denoted as Cov-MN), $\psi_{M^2}^{(b)}$ additionally considers the spatio-temporal semantics by $\theta_T$ and $\theta_S$, thereby our semantic second-order moment broadens the spatio-temporal receptive field from a fixed $1\times 1^2$ to $\tau^{(b)} \times \mathcal{R}^{\left(b\right)}$, as illustrated in Figure~\ref{fig:overview}. 
Diversifying various scales, $\psi_{M^2}$ results in a collection of fine-grained representations with distinct spatio-temporal receptive field $\{\tau^{(b)} \times \mathcal{R}^{\left(b\right)}\}_{b=1}^{B}$, as highlighted by orange region and green region in Figure~\ref{fig:overview}. Given these rich and powerful representation candidates, we will subsequently discuss \textit{how to achieve an adaptive alignment for query and support instance pairs.}


\subsection{Adaptive Alignment (\texorpdfstring{A$^2$}{A2} Module)}~\label{sec:emd}
The similarity of query and support instance representation output by M$^2$ block spanning various scales ($\mathbf{Q}$ and $\mathbf{S}$) can be obtained by a temporal alignment metric $\mathcal{M}$:
\begin{equation}\label{eq:FSL_math}
     \mathcal{M}\left(
     \{\fr{Q}{[1]},\cdots, \fr{Q}{[L]}\}, 
     \{\fr{S}{[1]},\cdots, \fr{S}{[L]}\}
     \right).
\end{equation}
The notation $\fr{Q}{[l]}$ or $\mathbf{S}^{[l]}$ $(l=1,\cdots,L)$ indicates the $l$-th representation in the sequence $\mathbf{Q}$ or $\mathbf{S}$, which alternates the dual scale and time stamp indexes which a single index $l$ for clarification. Thereby, its maximum temporal length $L$ equals to $\sum_{b=1}^B T^b$.
 
Due to the latent diverse subaction duration for different instances, it is favorable to make an adaptive alignment for $\mathbf{Q}$ and $\mathbf{S}$, which encourages each representation to find the informative candidate among all scales \& time stamps. 
As such, the selection plan corresponds to the transportation cost for moving $\mathbf{Q}$ to $\mathbf{S}$. Mathematically, it can be formulated as an Earth Mover's Distance (EMD)~\cite{EMD_2000}. 

In EMD methodology, the similarity score of query video $\mathcal{Q}_i$ and $\mathcal{S}_j$, $sim(i,j)\in\mathbb{R}$, can be defined as:
\begin{equation}\label{eq:emd_sim}
     sim(i,j)=\mathcal{M}\left(\mathbf{Q},\mathbf{S}\right)=\left<\mathcal{SIM}, \mathcal{A}^{*} \right>,
\end{equation}
where $\mathcal{A}^*\in\mathbb{R}^{L\times L}$ is the optimal alignment matrix to be solved; $\mathcal{SIM}\in\mathbb{R}^{L\times L}$ indicates a similarity matrix given by:
\begin{equation}\label{eq:sigma_xy}
    \mathcal{SIM}_{l,l'}=\frac {\left<\fr{Q}{[l]},\fr{S}{[l']}\right>} {\|\fr{Q}{[l]}\| \cdot \|\fr{S}{[l']}\|}.
\end{equation}
 The symbol $\langle \cdot, \cdot \rangle$ and $\| \cdot \|$ denote the inner product and Frobenius norm respectively. The optimal alignment matrix $\mathcal{A}^*$ can be obtained by solving a linear programming problem:
\begin{equation}\label{eq:emd_goal}
     \begin{cases}
        \mathcal{A}^*=\mathop{\arg\min}\limits_{\mathcal{A}} \left< 1-\mathcal{SIM},\mathcal{A} \right> \quad \\
        \begin{aligned}
        s.t.~~~~ & \sum_{l'=1}^{L}\mathcal{A}_{l,l'} = \mu_{l}, \\
       & \sum_{l=1}^{L}\mathcal{A}_{l,l'} = \gamma_{l'}, \\
       & \mathcal{A}_{l,l'}\ge 0.
        \end{aligned}
     \end{cases}
\end{equation}
Moreover, marginal probability masses $\mu, \gamma\in\mathbb{R}^L$ are determined through a cross-reference mechanism as suggested by~\cite{DeepEMD_2020}, extended to temporal dimension in our case:
\begin{equation}\label{eq:mu}
\begin{split}
        \mu_{l} = \left< \fr{Q}{[l]} ,\sum_{i=1}^{L}{\fr{S}{[i]}/L} \right>,
        \gamma_{l'} = \left< \fr{S}{[l']} , \sum_{i=1}^L {\fr{Q}{[i]}/L}\right>.
\end{split}
\end{equation}

\revision{In such dynamic temporal alignment mechanism, for each query descriptor $\mathbf{Q}^{[l]}$ that has low similarity to the support features (quantified by $\mu_{l}$ in Eqn.~\ref{eq:mu}), its alignment weight $\mathcal{A}_{l,:}$ is decreased (due to the first constraint in Eqn.~\ref{eq:emd_goal}), reflecting its insignificance to this task. On the contrary, when $\mathbf{Q}^{[l]}$ is highly similar to the support descriptors, its alignment weight is increased, indicating that it plays an important role in the alignment process. Thus, such dynamic process of alignment weight adjustment could select important descriptor, which is namely \textit{task-specific selection mechanism} in this paper.}

Ultimately, the similarity score between $\mathcal{Q}_i$ with $N$-support instances, i.e.,  $sim(i,*)\in\mathbb{R}^{N}$, are used to construct a softmax-like classifier.

\newcommand\sotaf[1]{\textcolor{red}{\textbf{#1}}}
\newcommand\sotas[1]{\underline{#1}}
\newcommand{\acc}[2]{#1$/$#2\%}
\newcommand\accthree[3]{#1$/$#2$/$#3$\%$}

\section{Experiments}

\subsection{Experimental Settings}
\noindent\textbf{Dataset} We conduct experiments on five popular few-shot action recognition datasets: SSV2-Full, SSV2-Small, K-100, UCF-101 and HMDB-51. Both SSV2-Full~\cite{OTAM_2020} and SSV2-Small~\cite{CMN_zhu2018} are subsets of the Something-Something V2 dataset~\cite{ssv2}, tailored for the few-shot setting with different base set scopes. The K-100~\cite{CMN_zhu2018} collection is a subset of  Kinetics-400~\cite{i3d_Carreira_2017_CVPR} dataset. UCF-101~\cite{UCF101_arxiv} and HMDB-51~\cite{HMDB_2011_ICCV} are reorganized for few-shot action setting.

\noindent\textbf{Training and Inference} In our approach, we \revision{extensively evaluate our A$^2$M$^2$-Net across various backbones, including 2D network (ResNet-50~\cite{ResNet_He_2016_CVPR} architecture pre-trained on ImageNet-1K~\cite{ImageNet-1K} dataset), 3D network (VideoMAE~\cite{MAE_2022} based on ViT-B) as well as multi-modality network (CLIP~\cite{CLIP} ViT-B/16). Unless otherwise specified, we use ResNet-50 as default configuration,} consistent with leading research~\cite{TRX_2021, STRM_2022, CPM_2022, Hybrid_2022, MoLo_cvpr23}.

The linear programming problem of EMD is solved by an off-the-shelf toolkit~\cite{OpenCVDocs}. $C_{in}$ is 2048 for ResNet-50; $C'$ and $C_{out}$ are set to 256 and 128, respectively. For video processing, we set the input resolution to $8\times 224 \times 224$ with uniform sampling~\cite{TSN_PAMI}. For data augmentation across all datasets, we implement scale jittering and random cropping, and additionally, we apply random horizontal flipping to all benchmarks except for the SSV2 datasets. Our training protocol involves pre-training processing followed by meta-training, as outlined in~\cite{CPM_2022}. For smaller datasets, BN remains frozen throughout the training phase. We employ SGD as the optimizer with a weight decay of 5e-4. 
In the pre-training process, we utilize a mini-batch size of 48. The initial learning rate, set to 9e-4 for SSV2-Full, 1.5e-3 for SSV2-Small and HMDB-51, and 1.25e-4 for K-100 and UCF-101, is decayed by a factor of 0.1 when the performance saturates.
Specifically, randomly initialized parameters are subjected to a higher learning rate for a faster convergence. To prevent overfitting, we adopt a dropout layer preceding the FC layer with rate 0.85. The mean accuracy across \revision{10,000} episodes is reported on a single-crop view as performance metric\revision{, accompanied by the 95\% confidence intervals for the results compared with the state-of-the-arts.}

\newcommand{\tauhl}[1]{\textcolor{colortauorg}{#1}}

\newcommand{\Rhl}[1]{\textcolor{colorR}{#1}}
\newcommand{\tube}[2]{$\tauhl{#1} \times \Rhl{#2}^2$}

\newcommand{\gap}[1]{\colorbox{colorbaseline}{#1}}
\newcommand{\psp}[1]{\colorbox{colorpsp}{#1}}
\newcommand{\ssgap}[1]{\colorbox{colorist}{#1}}
\newcommand{\hlours}[1]{\colorbox{colorA2M2}{#1}}
\newcommand{\accval}[2]{#1$/$#2}


\newcommand{\std}[1]{\scriptsize{$_{\pm#1}$}}
\newcommand{\stdd}[1]{\revision{\tiny{$_{\pm#1}$}}}
\begin{table*}[!htb]
\centering
\footnotesize
\tablestyle{0.5pt}{1.7}
\renewcommand\arraystretch{1.5}
\caption{5-Way accuracy (\%) on four datasets. $\star$: Results are adopted from literatures~\cite{OTAM_2020, Hybrid_2022}. All the results are evaluated on ResNet-50 architecture pretrained in ImageNet-1K, except for $\dag$, which is pretrained on Sports-1M dataset with C3D network. The best and second performances with same settings are noted by the bold in \textcolor{red}{red} and \underline{underline}, respectively.}\label{tab:sota_overview}
\begin{tabular}{l|ccc|ccc|ccc|ccc}
\hline
\multirow{2}{*}{Method} & \multicolumn{3}{c|}{SSV2-Full} & \multicolumn{3}{c|}{SSV2-Small} &  \multicolumn{3}{c|}{HMDB-51} & \multicolumn{3}{c}{UCF-101}  \\ 
& 1-shot & 3-shot & 5-shot & 1-shot & 3-shot & 5-shot & 1-shot & 3-shot & 5-shot & 1-shot & 3-shot & 5-shot \\

\hline
MatchingNet~\cite{matchingnet}$^{\star}$ & --   & --  & --   & 31.3 &      & 45.5  & -- & -- &--  & -- &  -- & --\\
MAML~\cite{maml_2017}$^{\star}$          & --   & --  & --   & 30.9 &      & 41.9  & -- & -- & -- & -- &  -- & -- \\  
CMN~\cite{CMN_zhu2018}$^{\star}$         & 36.2 &     & 48.8 & --   &  --  &  --   & -- & -- & -- & -- & --  & --\\ 
OTAM~\cite{OTAM_2020}$^{\star}$          & 42.8 & 51.5 & 52.3 & 36.4 & 45.9 & 48.0  & 54.5 & 65.7 & 68.0 & 79.9 & 87.0 & 88.9 \\
CMOT~\cite{CMOT}$^{\dag}$ & 46.8 \stdd{0.5} & -- & 55.9 \stdd{0.4} & -- & -- & -- & 66.9 \stdd{0.5} & -- & 81.6 \stdd{0.4} & 87.2 \stdd{0.4}  & -- & 95.7 \stdd{0.3} \\
Nguyen \textit{et al.}~\cite{Nguyen_ECCV22} & 43.8 \stdd{0.4}  & -- & 61.1 \stdd{0.4} & -- & -- & -- & 59.6 \stdd{0.4} & -- & 76.9 \stdd{0.4} & 84.9 \stdd{0.3} & -- & 95.9 \stdd{0.2} \\
ITANet~\cite{ITA_2021} & 49.2 \stdd{0.2} & 59.1 \stdd{0.2}  & 62.3 \stdd{0.3} & 39.8 \stdd{0.2} & 49.4 \stdd{0.4} & 53.7 \stdd{0.2} & -- & -- & -- & -- & -- & --\\ 
SA-TAP~\cite{TAP_2021} & 45.2 \stdd{0.4} & -- & 63.0 \stdd{0.4} & -- & --   & -- & 57.5 \stdd{0.4} & -- & 74.2 \stdd{0.4} & 83.9 \stdd{0.3} & -- & 95.4 \stdd{0.2} \\ 
PAL~\cite{PAL_21} & 46.4 & -- & 62.6 & -- &  --  & -- & 60.9  & -- & 75.8 & 85.3 & -- & 95.2\\
TRX~\cite{TRX_2021} & 42.0 & 57.7 & 64.6 & -- & -- & 59.1 & -- &  -- & 75.6 & -- & -- & 96.1 \\ 
STRM~\cite{STRM_2022} & 43.1 & 62.0 & 68.1 & -- &  -- &  55.3 & -- & -- & 77.3 & -- & -- & \sotaf{96.9} \\
MTFAN~\cite{MTFAN_2022} & 45.7 &  -- & 60.4  & -- & -- &  --  & 59.0 & -- & 74.6 & 84.8 & -- & 95.1 \\ 
HyRSM~\cite{Hybrid_2022} & 54.3 & 65.1 & 69.0 & 40.6 & 52.3 & 56.1 & 60.3 & 71.7 & 76.0 & 83.9 & 93.0 & 94.7 \\ 
Huang~\textit{et al.}~\cite{CPM_2022} & 49.3 &  --  & 66.7 & 38.9 & --  & \sotaf{61.6} & 60.1 & -- & 77.0 & 71.4  & -- & 91.0 \\ 
HCL~\cite{HCL_eccv22} & 47.3 & 59.0 & 64.9 & 38.7 & 49.1 & 55.4 & 59.1 & 71.2 & 76.3 & 82.6 & 91.0 & 94.5 \\
MT-PRENet~\cite{Liu2022MultidimensionalPR} & 42.1 & -- & 58.4 & --  & -- & -- & 57.3 & 72.4 & 76.8 & 82.0 & -- & \sotas{96.4} \\
SA-CT~\cite{STA_CT_MM23} & 48.9 & -- & 69.1 & -- & -- & -- & 60.4 & -- & \sotaf{78.3} & 85.4 & -- & \sotas{96.4} \\
MASTAF~\cite{MASTAF} & 46.9 & -- & 62.4 & 37.5 & -- & 50.2 & 54.8 & -- & 67.3  & 79.3 & -- & 90.3 \\
MoLo~\cite{MoLo_cvpr23} & \sotas{56.6} & \sotas{67.0}  & \sotas{70.6} & \sotas{42.7} & 52.9  & 56.4 & 60.8 & \sotas{72.0} & \sotas{77.4} & \sotas{86.0} & 93.5 & 95.5 \\
GgHM~\cite{GgHM_ICCV23} & 54.5 & -- & 69.2 & -- & -- & -- & \sotas{61.2} & -- & 76.9 & 85.2 & -- & 96.3 \\
RFPL~\cite{RFPL_ICCV23} & 47.0 & 58.3 & 61.0 & -- & -- & -- & -- & -- & -- &  84.3 & 90.2 & 92.1 \\
\hline
A$^2$M$^2$-Net (ours) & \sotaf{56.7}\stdd{0.4} &  \sotaf{69.7}\stdd{0.4} & \sotaf{74.1}\stdd{0.3} & \sotaf{42.9}\stdd{0.4} &  \sotaf{55.2}\stdd{0.4} & \sotas{\revision{60.5}}\stdd{0.4} & \sotaf{61.8}\stdd{0.4} &  \sotaf{73.4}\stdd{0.4} & 76.6\stdd{0.3} & \sotaf{86.1}\stdd{0.3} &  \sotaf{\revision{94.0}}\stdd{0.2} & 95.8\stdd{0.2} \\
\hline
\end{tabular}
\end{table*}


\begin{table}[!htb]
\centering
\tablestyle{6pt}{2.}
\renewcommand\arraystretch{1.5}
\caption{5-Way accuracy on K-100 dataset. $\star$: Results are adopted from literatures~\cite{OTAM_2020}. Results are conducted on ResNet-50 pretrained on ImageNet-1K, except for $\dag$: ResNet-152, pretrained on ImageNet-11K. The best and second performances are noted by the bold in \textcolor{red}{red} and \underline{underline}, respectively.}\label{tab:sota_k100}
\begin{tabular}{lccc}
\hline
Method & 1-shot & 3-shot & 5-shot \\
\hline
MatchingNet~\cite{matchingnet}$^{\star}$ & 53.3 & 69.2 & 74.6\\ 
MAML~\cite{maml_2017}$^{\star}$          & 54.2 & 70.0 & 75.3 \\  
CMN~\cite{CMN_zhu2018}$^{\star}$         & 60.5 & 75.6 & 78.9 \\ 
OTAM~\cite{OTAM_2020}                    & 73.0 & 78.7 & 85.8 \\ 
ITANet~\cite{ITA_2021}                   & 73.6 \stdd{0.2} & -- & 84.3 \stdd{0.3} \\
TRX~\cite{TRX_2021}                      & 63.6 & 81.8 & 85.9\\  
TA$^2$N~\cite{TAN2N_li2022ta2n}           & 72.8 & -- & 85.8 \\
STRM~\cite{STRM_2022}                    & -- & 81.1 & 86.7 \\ 
MTFAN~\cite{MTFAN_2022}                  & \sotas{74.6} & -- & 87.4  \\  
HyRSM~\cite{Hybrid_2022}                 & 73.7 & 83.5 & 86.1  \\  
Huang~\textit{et al.}~\cite{CPM_2022}    & 73.3 & -- & 86.4 \\ 
HCL~\cite{HCL_eccv22}                    & 73.7 & -- & 85.8 \\
MoLo~\cite{MoLo_cvpr23}                  & 74.0 & 83.7 & 85.6 \\
\hline
A$^2$M$^2$-Net (ours)                    & 74.5\stdd{0.4} & \sotas{\revision{84.0}}\stdd{0.4} & \sotas{87.5}\stdd{0.3} \\
A$^2$M$^2$-Net$^\dag$ (ours)             & \sotaf{80.9}\stdd{0.4} & \sotaf{88.7}\stdd{0.4} & \sotaf{90.8}\stdd{0.3} \\
\hline
\end{tabular}
\end{table}

\subsection{Comparison with State-of-the-arts}

\mnrevision{We conduct an extensive comparison with state-of-the-arts by using 2D backbone of ResNet-50 on five popular FSAR datasets, where results of 1-shot$/$3-shot$/$5-shot are presented in Tables~\ref{tab:sota_overview} and \ref{tab:sota_k100}. It is clear that our A$^2$M$^2$-Net achieves highly competitive performance across all benchmarks.}

In terms of comparison with novel metric-based counterparts, A$^2$M$^2$-Net displays marked superiority over MTFAN~\cite{MTFAN_2022}, OTAM~\cite{OTAM_2020} and HyRSM~\cite{Hybrid_2022}. Specifically, on the challenging SSV2-Full dataset, A$^2$M$^2$-Net has a significant advantage with \accthree{11.0}{--}{13.7} compared with MTFAN. Notably, with stronger C3D network pretrained on Sports-1M dataset, CMOT~\cite{CMOT} is significantly inferior to our A$^2$M$^2$-Net with \accthree{9.9}{--}{18.2} on SSV2-Full dataset. Furthermore, A$^2$M$^2$-Net surpasses OTAM on five datasets, with margins exceeding \accthree{6.2}{7.1}{6.9} on four of these datasets. Additionally, it outperforms the highly competitive HyRSM by \accthree{2.4}{4.6}{5.1} and \accthree{2.3}{2.9}{4.4} on the SSV2 datasets.
This observation underlines the effectiveness of the enhancement of A$^2$M$^2$-Net to the original EMD algorithm, given that MTFAN, despite its reliance on the EMD metric, does not demonstrate similar superiority over OTAM and HyRSM. 

For the frame-level temporal alignment approaches, A$^2$M$^2$-Net outperforms ITANet~\cite{ITA_2021}, OTAM~\cite{OTAM_2020} and SA-TAP~\cite{TAP_2021} with a clear margin, typically showing over 2.0\% performance superiority on four of the reported benchmarks. 

In the line of temporal relation studies, A$^2$M$^2$-Net notably outperforms TRX~\cite{TRX_2021} and STRM~\cite{STRM_2022} by margins of \accthree{14.7}{12.0}{9.5} and \accthree{13.6}{7.7}{6.0} on SSV2-Full dataset respectively. 
Moreover, A$^2$M$^2$-Net competes effectively with the recent MoLo~\cite{MoLo_cvpr23} work, achieving a favorable advantage in the 3-shot and 5-shot settings, e.g., \acc{2.7}{3.5} and \acc{2.3}{4.1} on SSV2 datasets. 
Notably, in comparison with the latest GgHM~\cite{GgHM_ICCV23} and RFPL~\cite{RFPL_ICCV23} studies, our A$^2$M$^2$-Net surpasses them on all benchmarks, demonstrating an impressive performance advantage. Particularly, it is superior to GgHM and RFPL with margins of \accthree{2.2}{--}{4.9} and \accthree{9.7}{11.4}{13.1}  on the SSV2-Full dataset respectively.

The consistent and notable superiority observed across extensive benchmarks convincingly affirm the robustness of A$^2$M$^2$-Net, providing an effective alternative for FSAR task.


\begin{table*}[t]
\vspace{-3mm}
\subfloat[\textbf{High-order counterparts}: evaluate various first-order and second-order moment fashion, in single-scale setting. \revision{KP: Kernel Pooling.}\label{tab:sop}]
{
\tablestyle{0.5pt}{1.5}
\begin{tabular}{l|ccc}
pooling & order & 1-shot & 5-shot \\
\hline
GAP & 1 & \gap{45.12} & \gap{61.02} \\ 
CoV &  2  & 45.87 & 61.72 \\
BCNN~\cite{BCNN_lin_2015_ICCV} &  2  & 46.22 & 62.83 \\
Cov-MN~\cite{iSQRT_2018_CVPR} &  2  & \psp{46.80}  &  \psp{64.02} \\ 
\revision{KP}~\cite{KP_Cui_2017_CVPR} & \revision{1+2+3} & \revision{45.49} & \revision{61.43} \\
\end{tabular}
}
\hspace{2pt}
\subfloat[\textbf{Multi-temporal scale $\tauhl{\tau}$}: keep each spatial scale $\Rhl{\mathcal{R}^{(b)}}$ with $\Rhl{1^2}$. params: parameters.
\label{tab:tau}]{
\tablestyle{1pt}{1.3}
\begin{tabular}{c:c|ccc}
\multicolumn{1}{c}{scale} & \multicolumn{1}{c|}{$\tauhl{\tau}~\times$ \Rhl{$\mathcal{R}(1^2)$} } & params. & 1-shot & 5-shot \\ 
\hline
\multicolumn{1}{c}{single} & \multicolumn{1}{c|}{\tauhl{(1)}} & 23.7 M &  \psp{46.80 }&\psp{64.02} \\
\hdashline
\multirow{5}{*}{\rotatebox{90}{\parbox[c][0.75cm][c]{1.5cm}{multiple}}} & \tauhl{(1, 3)}  & +1.7 M & 55.14 & 72.18  \\
& \tauhl{(1, 5)}  & +2.7 M & 55.46 & 72.92 \\
& \tauhl{(1, 7)}   & +3.8 M & 54.64 & 72.56 \\
& \tauhl{(1, 3, 5)}   & +4.4 M & \ssgap{56.08} & \ssgap{73.19} \\
& \tauhl{(1, 5, 7)}  & +6.5 M & 55.08 & 73.43 \\
& \tauhl{(1, 3, 5, 7)}  & +8.1 M & 56.04 & 73.55 \\
\end{tabular}}
\hspace{2pt}
\subfloat[\textbf{Multi-spatial scale \Rhl{$\mathcal{R}^{\star}$}}: keep temporal scale $\tauhl{\tau}$ with optimal $\tauhl{(1,3,5)}$. params.: parameters. \Rhl{$\revision{\tau}^{\star}$}: initial gridded size for \Rhl{$\mathcal{R}$}.
\label{tab:R}]{
\tablestyle{1pt}{1.75}
\begin{tabular}{c:c|ccc}
\multicolumn{1}{c}{scale} & \multicolumn{1}{c|}{\tauhl{$\tau(1,3,5)$} $\times$ $\Rhl{\mathcal{R}^{\star}}$}  & params. & 1-shot & 5-shot \\
\hline
\multirow{5}{*}{\rotatebox{90}{\parbox[c][0.75cm][c]{1.2cm}{multiple}}} & $\Rhl{(1^2,~1^2,~1^2)}$ & +4.4 M & \ssgap{56.08} & \ssgap{73.19} \\
& $\Rhl{(3^2,~3^2,~3^2)}$ & +5.8 M & 56.03 & 73.39\\
& $\Rhl{(5^2,~5^2,~5^2)}$ & +7.7 M &  55.56 & 73.04 \\
& $\Rhl{(1^2,~3^2,~5^2)}$ & +5.8 M & \hlours{56.73} & \hlours{74.14} \\
& $\Rhl{(5^2,~3^2,~1^2)}$ & +5.8 M & 56.01 & 74.13 \\
\end{tabular}}
\hspace{20pt}
\subfloat[\textbf{Temporal and spatial pooling layer $\theta$}. avg.: average pooling, conv.: convolution. \label{tab:theta}]
{
\tablestyle{4pt}{1.3}
\begin{tabular}{l:c:c|cc}
\multicolumn{1}{c}{scale} & \multicolumn{1}{c}{$\theta_S$} & \multicolumn{1}{c|}{$\theta_T$} &  1-shot & 5-shot \\
\hline
\multicolumn{1}{c}{single} & N/A & N/A  & \psp{46.80}  & \psp{64.02} \\
\hdashline
\multirow{5}{*}{\rotatebox{90}{\parbox[c][0.5cm][c]{1.8cm}{multiple}}}
 & \multirow{3}{*}{\rotatebox{90}{\parbox[c][0.5cm][c]{0.65cm}{fixed}}} & avg. & 52.01 & 70.32 \\
&  & max & 54.23 &  71.90 \\
&  & conv. & 54.02 & 72.04 \\
\cdashline{2-5}
 & \multirow{3}{*}{\rotatebox{90}{\parbox[c][0.5cm][c]{1.2cm}{semantic}}} & avg.  & 50.93 & 69.34 \\
&  & max   & 52.67 & 71.44 \\
&  & conv. & \hlours{56.73} & \hlours{74.14} \\
\end{tabular}
}
\hspace{2pt}
\subfloat[\textbf{Adaptive alignment (A$^2$ module):} evaluating A$^2$ module for second-order moment under single- and multi-scale settings. \revision{p-p: point-to-point, cr.: cross.} \label{tab:A2}]
{
\tablestyle{4pt}{1.75}
\begin{tabular}{cc|cc}
scale & A$^2$ & 1-shot & 5-shot \\
\hline
\multirow{2}{*}{\rotatebox{90}{\parbox[c][0.5cm][c]{0.7cm}{\revision{single}}}}  & \revision{$\times$} & 45.49  & 63.33\\
 & \checkmark &  \psp{46.80} & \psp{64.02}  \\
\hdashline
\multirow{3}{*}{\rotatebox{90}{\parbox[c][0.5cm][c]{1.2cm}{\revision{multiple}}}}  & \revision{$\times$,p-p}  & 48.03 & 65.70 \\
 & \revision{$\times$,cr.} & \revision{42.40} & \revision{57.31} \\
  & \checkmark & \hlours{56.73} & \hlours{74.14} \\
\end{tabular}
}
\hspace{3pt}
\subfloat[\textbf{Multi-scale Moment (M$^2$ block)}: component effectiveness of M$^2$ block with using A$^2$ module. M-S: multi-scale, SM: Second-order moment.\label{tab:M2}]
{
\tablestyle{1.5pt}{1.75}
\begin{tabular}{l|cccccc}
\multirow{2}{*}{method} & \multicolumn{2}{c}{M$^2$} & \multirow{2}{*}{A$^2$} & \multirow{2}{*}{1-shot} & \multirow{2}{*}{5-shot} \\
\cdashline{2-3}
 & M-S & SM &  &  &   \\
\hline
baseline & &  & \checkmark & \gap{45.12} & \gap{61.02} \\
Cov-MN &  &  \checkmark  & \checkmark & \psp{46.80} & \psp{64.02} \\
Multi-scale & \checkmark &  & \checkmark & 53.50 & 70.14 \\
\hdashline
A$^2$M$^2$-Net (ours) & \checkmark & \checkmark  & \checkmark & \hlours{56.73} & \hlours{74.14} \\
\end{tabular}
}
\caption{Ablation study of A$^2$M$^2$-Net on SSV2-Full dataset.}
\label{tab:ablations}
\end{table*}

\subsection{Ablation Study}\label{sec:ablation}
The ablation study is conducted on the SSV2-Full dataset, reporting 5-way 1-shot and 5-way 5-shot accuracy. The results are presented in Table~\ref{tab:ablations}, where the performances highlighted with the same color denoting the same settings. For clarification, we refer to the baseline as GAP combined with EMD.

\noindent \textbf{Evaluation on high-order statistics.} We firstly evaluate the effect of feature statistics. In Table~\ref{tab:sop}, we report the first-order approach (GAP), \revision{a third-order method (Kernel Pooling~\cite{KP_Cui_2017_CVPR}, KP)} and various second-order implementations in a vanilla setting (namely, without considering spatio-temporal relation), where Cov denotes the plain covariance pooling, BCNN~\cite{BCNN_lin_2015_ICCV} and Cov-MN~\cite{iSQRT_2018_CVPR} that perform an element-wise signed square-root followed by $\ell_2$ normalization and matrix normalization on Cov, respectively. \revision{Kernel pooling performs a differential approximate higher-order statistics in frequency domain by Fast Fourier Transform (FFT) and Inverse Fast Fourier Transform (IFFT).} It can be observed that, firstly, various second-order approaches  are typically superior to the first-order one, with over \acc{0.75}{0.70} performance gains. \revision{However, beyond second order, the higher-order approach, such as third-order makes no further performance improvements, but suffers from high-dimensional computation consumption ($C_{out}^3$-dimensional)}. Moreover, among various second-order counterparts, the matrix normalized one obtains the best performance. And thus, Cov-MN is instantiated as our second-order approach in follow-up experiments.

\noindent \textbf{Evaluation on multiple scales $\tauhl{\tau}\times\Rhl{\mathcal{R}}$.} To determine the optimal spatio-temporal size for the multi-scale regime, we initially focus on temporal sizes ($\tauhl{\tau}$), while keeping the spatial kernel (\textcolor{colorR}{$\mathcal{R}$}) fixed to \Rhl{$1^2$} for each scale. For instance, the notation $\tauhl{\left(a, b\right)}$ represents a dual-scale configuration (i.e., with $B$=2) with sizes $\tauhl{a}\times$\Rhl{$1^2$} and $\tauhl{b}\times$\Rhl{$1^2$}. Given an input temporal length of $T=8$, we experimented with $\tauhl{\tau}$ values of \tauhl{1, 3, 5}, and \tauhl{7}. Results are reported in Table~\ref{tab:tau}. The results indicate that employing a multi-scale strategy consistently results in substantial performance improvements, surpassing single-scale models by margins exceeding $8.34/8.16\%$. 
Furthermore, transitioning from a two-scale to a three-scale model enhances performance, achieving the optimal performance-consumption trade-off at $\tauhl{(1,3,5)}$. 
However, further appending the temporal kernel \tauhl{7} yields only modest performance gains \acc{-0.04}{+0.36}, but at the cost of a considerable parameter increase, specifically by a factor of $\times 1.8$.
Therefore, $\tauhl{\tau=(1,3,5)}$ is implemented as the default configuration of $\tauhl{\tau}$ for A$^2$M$^2$-Net.

By maintaining $\tauhl{\tau=(1,3,5)}$, we proceed to assess various spatial scales \Rhl{$\mathcal{R}$} with optimal temporal scale $\tauhl{(1,3,5)}$, as detailed in Table~\ref{tab:R}. Due to the irregular size of \Rhl{$\mathcal{R}$}, we evaluate \Rhl{$\mathcal{R}$} by its initial gridded size \Rhl{$\mathcal{R}^{\star}$}. Results of the table reveal that customizing the spatial scales for each temporal scale yields performance improvements, \acc{0.65}{0.95}. 

In conclusion, the optimal multi-scale for A$^2$M$^2$-Net is $\tauhl{(1,3,5)}\times \Rhl{(1^2,3^2,5^2)}$.

\begin{figure*}[!htb]
  \centering
  \subfloat[$N$-way 1-shot]{\label{fig:ways}
    \includegraphics[width=.243\textwidth]{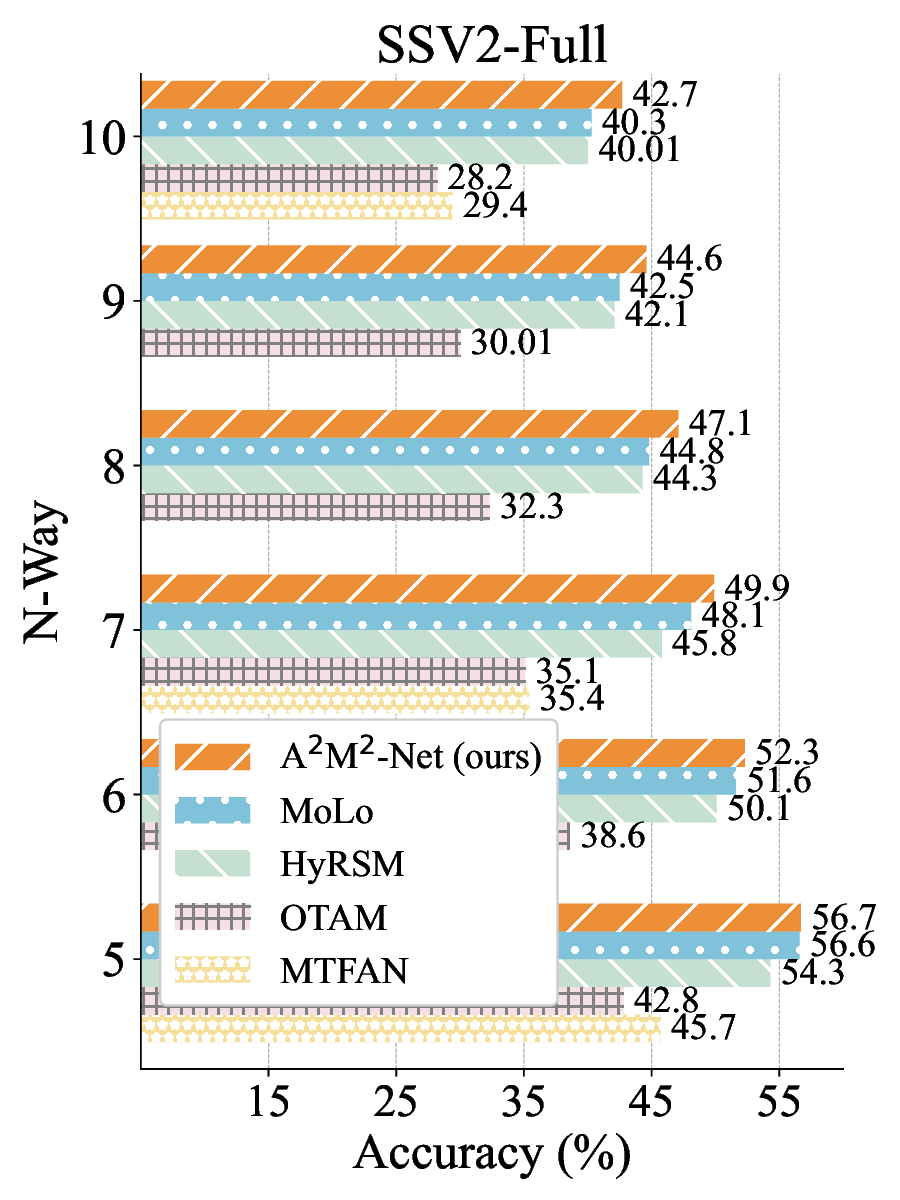}%
    \includegraphics[width=.243\textwidth]{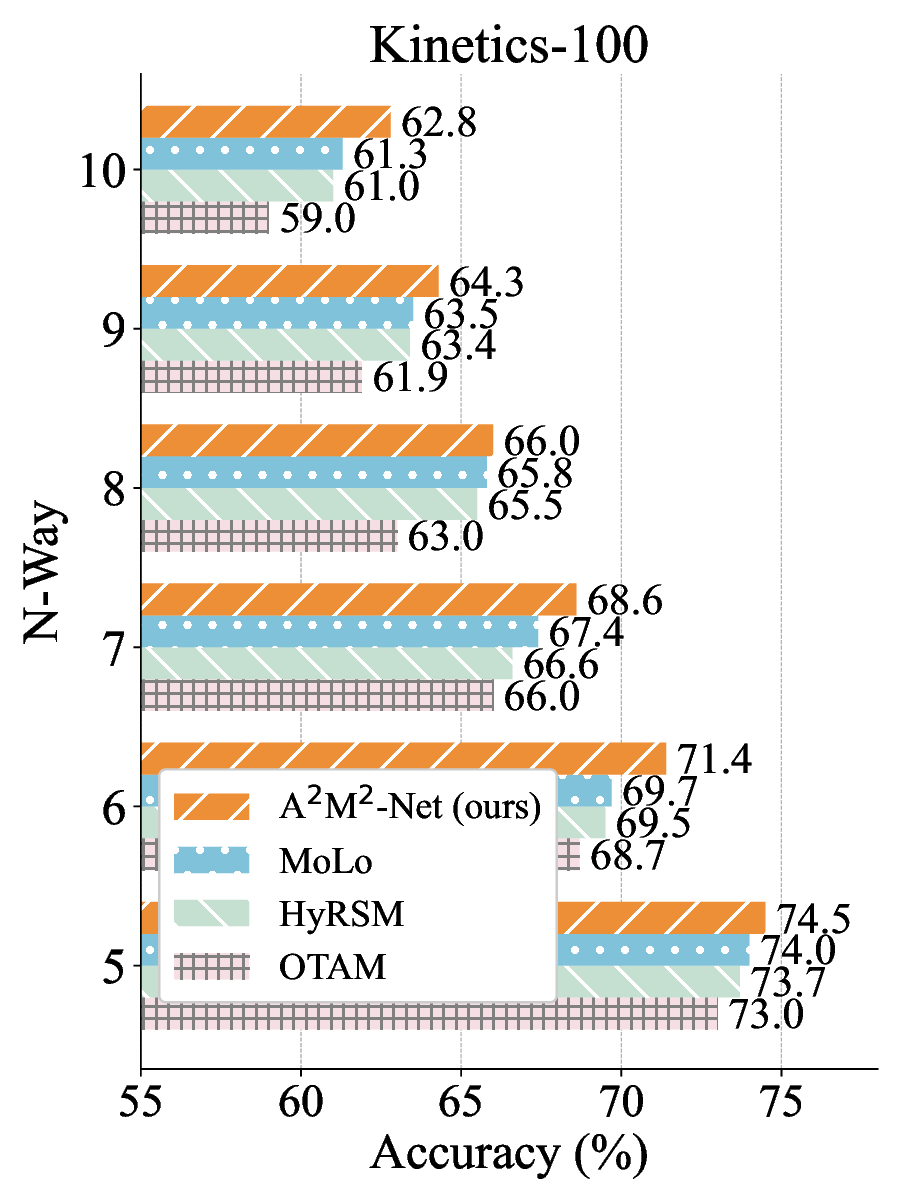}%
  }
  \vspace{0.0cm}
  \subfloat[5-Way $K$-shot ]{\label{fig:shots}
    \includegraphics[width=.243\textwidth]{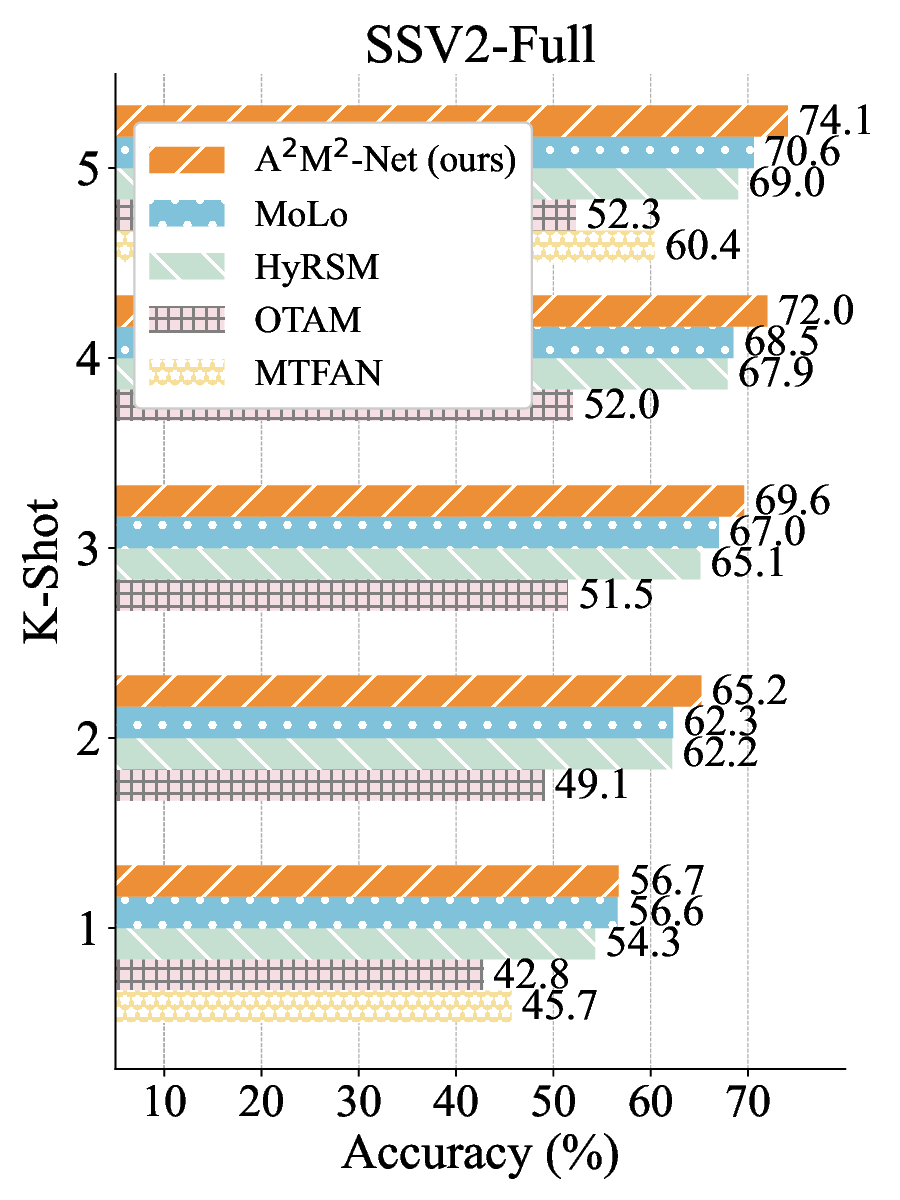}%
    \includegraphics[width=.243\textwidth]{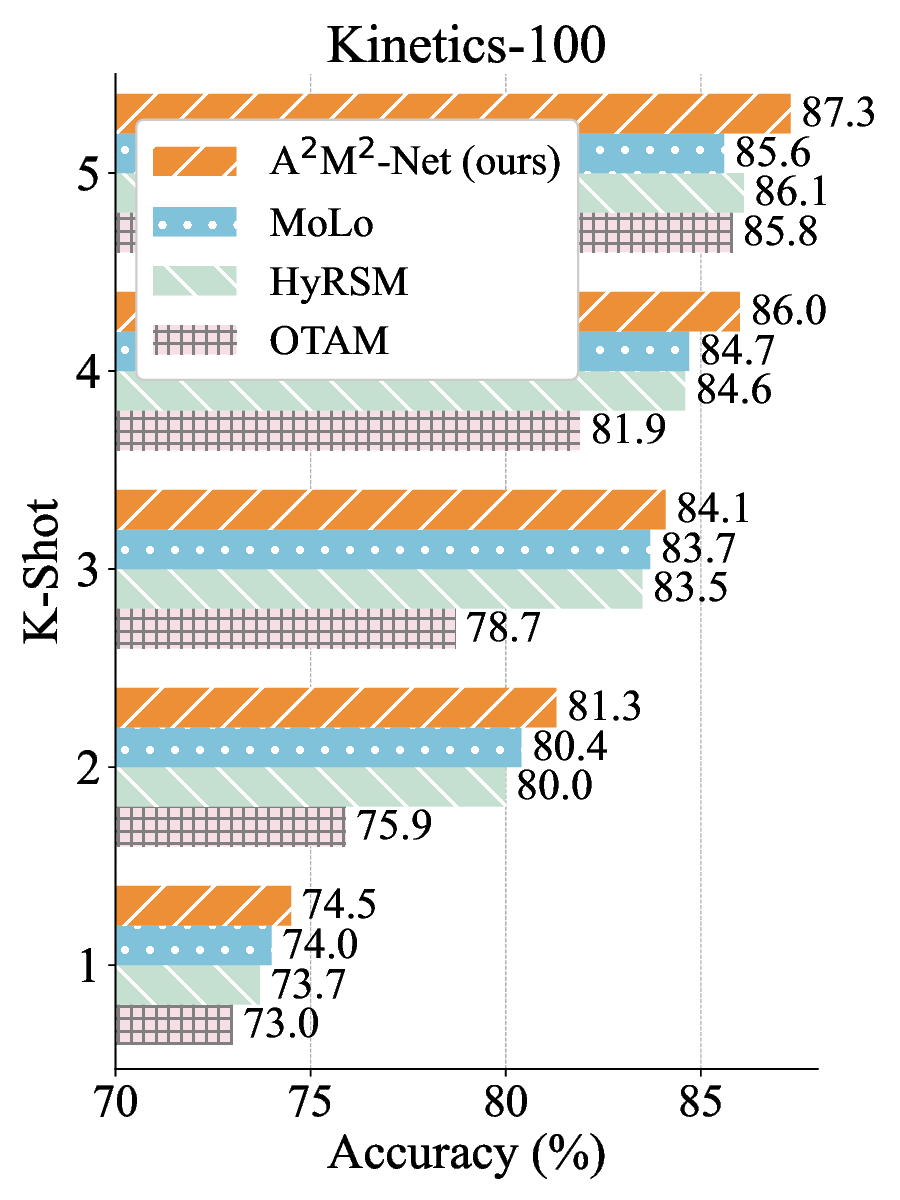}%
  }
  \caption{\revision{Evaluation various ways and shots.}}
  \label{Fig:generalization}
\end{figure*}

\noindent \textbf{Evaluation of $\theta_T$ and $\theta_S$ functions.} We evaluate the effectiveness of spatio-temporal modeling and its implementation function. To be specific, for $\theta_S$, we compare semantic  and fixed (\textit{w.r.t.}, conventional convolution) ones. For $\theta_T$, we assess average (avg.), maximum (max), and convolution (conv.) temporal pooling functions. The results, detailed in Table~\ref{tab:theta}, also include the performance of Cov-MN, which not utilizes $\theta_T$ and $\theta_S$, denoted as N/A. The findings indicate substantial performance enhancements when employing multi-scale protocols with $\theta_T$ or $\theta_S$, leading to the gains over \acc{4.13}{5.32} compared to the single-scale Cov-MN. In particular, for fixed $\theta_S$, both max and convolution layers show equivalent performance, outperforming average pooling by a margin of approximately \acc{2.0}{1.6}. Remarkably, when switching $\theta_S$ to the semantic approach, the configuration of temporal convolution layer demonstrates a clear advantage, outperforming max pooling with a performance of \acc{56.73}{74.14}. Therefore, A$^2$M$^2$-Net employs the optimal  semantic $\theta_S$ together with convolutional $\theta_T$ as default.

\noindent \textbf{Evaluation on adaptive alignment (A$^2$ module).} We evaluate the effectiveness of adaptive alignment scheme (A$^2$ module) in both single-scale and multi-scale settings with second-order moment. Particularly, in the absence of the A$^2$ metric, namely using a fixed alignment scheme,  
\revision{it computes the cosine similarity between query and support features at corresponding timestamps within each scale (point-to-point, p-p) or cross all timestamps among all scales (cr.). Notably, the latter one corresponds to the setting of directly using $\mathcal{SIM}$ for prediction. These similarity scores are then averaged along timestamps and  to derive}

\begin{table}[!htbp]
\centering
\caption{\revision{Efficiency evaluation with 2D approaches. We report the parameter on ResNet50 backbone and meta-test latency for a single task (5-way 1-shot setting) by averaging 10,000 tasks test time, denoted in milliseconds (\textit{ms}).}}
\label{tab:efficiency}
\tablestyle{6pt}{1.5}
\begin{tabular}{lcccc}
\hline
Method  & Parameter & Latency & 1-shot  \\
\hline
OTAM~\cite{OTAM_2020} & 23.6M & 95.3 \textit{ms} & 42.8 \\
TRX~\cite{TRX_2021} & 47.1M & 105.4 \textit{ms} & 42.0 \\
HyRSM~\cite{Hybrid_2022} & 65.6M & 89.2 \textit{ms} & 54.3 \\
A$^2$M$^2$-Net (ours) & 29.9M & 81.4 \textit{ms} &56.7 \\
\gray{~$\sim$ A$^2$ module (ours)} & -- & \gray{11.1 \textit{ms}} & \gray{---}  \\
\hline
\end{tabular}
\end{table}
\noindent \revision{the final similarity score.} Detailed structures of various alignment paradigms are available in the supplemental materials. The results are presented in  Table~\ref{tab:A2}. It can be observed that the adaptive alignment scheme consistently outperforms the fixed alignment in both scenarios. 
Moreover, when expanding from a single scale to multiple scales, the fixed alignment mechanism obtains favorable \revision{performance improvements (\acc{2.54}{2.37}) on p-p setting but resulting no gains on cr. setting}. However, it still significantly lags behind the A$^2$ module fashion with \acc{9.93}{10.12}. 
These findings indicate the generalization of A$^2$ module across both single and multiple scales, and highlight its promising performance when given rich motion descriptors.

\begin{figure}[!htbp]
    \centering
    \includegraphics[width=.49\textwidth]{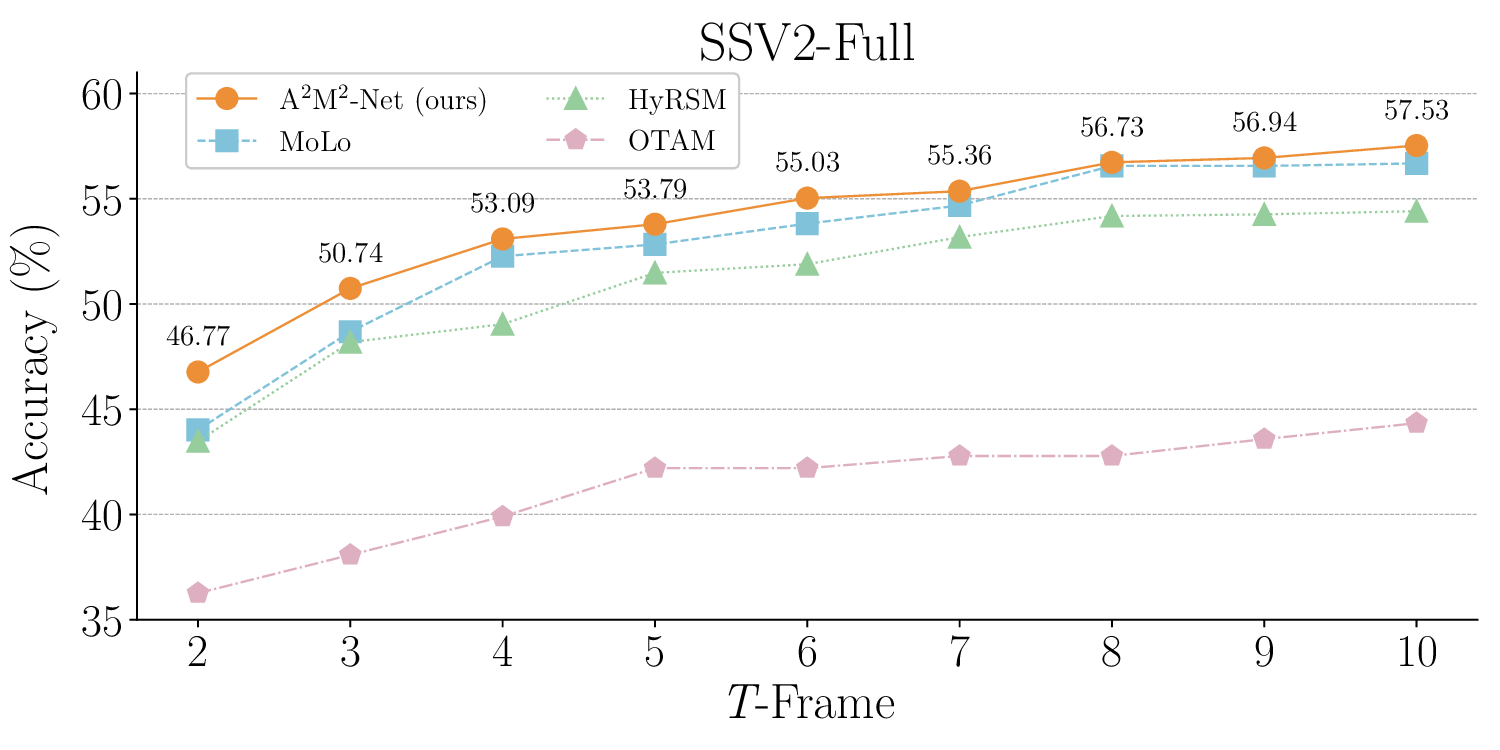}
    \caption{Evaluation on various frames (5-way 1-shot).}
    \label{fig:frames}
\end{figure}

\noindent \textbf{Evaluation on multi-scale moment (M$^2$ block).}
We conduct a components ablation study for M$^2$ block, as detailed in Table~\ref{tab:M2}. Specifically, we assess the impact of incorporating a second-order moment (SM) and a multi-scale (M-S) mechanism into M$^2$ block. The results indicate that the multi-scale strategy and second-order moment individually contribute performance improvements of \acc{1.68}{3.00} and \acc{8.38}{9.12}, respectively. Furthermore, these components are found to be complementary: their combination (M$^2$ block) leads to a performance of \acc{56.73}{74.14}, resulting in overall performance gains of \acc{11.61}{13.12}. Therefore, the devise of multi-scale second-order moment could effectively boost performance.

\noindent \textbf{\revision{Efficiency evaluation.}}
\revision{To evaluate the efficiency of our A$^2$M$^2$-Net in comparison to state-of-the-art approaches \mnrevision{with 2D network}, we measure the inference speed of our model alongside prior methods on a workstation equipped with a single NVIDIA RTX 3090 GPU and an Intel i9-10900X@3.7GHz CPU. We report the latency for a single task meta-test (5-way 1-shot, 1 query video per category). \mnrevision{The results are presented in Table~\ref{tab:efficiency}.}}

\revision{Compared to some state-of-the-art methods, such as OTAM~\cite{OTAM_2020} and TRX~\cite{TRX_2021}, our A$^2$M$^2$-Net maintains a competitive margin in both efficiency and accuracy. To be specific, our model achieves a 15\% speedup over OTAM and a 22\% speedup over TRX with delivering performance gains of 13.9\% and 14.7\% respectively on Accuracy. Additionally, our A$^2$M$^2$-Net outperforms HyRSM~\cite{Hybrid_2022} by 2.4\% accuracy while utilizing only 45\% of the parameter count. Specifically, the Adaptive Alignment (A$^2$) module accounts for approximately 14\% of the total latency. This indicates that the A$^2$ module does not introduce a significant computational burden, ensuring that our model remains both effective and efficient.} \mnrevision{These results demonstrate that our A$^2$M$^2$-Net achieves high computational efficiency while maintaining high performance.}

\subsection{Generalization Evaluation}
To evaluate the generalization of A$^2$M$^2$-Net, we make a further exploration on various shots, ways, frames and combine our M$^2$ block with other metrics. The experiments are performed on the SSV2-Full and K-100 benchmarks.

 \noindent \textbf{Results with various ways and shots.}
We compare the performance of A$^2$M$^2$-Net with state-of-the-arts on more challenging ($N$-way 1-shot) and more practical (5-way $K$-shot) scenarios, including MoLo~\cite{MoLo_cvpr23}, HyRSM~\cite{Hybrid_2022}, MTFAN~\cite{MTFAN_2022} and OTAM~\cite{OTAM_2020}. The results, as shown in Figure~\ref{Fig:generalization}, clearly demonstrate that A$^2$M$^2$-Net consistently outperforms these leading approaches across both various ways and various shots settings. It is notable that MTFAN, a counterpart utilizing the EMD approach, consistently underperforms our A$^2$M$^2$-Net by approximately 10\% across various ways and shots settings on the SSV2-Full dataset (with K-100 results not available). 

Furthermore, although MoLo and HyRSM are formidable competitors, A$^2$M$^2$-Net exhibits surprising performance superiority over them in more shots or ways settings, specifically with an advantage of over 2.4\% (10-way 1-shot) and 3.5\% (5-way 5-shot) on the SSV2-Full dataset.
Besides SSV2-Full database, A$^2$M$^2$-Net still maintains its strong performance on K-100, showing a healthy improvement of 0.2\%$\sim$1.7\% and 0.4\%$\sim$1.3\%  over the runner-up across the scenarios ranging from 5-way to 10-way and 1-shot to 5-shot respectively. These observation suggest that A$^2$M$^2$-Net has the potential to face both highly challenging and realistic few-shot scenarios.

\noindent \textbf{Results with various frames.}
Following pioneering work~\cite{MoLo_cvpr23}, we assess the 5-way 1-shot performance of A$^2$M$^2$-Net and other leading methods across a wide range of input frames ($T$), from 2-frame to 10-frame setups, as depicted in Figure~\ref{fig:frames}. The results reveal that, from 2-frame to 10-frame, our A$^2$M$^2$-Net shows very competitive performance. In particular, in the 2-frame setting, A$^2$M$^2$-Net surpasses other methods by over 2.8\%. This demonstrates that the multi-scale mechanism can effectively explore representative characteristics in very limited settings.

\begin{table}[!htbp]
\centering
\caption{Combining our M$^2$ block with other metrics. $\dag$: our implementation.}
\label{tab:metric}
\tablestyle{3pt}{1.75}
\begin{tabular}{cl|cccccc}
\hline
\makecell {\revision{Flexible} \\ \revision{Alignment}}& Method & 1-shot & 5-shot \\
\hline
\multirow{2}{*}{\revision{$\times$}} & OTAM~\cite{OTAM_2020} & 42.8 & 52.3 \\
& OTAM + M$^2$ (ours) & 50.82 & 59.73 \\
\hdashline
\multirow{2}{*}{\revision{$\checkmark$}} & \revision{Drop-DTW}$\dag$~\cite{Drop_DTW} & \revision{43.8} & \revision{56.8}\\
& \revision{Drop-DTW + M$^2$(ours)} &\revision{51.2} & \revision{69.6}\\
\hdashline
\multirow{2}{*}{\revision{$\times$}}& Bi-MHM~\cite{Hybrid_2022} & 44.6 & 56.0 \\
& Bi-MHM + M$^2$ (ours) & \revision{54.91} & \revision{72.23} \\
\hdashline
\multirow{2}{*}{\revision{$\checkmark$}}& EMD$\dag$ & \gap{45.12} & \gap{61.02} \\
& EMD + M$^2$ (ours) & \hlours{56.73} & \hlours{74.14} \\
\hline
\end{tabular}
\end{table}

\begin{table}[!htbp]
\centering
\caption{\revision{Combining our A$^2$ module with other feature modeling approaches.}}
\label{tab:A2+others}
\tablestyle{12pt}{1.75}
\begin{tabular}{l|ccc}
\hline
 Method & 1-shot & 5-shot \\
\hline
STRM~\cite{STRM_2022} & 43.1 & 68.1 \\
STRM + A$^2$ (ours) & 48.7 & 70.2 \\
\hdashline
HyRSM~\cite{Hybrid_2022} & 54.3 & 69.0 \\
HyRSM + A$^2$ (ours) & 54.6 & 70.9 \\
\hdashline
M$^2$ + A$^2$ (ours) & \hlours{56.73} & \hlours{74.14} \\
\hline
\end{tabular}
\end{table}

\noindent \textbf{Results with various metrics.} We further combine our strong representation output by M$^2$ module with several well-known metrics on SSV2-Full dataset, including OTAM~\cite{OTAM_2020}, \revision{Drop-DTW~\cite{Drop_DTW}}, Bi-MHM~\cite{Hybrid_2022}, and EMD~\cite{EMD_2000}, with results detailed in Table~\ref{tab:metric}. Initially, in the setting without M$^2$ block, EMD demonstrates the highest accuracy among three metrics. Additionally, our M$^2$ module generally enhances the performance across all metrics, yielding improvements of \acc{8.02}{7.43} for OTAM, \revision{\acc{7.4}{12.8} for Drop-DTW}, \revision{\acc{10.31}{16.23}} for Bi-MHM, and \acc{11.61}{13.12} for EMD, respectively. \revision{Notably, as flexible alignment methods, Drop-DTW and EMD achieve remarkable performance improvements when integrated with our M$^2$ block, highlighting the potential of adaptive alignment for FSAR task. Overall, t}hese results confirm that M$^2$ block is a metric-agnostic approach, exhibiting broad generalizability across various metrics, while the adaptive alignment mechanism is the most promising metric for strong representation from M$^2$ block.

\begin{table}[!htb]
\centering
\tablestyle{3pt}{1.75}
\renewcommand\arraystretch{1.75}
\caption{\revision{Results on \mnrevision{large-scale pre-trained models (CLIP and VideoMAE, using ViT-B/16) on SSV2-Full dataset.} The best and second performances are noted by the bold in \textcolor{red}{red} and \underline{underline}, respectively.}}\label{tab:clip}
\begin{tabular}{lcccccc}
\hline
Method &  Pre-train & 1-shot & 3-shot & 5-shot \\
\hline
Shi et al.~\cite{KP} & \multirow{6}{*}{\rotatebox{90}{CLIP-ViT 400M}} &  ---  & ---  & 62.4 \\

CLIP-CPM$^2$C~\cite{CLIPCPM2C} &  & 60.1 & 68.9 & 72.8 \\

CLIP-MDMF~\cite{CLIPM2DF} & & 60.1 & 68.9 & 72.7  \\

CLIP-FSAR~\cite{CLIPFSAR} &  & 61.9 &68.1  & 72.1  \\

MA-CLIP~\cite{MACLIP} &  &  63.3 & --   & 72.3  \\ 

CLIP-A$^2$M$^2$ (ours) & &\sotas{65.9} & \sotas{75.4} & \sotas{82.0}\\
\hdashline
\mnrevision{MAE-A$^2$M$^2$ (ours)} & \mnrevision{VideoMAE} & \sotaf{69.7} & \sotaf{77.6} & \sotaf{82.5} \\
\hline
\end{tabular}
\end{table}

\noindent \textbf{\revision{Results with various feature modeling approaches.}} \revision{We conducted ablation experiments by integrating our A$^2$ module with various feature modeling approaches on the SSV2-Full dataset, specifically with STRM~\cite{STRM_2022} and HyRSM~\cite{Hybrid_2022}. For STRM~\cite{STRM_2022}, the A$^2$ module performs temporal alignment on the query and support tuple collections by setting cardinalities $\Omega=\{2\}$, as suggested in~\cite{Hybrid_2022}. In the case of HyRSM~\cite{Hybrid_2022}, we replaced the Bi-MHM metric with our A$^2$ module to facilitate temporal alignment for its task-specific features. The results presented in Table~\ref{tab:A2+others}, demonstrate that our A$^2$ module achieves significant performance improvements over the original metrics, specifically \acc{5.6}{2.1}  for STRM and \acc{0.3}{1.9} for HyRSM. While both configurations show notable enhancements, they still underperform compared to our A$^2$M$^2$-Net. This demonstrates that our A$^2$ module generalizes well and also effectively complements the M$^2$ block.}

\noindent \textbf{\mnrevision{Results on large-scale pre-trained models.}} \mnrevision{We evaluated the generalization of our A$^2$M$^2$-net on two large-scale pre-trained models on the SSV2-Full~\cite{OTAM_2020} dataset, including CLIP~\cite{CLIP} and VideoMAE~\cite{MAE_2022} with backbone of ViT-B/16. Specifically, our CLIP-A$^2$M$^2$ employs the CLIP visual encoder and inherits its multi-modality setup; detailed architecture and experimental settings are provided in the supplemental material. For our MAE-A$^2$M$^2$, we directly integrate the M$^2$ block and A$^2$ module immediately after the encoder of VideoMAE, using patch tokens without any structural changes. The results are presented in Table~\ref{tab:clip}. Our A$^2$M$^2$-nets obtain \accthree{65.9}{75.4}{82.0} and \accthree{69.5}{77.56}{82.5} by using CLIP and VideoMAE, respectively. Notably, our CLIP-A$^2$M$^2$ exceeds all competing methods using CLIP network, especially in 5-shot accuracy (with over 10\% improvement). These results demonstrates that our A$^2$M$^2$-net can be well generalized to large-scale pre-trained models, and benefit from their strong representation capabilities.}

In summary, our A$^2$M$^2$-Net exhibits impressive superiority to state-of-the-arts across various shots, ways, frames and metrics, \mnrevision{and also benefits from large-scale pre-trained models,} demonstrating strong generalization potential and the ability to handle more challenging or realistic scenarios.

\begin{center}
\begin{figure*}[ht!]
    \centering
    \subfloat[Long jump. \label{fig:vis_ssv2}]{
        \includegraphics[width=0.9\textwidth]{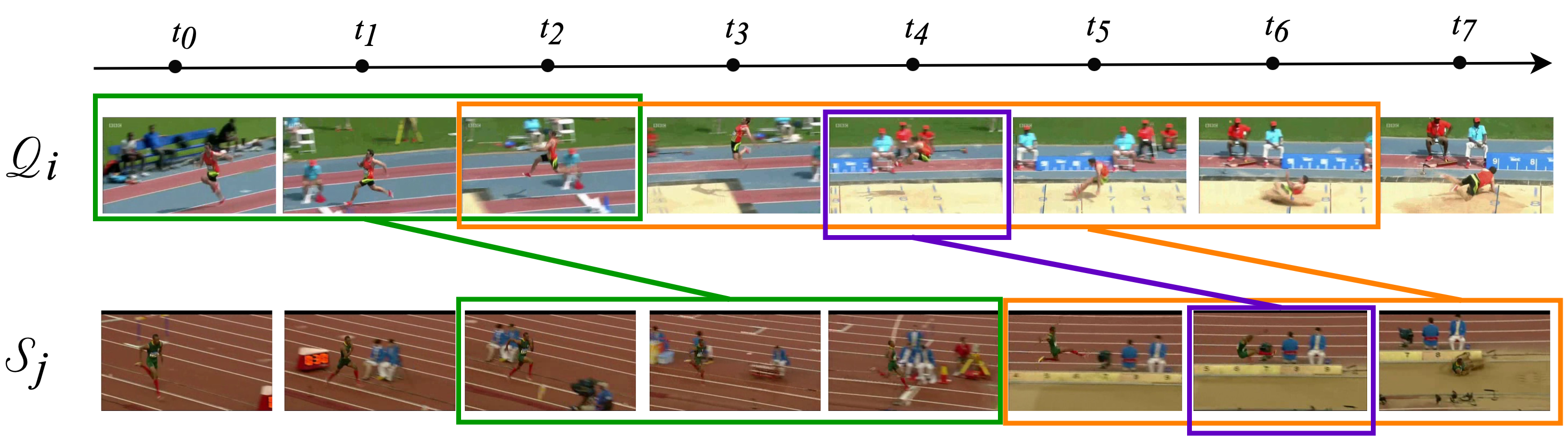}
    }
    
    \subfloat[Breast stroke. \label{fig:vis_ucf}]{
        \includegraphics[width=.9\textwidth]{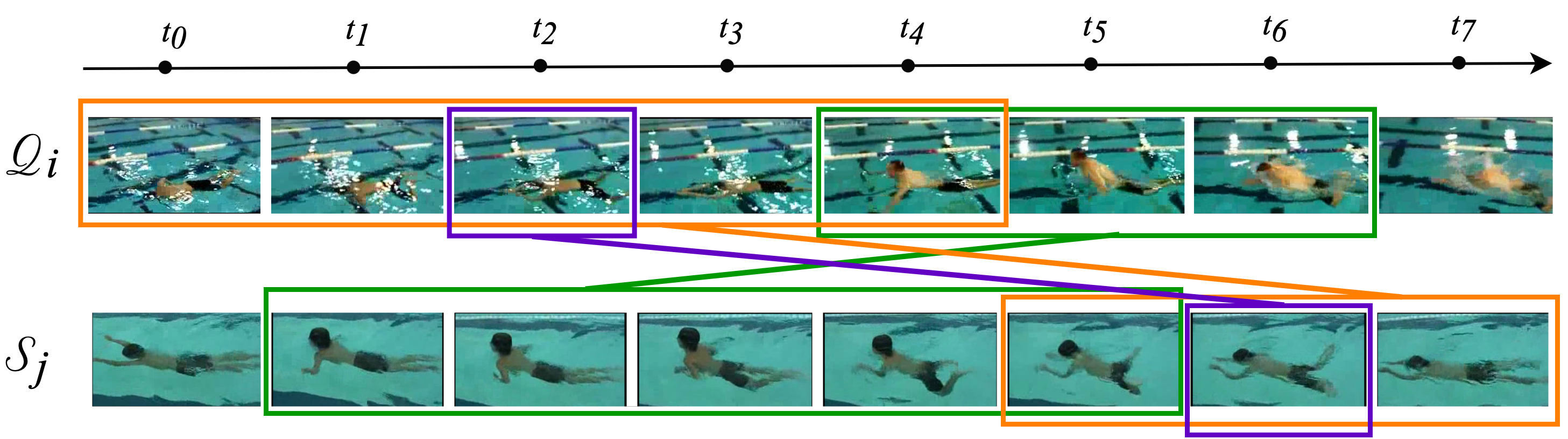}
    }

    \caption{Visualization of A$^2$M$^2$-Net alignment \revision{positive examples}. $\mathcal{Q}_i$: Query video, $\mathcal{S}_{j}$: Positive support video of $\mathcal{Q}_i$. This figure depicts some accurate prediction samples on the UCF-101 dataset, each of which shows three top-3 aligned pairs with highest matching score. 
    The aligned candidates in each pair are indicated by boxes of the same color. Notably, our A$^2$M$^2$-Net successfully address challenges of temporal misalignment, including (1) different subaction timestamps: (a) \textcolor[RGB]{65,146,41}{$\mathcal{Q}_i(t_0\sim t_2)$} \& \textcolor[RGB]{65,146,41}{$\mathcal{S}_j(t_2\sim t_4)$}; (2) varying subaction durations: (b) \textcolor[RGB]{239,135,51}{$\mathcal{Q}_i(t_0\sim t_4)$} \& \textcolor[RGB]{239,135,51}{$\mathcal{S}_j(t_5\sim t_7)$}; 
    (3) inconsistent subaction order (b) [\textcolor[RGB]{239,135,51}{$\mathcal{Q}_i(t_0\sim t_4)$} $\rightarrow$ \textcolor[RGB]{65,146,41}{$\mathcal{Q}_i(t_4\sim t_6)$}] \& [\textcolor[RGB]{239,135,51}{$\mathcal{S}_j(t_5\sim t_7)$} $\rightarrow$ \textcolor[RGB]{65,146,41}{$\mathcal{S}_j(t_1\sim t_5)$}]. 
    In addition, on a short-term scale, the focus is predominantly on key frames, for instance, (b) \textit{flight} in \textit{Long jump} \textcolor[RGB]{89,13,188}{$\mathcal{Q}_i(t_4)$} \& \textcolor[RGB]{89,13,188}{$\mathcal{S}_{j}(t_6)$}. Meanwhile, the longer-term scales often related to the motion revealing completed procedure, as demonstrated by (a) \textcolor[RGB]{239,135,51}{$\mathcal{Q}_i(t_2\sim t_6)$} and (b) \textcolor[RGB]{239,135,51}{$\mathcal{Q}_i(t_0\sim t_4)$}.
}
    \label{fig:visualization}
    \end{figure*}

\end{center}

\subsection{Visualization}\label{sec:vis}
To visualize the temporal alignment results of A$^2$M$^2$-Net, we display the alignment pairs across various scales for selected positive instances and \revision{negative alignment case. For clarity, only the top-3 pairs with the highest alignment scores are highlighted, using boxes of the same color.}

\begin{center}
\begin{figure*}[ht!]
    \centering
        \subfloat[$\mathcal{Q}_i:$ \textit{Pouring sth into sth until it overflows} $\sim\mathcal{S}_{neg}$:\textit{Pouring sth into sth} \label{fig:vis_neg}]{
        \includegraphics[width=.9\textwidth]{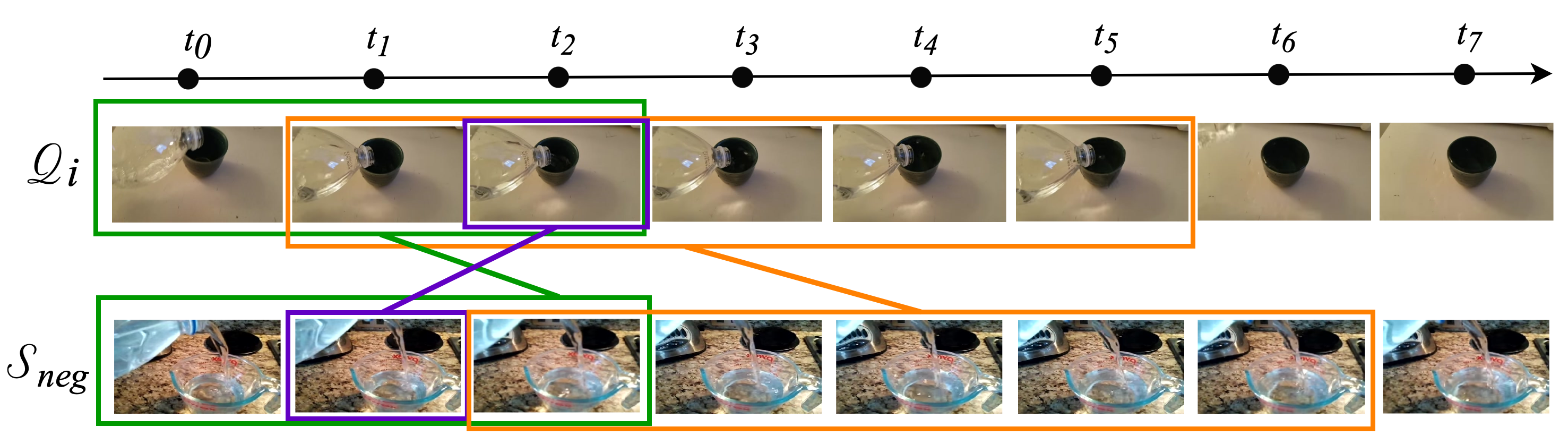}
    }

    \subfloat[$\mathcal{Q}_i~\&~\mathcal{S}_{pos}:$\textit{Pouring sth into sth until it overflows} \label{fig:vis_pos}]{
        \includegraphics[width=.9\textwidth]{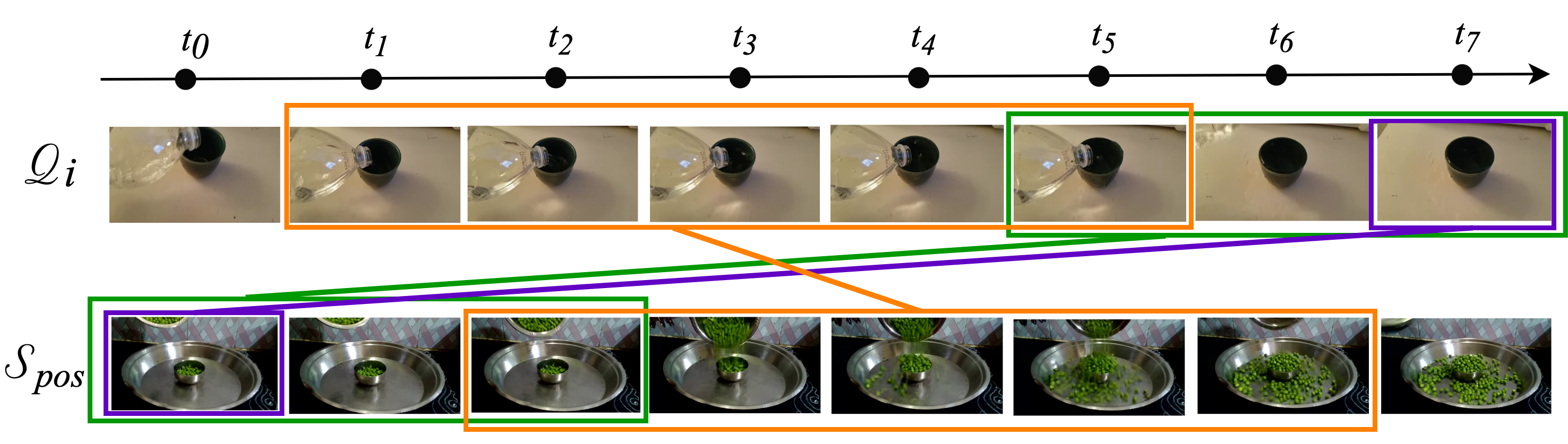}
    }

    \caption{\revision{Failure examples of A$^2$M$^2$-Net alignment results. $\mathcal{Q}_i$: Query video, $\mathcal{S}_{pos}$: Positive support video, $\mathcal{S}_{neg}$: Negative support video. This figure depicts failure prediction sample on the SSV2-Full dataset, each of which shows three top-3 aligned pairs with highest matching score. In this instance, the query video $\mathcal{Q}_i$ was incorrectly predicted to match with $\mathcal{S}_{neg}$. The failure alignment is mainly due to the high similarity between $\mathcal{Q}_i$ and $\mathcal{S}_{neg}$ as well as the subaction (\textit{``overflowing"}) in $\mathcal{Q}_i$ not being prominently visible. }
}
    \label{fig:vis_failure}
    \end{figure*}

\end{center}

\noindent \textbf{\revision{Positive examples.}}
\revision{For positive instances} (denoted as $\mathcal{Q}_i$ \& $\mathcal{S}_{j}$) in the UCF-101 dataset, referring to the activities \textit{long jump} and \textit{breast stroke}, as shown in Figure~\ref{fig:visualization}. Notably, our A$^2$M$^2$-Net successfully address the temporal misalignment challenges including (1) different subaction timestamps: (a) \textcolor[RGB]{65,146,41}{$\mathcal{Q}_i(t_0\sim t_2)$}\footnote{For convenience, here we abbreviate $t_0+i$ to $t_i$.} 
(start from $t_0$ frame) \& \textcolor[RGB]{65,146,41}{$\mathcal{S}_j(t_2\sim t_4)$} (start from $t_2$ frame); (2) varying subaction duration: (b) \textcolor[RGB]{239,135,51}{$\mathcal{Q}_i(t_0\sim t_4)$} (with 5-frame duration) \& \textcolor[RGB]{239,135,51}{$\mathcal{S}_j(t_5\sim t_7)$} (with 3-frame duration); (3) inconsistent subaction order: (b) \textcolor[RGB]{239,135,51}{$\mathcal{Q}_i(t_0\sim t_4)$} $\rightarrow$ \textcolor[RGB]{65,146,41}{$\mathcal{Q}_i(t_4\sim t_6)$} (\textcolor[RGB]{239,135,51}{\textit{stroking} \& \textit{inhaling}} $\rightarrow$ \textcolor[RGB]{65,146,41}{\textit{leg kicking}})~\textit{vs.}~\textcolor[RGB]{65,146,41}{$\mathcal{S}_j(t_1\sim t_5)$} $\rightarrow$ \textcolor[RGB]{239,135,51}{$\mathcal{S}_j(t_5\sim t_7)$} (\textcolor[RGB]{65,146,41}{\textit{leg kicking}} $\rightarrow$ \textcolor[RGB]{239,135,51}{\textit{stroking} \& \textit{inhaling}}).

In addition, the motion characteristics observed at different scales are complementary. On a short-term scale, the focus is predominantly on key frames, such as the \textit{\textcolor[RGB]{89,13,188}{flight}} moments presented in (a) \textcolor[RGB]{89,13,188}{$\mathcal{Q}_i(t_4)$} and \textcolor[RGB]{89,13,188}{$\mathcal{S}_{j}(t_6)$}. Conversely, longer-term scales are more oriented towards revealing the motion process, exemplified by (a) \textcolor[RGB]{239,135,51}{$\mathcal{Q}_i(t_2\sim t_6)$} and (b) \textcolor[RGB]{239,135,51}{$\mathcal{Q}_i(t_0\sim t_4)$}.

In summary, our proposed adaptive alignment mechanism effectively resolves temporal misalignment challenges by enabling flexible selection of strong representation candidates across multiple scales, which is consistent with the qualitative results in Table~\ref{tab:sota_overview}.

\noindent \textbf{Failure examples.}
\revision{To comprehensively showcase the alignment results of our A$^2$M$^2$-Net, we visualize a failure example on SSV2-Full dataset, as shown in Figure~\ref{fig:vis_failure} \mnrevision{(more examples could be found in supplemental matertial)}. Specifically, the query video $\mathcal{Q}_i$ and its positive support video $\mathcal{S}_{pos}$ belong to the category \textit{“Pouring something into something until it overflows”}, while the negative support video $\mathcal{S}_{neg}$ is labeled as \textit{“Pouring something into something”}. In this instance, the query video $\mathcal{Q}_i$ was incorrectly predicted to match with $\mathcal{S}_{neg}$.}

\revision{From the figure, we observe that $\mathcal{Q}_i$ and $\mathcal{S}_{neg}$ share very similar appearance features and overlapping subactions. Although $\mathcal{Q}_i$ and $\mathcal{S}_{pos}$ belong to the same category, they exhibit significant differences in their motion subjects (transparent pure water \textit{vs}. green beans). Additionally, the motion subject in $\mathcal{Q}_i$ is transparent, making the critical subaction \textit{“overflowing”} subtle and not prominently visible. These factors contribute to the high similarity between $\mathcal{Q}_i$ and $\mathcal{S}_{neg}$, leading to misalignment and erroneous prediction.}

\revision{This failure case highlights that high appearance similarity in negative pairs can lead to misalignment, which can be further explored in future work.}

\section{Conclusion}

In this paper, we proposed an adaptively aligned multi-scale moment
(namely A$^2$M$^2$-Net) to handle the temporal misalignment challenge, aiming to improve performance of few-shot action recognition (FSAR). To
this end, two key components (i.e., A$^2$ alignment approach and M$^2$ representation module) are collaborated to selectively utilize the multi-scale second-order moment representations candidates for temporal alignment in an adaptive manner. 
Due to support aligning multi-scale representation cross scales, A$^2$M$^2$-Net is friendly to diverse subaction durations and varying subaction orders and thus robust to the temporal misalignment challenge.
The extensive experiments on five popular FSAR datasets clearly demonstrate the effectiveness of our proposed method.

\backmatter

\bmhead{Acknowledgements}

This work was supported in part by the National
Natural Science Foundation of China under Grant 62471083, Grant
61971086, Grant 62276186, and Grant 61925602; in part by the Haihe
Lab of ITAI under Grant 22HHXCJC00002; and in part by the Science
and Technology Development Program Project of Jilin Province under
Grant 20230201111GX.

\bmhead{Data Availability}

All experiments are conducted on publicly available datasets. To be specific, SSV2 dataset can be found at \url{https://developer.qualcomm.com/software/ai-datasets/something-something}, Kinetics dataset is available at \url{http://deepmind.com/kinetics}, HMDB-51 dataset is at \url{https://serre-lab.clps.brown.edu/resource/hmdb-a-large-human-motion-database}, and UCF-101 dataset is accessible at URL \url{https://www.crcv.ucf.edu/data/UCF101.php}.



\begin{thebibliography}{89}
\ifx \bisbn   \undefined \def \bisbn  #1{ISBN #1}\fi
\ifx \binits  \undefined \def \binits#1{#1}\fi
\ifx \bauthor  \undefined \def \bauthor#1{#1}\fi
\ifx \batitle  \undefined \def \batitle#1{#1}\fi
\ifx \bjtitle  \undefined \def \bjtitle#1{#1}\fi
\ifx \bvolume  \undefined \def \bvolume#1{\textbf{#1}}\fi
\ifx \byear  \undefined \def \byear#1{#1}\fi
\ifx \bissue  \undefined \def \bissue#1{#1}\fi
\ifx \bfpage  \undefined \def \bfpage#1{#1}\fi
\ifx \blpage  \undefined \def \blpage #1{#1}\fi
\ifx \burl  \undefined \def \burl#1{\textsf{#1}}\fi
\ifx \doiurl  \undefined \def \doiurl#1{\url{https://doi.org/#1}}\fi
\ifx \betal  \undefined \def \betal{\textit{et al.}}\fi
\ifx \binstitute  \undefined \def \binstitute#1{#1}\fi
\ifx \binstitutionaled  \undefined \def \binstitutionaled#1{#1}\fi
\ifx \bctitle  \undefined \def \bctitle#1{#1}\fi
\ifx \beditor  \undefined \def \beditor#1{#1}\fi
\ifx \bpublisher  \undefined \def \bpublisher#1{#1}\fi
\ifx \bbtitle  \undefined \def \bbtitle#1{#1}\fi
\ifx \bedition  \undefined \def \bedition#1{#1}\fi
\ifx \bseriesno  \undefined \def \bseriesno#1{#1}\fi
\ifx \blocation  \undefined \def \blocation#1{#1}\fi
\ifx \bsertitle  \undefined \def \bsertitle#1{#1}\fi
\ifx \bsnm \undefined \def \bsnm#1{#1}\fi
\ifx \bsuffix \undefined \def \bsuffix#1{#1}\fi
\ifx \bparticle \undefined \def \bparticle#1{#1}\fi
\ifx \barticle \undefined \def \barticle#1{#1}\fi
\bibcommenthead
\ifx \bconfdate \undefined \def \bconfdate #1{#1}\fi
\ifx \botherref \undefined \def \botherref #1{#1}\fi
\ifx \url \undefined \def \url#1{\textsf{#1}}\fi
\ifx \bchapter \undefined \def \bchapter#1{#1}\fi
\ifx \bbook \undefined \def \bbook#1{#1}\fi
\ifx \bcomment \undefined \def \bcomment#1{#1}\fi
\ifx \oauthor \undefined \def \oauthor#1{#1}\fi
\ifx \citeauthoryear \undefined \def \citeauthoryear#1{#1}\fi
\ifx \endbibitem  \undefined \def \endbibitem {}\fi
\ifx \bconflocation  \undefined \def \bconflocation#1{#1}\fi
\ifx \arxivurl  \undefined \def \arxivurl#1{\textsf{#1}}\fi
\csname PreBibitemsHook\endcsname

\bibitem[\protect\citeauthoryear{Cao et~al.}{2020}]{OTAM_2020}
\begin{bchapter}
\bauthor{\bsnm{Cao}, \binits{K.}},
\bauthor{\bsnm{Ji}, \binits{J.}},
\bauthor{\bsnm{Cao}, \binits{Z.}},
\bauthor{\bsnm{Chang}, \binits{C.-Y.}},
\bauthor{\bsnm{Niebles}, \binits{J.C.}}:
\bctitle{Few-shot video classification via temporal alignment}.
In: \bbtitle{CVPR},
pp. \bfpage{10618}--\blpage{10627}
(\byear{2020})
\end{bchapter}
\endbibitem

\bibitem[\protect\citeauthoryear{Wang et~al.}{2022}]{Hybrid_2022}
\begin{bchapter}
\bauthor{\bsnm{Wang}, \binits{X.}},
\bauthor{\bsnm{Zhang}, \binits{S.}},
\bauthor{\bsnm{Qing}, \binits{Z.}},
\bauthor{\bsnm{Tang}, \binits{M.}},
\bauthor{\bsnm{Zuo}, \binits{Z.}},
\bauthor{\bsnm{Gao}, \binits{C.}},
\bauthor{\bsnm{Jin}, \binits{R.}},
\bauthor{\bsnm{Sang}, \binits{N.}}:
\bctitle{Hybrid relation guided set matching for few-shot action recognition}.
In: \bbtitle{CVPR},
pp. \bfpage{19948}--\blpage{19957}
(\byear{2022})
\end{bchapter}
\endbibitem

\bibitem[\protect\citeauthoryear{Wu et~al.}{2022}]{MTFAN_2022}
\begin{bchapter}
\bauthor{\bsnm{Wu}, \binits{J.}},
\bauthor{\bsnm{Zhang}, \binits{T.}},
\bauthor{\bsnm{Zhang}, \binits{Z.}},
\bauthor{\bsnm{Wu}, \binits{F.}},
\bauthor{\bsnm{Zhang}, \binits{Y.}}:
\bctitle{Motion-modulated temporal fragment alignment network for few-shot
  action recognition}.
In: \bbtitle{CVPR},
pp. \bfpage{9151}--\blpage{9160}
(\byear{2022})
\end{bchapter}
\endbibitem

\bibitem[\protect\citeauthoryear{Lu et~al.}{2021}]{CMOT}
\begin{botherref}
\oauthor{\bsnm{Lu}, \binits{S.}},
\oauthor{\bsnm{Ye}, \binits{H.-J.}},
\oauthor{\bsnm{Zhan}, \binits{D.-C.}}:
Few-shot action recognition with compromised metric via optimal transport.
arXiv preprint arXiv:2104.03737
(2021)
\end{botherref}
\endbibitem

\bibitem[\protect\citeauthoryear{Zhang et~al.}{2021}]{ITA_2021}
\begin{bchapter}
\bauthor{\bsnm{Zhang}, \binits{S.}},
\bauthor{\bsnm{Zhou}, \binits{J.}},
\bauthor{\bsnm{He}, \binits{X.}}:
\bctitle{Learning implicit temporal alignment for few-shot video
  classification}.
In: \bbtitle{IJCAI}
(\byear{2021})
\end{bchapter}
\endbibitem

\bibitem[\protect\citeauthoryear{Wanyan et~al.}{2023}]{AMFAR_cvpr23}
\begin{bchapter}
\bauthor{\bsnm{Wanyan}, \binits{Y.}},
\bauthor{\bsnm{Yang}, \binits{X.}},
\bauthor{\bsnm{Chen}, \binits{C.}},
\bauthor{\bsnm{Xu}, \binits{C.}}:
\bctitle{Active exploration of multimodal complementarity for few-shot action
  recognition}.
In: \bbtitle{CVPR},
pp. \bfpage{6492}--\blpage{6502}
(\byear{2023})
\end{bchapter}
\endbibitem

\bibitem[\protect\citeauthoryear{Cao et~al.}{2021}]{RVN_CVIU}
\begin{barticle}
\bauthor{\bsnm{Cao}, \binits{C.}},
\bauthor{\bsnm{Li}, \binits{Y.}},
\bauthor{\bsnm{Lv}, \binits{Q.}},
\bauthor{\bsnm{Wang}, \binits{P.}},
\bauthor{\bsnm{Zhang}, \binits{Y.}}:
\batitle{Few-shot action recognition with implicit temporal alignment and pair
  similarity optimization}.
\bjtitle{Comput. Vis. Image Underst.}
\bvolume{210},
\bfpage{103250}
(\byear{2021})
\end{barticle}
\endbibitem

\bibitem[\protect\citeauthoryear{He and Gao}{2021}]{TBSN}
\begin{bchapter}
\bauthor{\bsnm{He}, \binits{J.}},
\bauthor{\bsnm{Gao}, \binits{S.}}:
\bctitle{{TBSN}: Sparse-transformer based siamese network for few-shot action
  recognition}.
In: \bbtitle{ICTC},
pp. \bfpage{47}--\blpage{53}
(\byear{2021})
\end{bchapter}
\endbibitem

\bibitem[\protect\citeauthoryear{Wang et~al.}{2023}]{MoLo_cvpr23}
\begin{bchapter}
\bauthor{\bsnm{Wang}, \binits{X.}},
\bauthor{\bsnm{Zhang}, \binits{S.}},
\bauthor{\bsnm{Qing}, \binits{Z.}},
\bauthor{\bsnm{Gao}, \binits{C.}},
\bauthor{\bsnm{Zhang}, \binits{Y.}},
\bauthor{\bsnm{Zhao}, \binits{D.}},
\bauthor{\bsnm{Sang}, \binits{N.}}:
\bctitle{{Mo{L}o}: Motion-augmented long-short contrastive learning for
  few-shot action recognition}.
In: \bbtitle{CVPR},
pp. \bfpage{18011}--\blpage{18021}
(\byear{2023})
\end{bchapter}
\endbibitem

\bibitem[\protect\citeauthoryear{Bishay et~al.}{2019}]{TARN_bishay2019tarn}
\begin{bchapter}
\bauthor{\bsnm{Bishay}, \binits{M.}},
\bauthor{\bsnm{Zoumpourlis}, \binits{G.}},
\bauthor{\bsnm{Patras}, \binits{I.}}:
\bctitle{{TARN:} {T}emporal attentive relation network for few-shot and
  zero-shot action recognition}.
In: \bbtitle{BMVC}
(\byear{2019})
\end{bchapter}
\endbibitem

\bibitem[\protect\citeauthoryear{Zhang et~al.}{2020}]{ZHANG_PR}
\begin{barticle}
\bauthor{\bsnm{Zhang}, \binits{L.}},
\bauthor{\bsnm{Chang}, \binits{X.}},
\bauthor{\bsnm{Liu}, \binits{J.}},
\bauthor{\bsnm{Luo}, \binits{M.}},
\bauthor{\bsnm{Prakash}, \binits{M.}},
\bauthor{\bsnm{Hauptmann}, \binits{A.G.}}:
\batitle{Few-shot activity recognition with cross-modal memory network}.
\bjtitle{Pattern Recognition}
\bvolume{108},
\bfpage{107348}
(\byear{2020})
\end{barticle}
\endbibitem

\bibitem[\protect\citeauthoryear{Perrett et~al.}{2021}]{TRX_2021}
\begin{bchapter}
\bauthor{\bsnm{Perrett}, \binits{T.}},
\bauthor{\bsnm{Masullo}, \binits{A.}},
\bauthor{\bsnm{Burghardt}, \binits{T.}},
\bauthor{\bsnm{Mirmehdi}, \binits{M.}},
\bauthor{\bsnm{Damen}, \binits{D.}}:
\bctitle{Temporal-relational {C}ross{T}ransformers for few-shot action
  recognition}.
In: \bbtitle{CVPR},
pp. \bfpage{475}--\blpage{484}
(\byear{2021})
\end{bchapter}
\endbibitem

\bibitem[\protect\citeauthoryear{Yu et~al.}{2023}]{MGCSM_MM23}
\begin{bchapter}
\bauthor{\bsnm{Yu}, \binits{T.}},
\bauthor{\bsnm{Chen}, \binits{P.}},
\bauthor{\bsnm{Dang}, \binits{Y.}},
\bauthor{\bsnm{Huan}, \binits{R.}},
\bauthor{\bsnm{Liang}, \binits{R.}}:
\bctitle{Multi-speed global contextual subspace matching for few-shot action
  recognition}.
In: \bbtitle{ACM Multimedia},
pp. \bfpage{2344}--\blpage{2352}
(\byear{2023})
\end{bchapter}
\endbibitem

\bibitem[\protect\citeauthoryear{Wang et~al.}{2020}]{PAMI20}
\begin{barticle}
\bauthor{\bsnm{Wang}, \binits{Q.}},
\bauthor{\bsnm{Xie}, \binits{J.}},
\bauthor{\bsnm{Zuo}, \binits{W.}},
\bauthor{\bsnm{Zhang}, \binits{L.}},
\bauthor{\bsnm{Li}, \binits{P.}}:
\batitle{Deep {CNN}s meet global covariance pooling: Better representation and
  generalization}.
\bjtitle{{IEEE} Trans. on Pattern Analysis and Machine Intelligence}
\bvolume{43}(\bissue{8}),
\bfpage{2582}--\blpage{2597}
(\byear{2020})
\end{barticle}
\endbibitem

\bibitem[\protect\citeauthoryear{Rubner et~al.}{2000}]{EMD_2000}
\begin{barticle}
\bauthor{\bsnm{Rubner}, \binits{Y.}},
\bauthor{\bsnm{Tomasi}, \binits{C.}},
\bauthor{\bsnm{Guibas}, \binits{L.J.}}:
\batitle{The earth mover's distance as a metric for image retrieval}.
\bjtitle{Int. J. Comput. Vision}
\bvolume{40}(\bissue{2}),
\bfpage{99}--\blpage{121}
(\byear{2000})
\end{barticle}
\endbibitem

\bibitem[\protect\citeauthoryear{Zhu and Yang}{2018}]{CMN_zhu2018}
\begin{bchapter}
\bauthor{\bsnm{Zhu}, \binits{L.}},
\bauthor{\bsnm{Yang}, \binits{Y.}}:
\bctitle{Compound memory networks for few-shot video classification}.
In: \bbtitle{ECCV},
pp. \bfpage{751}--\blpage{766}
(\byear{2018})
\end{bchapter}
\endbibitem

\bibitem[\protect\citeauthoryear{Soomro et~al.}{2012}]{UCF101_arxiv}
\begin{botherref}
\oauthor{\bsnm{Soomro}, \binits{K.}},
\oauthor{\bsnm{Zamir}, \binits{A.R.}},
\oauthor{\bsnm{Shah}, \binits{M.}}:
{UCF101}: A dataset of 101 human actions classes from videos in the wild.
arXiv preprint arXiv:1212.0402
(2012)
\end{botherref}
\endbibitem

\bibitem[\protect\citeauthoryear{Hilde et~al.}{2011}]{HMDB_2011_ICCV}
\begin{bchapter}
\bauthor{\bsnm{Hilde}, \binits{K.}},
\bauthor{\bsnm{Hueihan}, \binits{J.}},
\bauthor{\bsnm{Stiefel}, \binits{h.}},
\bauthor{\bsnm{Thomas}, \binits{S.}}:
\bctitle{{HMDB}: A large video database for human motion recognition}.
In: \bbtitle{ICCV},
pp. \bfpage{2556}--\blpage{2563}
(\byear{2011})
\end{bchapter}
\endbibitem

\bibitem[\protect\citeauthoryear{Snell et~al.}{2017}]{proto_2017}
\begin{bchapter}
\bauthor{\bsnm{Snell}, \binits{J.}},
\bauthor{\bsnm{Swersky}, \binits{K.}},
\bauthor{\bsnm{Zemel}, \binits{R.S.}}:
\bctitle{Prototypical networks for few-shot learning}.
In: \bbtitle{NIPS},
vol. \bseriesno{30},
pp. \bfpage{4080}--\blpage{4090}
(\byear{2017})
\end{bchapter}
\endbibitem

\bibitem[\protect\citeauthoryear{Sung et~al.}{2018}]{relationnet}
\begin{bchapter}
\bauthor{\bsnm{Sung}, \binits{F.}},
\bauthor{\bsnm{Yang}, \binits{Y.}},
\bauthor{\bsnm{Zhang}, \binits{L.}},
\bauthor{\bsnm{Xiang}, \binits{T.}},
\bauthor{\bsnm{Torr}, \binits{P.H.}},
\bauthor{\bsnm{Hospedales}, \binits{T.M.}}:
\bctitle{Learning to compare: Relation network for few-shot learning}.
In: \bbtitle{CVPR},
pp. \bfpage{1199}--\blpage{1208}
(\byear{2018})
\end{bchapter}
\endbibitem

\bibitem[\protect\citeauthoryear{Lifchitz et~al.}{2019}]{DC}
\begin{bchapter}
\bauthor{\bsnm{Lifchitz}, \binits{Y.}},
\bauthor{\bsnm{Avrithis}, \binits{Y.}},
\bauthor{\bsnm{Picard}, \binits{S.}},
\bauthor{\bsnm{Bursuc}, \binits{A.}}:
\bctitle{Dense classification and implanting for few-shot learning}.
In: \bbtitle{CVPR},
pp. \bfpage{9258}--\blpage{9267}
(\byear{2019})
\end{bchapter}
\endbibitem

\bibitem[\protect\citeauthoryear{Hao et~al.}{2019}]{SAML}
\begin{bchapter}
\bauthor{\bsnm{Hao}, \binits{F.}},
\bauthor{\bsnm{He}, \binits{F.}},
\bauthor{\bsnm{Cheng}, \binits{J.}},
\bauthor{\bsnm{Wang}, \binits{L.}},
\bauthor{\bsnm{Cao}, \binits{J.}},
\bauthor{\bsnm{Tao}, \binits{D.}}:
\bctitle{Collect and select: {S}emantic alignment metric learning for few-shot
  learning}.
In: \bbtitle{ICCV},
pp. \bfpage{8460}--\blpage{8469}
(\byear{2019})
\end{bchapter}
\endbibitem

\bibitem[\protect\citeauthoryear{Xie et~al.}{2022}]{DeepBDC_22}
\begin{botherref}
\oauthor{\bsnm{Xie}, \binits{J.}},
\oauthor{\bsnm{Long}, \binits{F.}},
\oauthor{\bsnm{Lv}, \binits{J.}},
\oauthor{\bsnm{Wang}, \binits{Q.}},
\oauthor{\bsnm{Li}, \binits{P.}}:
Joint distribution matters: Deep brownian distance covariance for few-shot
  classification.
CVPR,
7962--7971
(2022)
\end{botherref}
\endbibitem

\bibitem[\protect\citeauthoryear{Zhang et~al.}{2020}]{DeepEMD_2020}
\begin{bchapter}
\bauthor{\bsnm{Zhang}, \binits{C.}},
\bauthor{\bsnm{Cai}, \binits{Y.}},
\bauthor{\bsnm{Lin}, \binits{G.}},
\bauthor{\bsnm{Shen}, \binits{C.}}:
\bctitle{Deep{EMD}: Few-shot image classification with differentiable earth
  mover’s distance and structured classifiers}.
In: \bbtitle{CVPR},
pp. \bfpage{12200}--\blpage{12210}
(\byear{2020})
\end{bchapter}
\endbibitem

\bibitem[\protect\citeauthoryear{Zhang et~al.}{2022}]{DeepEMD_PAMI}
\begin{botherref}
\oauthor{\bsnm{Zhang}, \binits{C.}},
\oauthor{\bsnm{Cai}, \binits{Y.}},
\oauthor{\bsnm{Lin}, \binits{G.}},
\oauthor{\bsnm{Shen}, \binits{C.}}:
Deep{EMD}: Differentiable earth mover's distance for few-shot learning.
IEEE Trans. Pattern Anal. Mach. Intell.
\textbf{45}
(2022)
\end{botherref}
\endbibitem

\bibitem[\protect\citeauthoryear{Doersch et~al.}{2020}]{CTX}
\begin{bchapter}
\bauthor{\bsnm{Doersch}, \binits{C.}},
\bauthor{\bsnm{Gupta}, \binits{A.}},
\bauthor{\bsnm{Zisserman}, \binits{A.}}:
\bctitle{Cross{T}ransformers: {S}patially-aware few-shot transfer}.
In: \bbtitle{NuerIPS},
vol. \bseriesno{33},
pp. \bfpage{21981}--\blpage{21993}
(\byear{2020})
\end{bchapter}
\endbibitem

\bibitem[\protect\citeauthoryear{Wang et~al.}{2023}]{FSL_Transf_TCSVT}
\begin{barticle}
\bauthor{\bsnm{Wang}, \binits{X.}},
\bauthor{\bsnm{Wang}, \binits{X.}},
\bauthor{\bsnm{Jiang}, \binits{B.}},
\bauthor{\bsnm{Luo}, \binits{B.}}:
\batitle{Few-shot learning meets transformer: Unified query-support
  {T}ransformers for few-shot classification}.
\bjtitle{IEEE Trans. Circuits Syst. Video Technol.}
\bvolume{33},
\bfpage{7789}--\blpage{7802}
(\byear{2023})
\end{barticle}
\endbibitem

\bibitem[\protect\citeauthoryear{Ye et~al.}{2020}]{feat_2020}
\begin{bchapter}
\bauthor{\bsnm{Ye}, \binits{H.-J.}},
\bauthor{\bsnm{Hu}, \binits{H.}},
\bauthor{\bsnm{Zhan}, \binits{D.-C.}},
\bauthor{\bsnm{Sha}, \binits{F.}}:
\bctitle{Few-shot learning via embedding adaptation with set-to-set functions}.
In: \bbtitle{CVPR},
pp. \bfpage{8808}--\blpage{8817}
(\byear{2020})
\end{bchapter}
\endbibitem

\bibitem[\protect\citeauthoryear{Finn et~al.}{2017}]{ModelAgnostic}
\begin{bchapter}
\bauthor{\bsnm{Finn}, \binits{C.}},
\bauthor{\bsnm{Abbeel}, \binits{P.}},
\bauthor{\bsnm{Levine}, \binits{S.}}:
\bctitle{Model-agnostic meta-learning for fast adaptation of deep networks}.
In: \bbtitle{ICML},
pp. \bfpage{1126}--\blpage{1135}
(\byear{2017})
\end{bchapter}
\endbibitem

\bibitem[\protect\citeauthoryear{Wang et~al.}{2022}]{Wang2022GlobalCO}
\begin{bchapter}
\bauthor{\bsnm{Wang}, \binits{H.}},
\bauthor{\bsnm{Wang}, \binits{Y.}},
\bauthor{\bsnm{Sun}, \binits{R.}},
\bauthor{\bsnm{Li}, \binits{B.}}:
\bctitle{Global convergence of {MAML} and theory-inspired neural architecture
  search for few-shot learning}.
In: \bbtitle{CVPR},
pp. \bfpage{9787}--\blpage{9798}
(\byear{2022})
\end{bchapter}
\endbibitem

\bibitem[\protect\citeauthoryear{Antoniou et~al.}{2018}]{Antoniou2018HowTT}
\begin{botherref}
\oauthor{\bsnm{Antoniou}, \binits{A.}},
\oauthor{\bsnm{Edwards}, \binits{H.}},
\oauthor{\bsnm{Storkey}, \binits{A.J.}}:
How to train your {MAML}.
arXiv preprint arXiv:1810.09502
(2018)
\end{botherref}
\endbibitem

\bibitem[\protect\citeauthoryear{Andrychowicz
  et~al.}{2016}]{Andrychowicz2016LearningTL}
\begin{bchapter}
\bauthor{\bsnm{Andrychowicz}, \binits{M.}},
\bauthor{\bsnm{Denil}, \binits{M.}},
\bauthor{\bsnm{Colmenarejo}, \binits{S.G.}},
\bauthor{\bsnm{Hoffman}, \binits{M.W.}},
\bauthor{\bsnm{Pfau}, \binits{D.}},
\bauthor{\bsnm{Schaul}, \binits{T.}},
\bauthor{\bsnm{Freitas}, \binits{N.}}:
\bctitle{Learning to learn by gradient descent by gradient descent}.
In: \bbtitle{NIPS},
pp. \bfpage{3988}--\blpage{3996}
(\byear{2016})
\end{bchapter}
\endbibitem

\bibitem[\protect\citeauthoryear{Ravi and
  Larochelle}{2016}]{Ravi2016OptimizationAA}
\begin{bchapter}
\bauthor{\bsnm{Ravi}, \binits{S.}},
\bauthor{\bsnm{Larochelle}, \binits{H.}}:
\bctitle{Optimization as a model for few-shot learning}.
In: \bbtitle{ICLR}
(\byear{2016})
\end{bchapter}
\endbibitem

\bibitem[\protect\citeauthoryear{Carreira and
  Zisserman}{2017}]{i3d_Carreira_2017_CVPR}
\begin{bchapter}
\bauthor{\bsnm{Carreira}, \binits{J.}},
\bauthor{\bsnm{Zisserman}, \binits{A.}}:
\bctitle{Quo vadis, action recognition? {A} new model and the {K}inetics
  dataset}.
In: \bbtitle{CVPR},
pp. \bfpage{6299}--\blpage{6308}
(\byear{2017})
\end{bchapter}
\endbibitem

\bibitem[\protect\citeauthoryear{Tran et~al.}{2018}]{R2+1D_tran_2018_CVPR}
\begin{bchapter}
\bauthor{\bsnm{Tran}, \binits{D.}},
\bauthor{\bsnm{Wang}, \binits{H.}},
\bauthor{\bsnm{Torresani}, \binits{L.}},
\bauthor{\bsnm{Ray}, \binits{J.}},
\bauthor{\bsnm{LeCun}, \binits{Y.}},
\bauthor{\bsnm{Paluri}, \binits{M.}}:
\bctitle{A closer look at spatiotemporal convolutions for action recognition}.
In: \bbtitle{CVPR},
pp. \bfpage{6450}--\blpage{6459}
(\byear{2018})
\end{bchapter}
\endbibitem

\bibitem[\protect\citeauthoryear{Feichtenhofer
  et~al.}{2019}]{SlowFast_Feichtenhofer_2019_ICCV}
\begin{bchapter}
\bauthor{\bsnm{Feichtenhofer}, \binits{C.}},
\bauthor{\bsnm{Fan}, \binits{H.}},
\bauthor{\bsnm{Malik}, \binits{J.}},
\bauthor{\bsnm{He}, \binits{K.}}:
\bctitle{Slow{F}ast networks for video recognition}.
In: \bbtitle{ICCV},
pp. \bfpage{6202}--\blpage{6211}
(\byear{2019})
\end{bchapter}
\endbibitem

\bibitem[\protect\citeauthoryear{Feichtenhofer}{2020}]{X3D_Feichtenhofer_2020_CVPR}
\begin{bchapter}
\bauthor{\bsnm{Feichtenhofer}, \binits{C.}}:
\bctitle{X3{D}: Expanding architectures for efficient video recognition}.
In: \bbtitle{ECCV},
pp. \bfpage{203}--\blpage{213}
(\byear{2020})
\end{bchapter}
\endbibitem

\bibitem[\protect\citeauthoryear{Wang et~al.}{2023}]{AMSNet}
\begin{botherref}
\oauthor{\bsnm{Wang}, \binits{Q.}},
\oauthor{\bsnm{Hu}, \binits{Q.}},
\oauthor{\bsnm{Gao}, \binits{Z.}},
\oauthor{\bsnm{Li}, \binits{P.}},
\oauthor{\bsnm{Hu}, \binits{Q.}}:
A{MS-Net:} modeling adaptive multi-granularity spatio-temporal cues for video
  action recognition.
IEEE Trans. Neural Networks Learn. Syst.
(2023)
\end{botherref}
\endbibitem

\bibitem[\protect\citeauthoryear{Lin et~al.}{2024}]{VLG_IJCV}
\begin{botherref}
\oauthor{\bsnm{Lin}, \binits{J.}},
\oauthor{\bsnm{Liu}, \binits{Z.}},
\oauthor{\bsnm{Wang}, \binits{W.}},
\oauthor{\bsnm{Wu}, \binits{W.}},
\oauthor{\bsnm{Wang}, \binits{L.}}:
V{LG}: General video recognition with web textual knowledge.
Int J Comput Vis.,
1--26
(2024)
\end{botherref}
\endbibitem

\bibitem[\protect\citeauthoryear{Wang et~al.}{2018}]{TDO_IJCV}
\begin{barticle}
\bauthor{\bsnm{Wang}, \binits{L.}},
\bauthor{\bsnm{Wang}, \binits{Z.}},
\bauthor{\bsnm{Qiao}, \binits{Y.}},
\bauthor{\bsnm{Van~Gool}, \binits{L.}}:
\batitle{Transferring deep object and scene representations for event
  recognition in still images}.
\bjtitle{Int J Comput Vis.}
\bvolume{126},
\bfpage{390}--\blpage{409}
(\byear{2018})
\end{barticle}
\endbibitem

\bibitem[\protect\citeauthoryear{Tian et~al.}{2022}]{EAN_IJCV}
\begin{barticle}
\bauthor{\bsnm{Tian}, \binits{Y.}},
\bauthor{\bsnm{Yan}, \binits{Y.}},
\bauthor{\bsnm{Zhai}, \binits{G.}},
\bauthor{\bsnm{Guo}, \binits{G.}},
\bauthor{\bsnm{Gao}, \binits{Z.}}:
\batitle{E{AN}: event adaptive network for enhanced action recognition}.
\bjtitle{Int J Comput Vis.}
\bvolume{130}(\bissue{10}),
\bfpage{2453}--\blpage{2471}
(\byear{2022})
\end{barticle}
\endbibitem

\bibitem[\protect\citeauthoryear{Wu et~al.}{2021}]{C2F_IJCV}
\begin{barticle}
\bauthor{\bsnm{Wu}, \binits{Z.}},
\bauthor{\bsnm{Li}, \binits{H.}},
\bauthor{\bsnm{Zheng}, \binits{Y.}},
\bauthor{\bsnm{Xiong}, \binits{C.}},
\bauthor{\bsnm{Jiang}, \binits{Y.-G.}},
\bauthor{\bsnm{Davis}, \binits{L.S.}}:
\batitle{A coarse-to-fine framework for resource efficient video recognition}.
\bjtitle{Int J Comput Vis.}
\bvolume{129}(\bissue{11}),
\bfpage{2965}--\blpage{2977}
(\byear{2021})
\end{barticle}
\endbibitem

\bibitem[\protect\citeauthoryear{Su and Wen}{2021}]{TAP_2021}
\begin{bchapter}
\bauthor{\bsnm{Su}, \binits{B.}},
\bauthor{\bsnm{Wen}, \binits{J.-R.}}:
\bctitle{Temporal alignment prediction for supervised representation learning
  and few-shot sequence classification}.
In: \bbtitle{ICLR}
(\byear{2021})
\end{bchapter}
\endbibitem

\bibitem[\protect\citeauthoryear{Wang et~al.}{2021}]{CML}
\begin{bchapter}
\bauthor{\bsnm{Wang}, \binits{J.}},
\bauthor{\bsnm{Wang}, \binits{Y.}},
\bauthor{\bsnm{Liu}, \binits{S.}},
\bauthor{\bsnm{Li}, \binits{A.}}:
\bctitle{Few-shot fine-grained action recognition via bidirectional attention
  and contrastive meta-learning}.
In: \bbtitle{ACM Multimedia},
pp. \bfpage{582}--\blpage{591}
(\byear{2021})
\end{bchapter}
\endbibitem

\bibitem[\protect\citeauthoryear{Yang et~al.}{2020}]{TPN_CVPR20}
\begin{bchapter}
\bauthor{\bsnm{Yang}, \binits{C.}},
\bauthor{\bsnm{Xu}, \binits{Y.}},
\bauthor{\bsnm{Shi}, \binits{J.}},
\bauthor{\bsnm{Dai}, \binits{B.}},
\bauthor{\bsnm{Zhou}, \binits{B.}}:
\bctitle{Temporal pyramid network for action recognition}.
In: \bbtitle{CVPR},
pp. \bfpage{591}--\blpage{600}
(\byear{2020})
\end{bchapter}
\endbibitem

\bibitem[\protect\citeauthoryear{Zheng et~al.}{2022}]{HCL_eccv22}
\begin{bchapter}
\bauthor{\bsnm{Zheng}, \binits{S.}},
\bauthor{\bsnm{Chen}, \binits{S.}},
\bauthor{\bsnm{Jin}, \binits{Q.}}:
\bctitle{Few-shot action recognition with hierarchical matching and contrastive
  learning}.
In: \bbtitle{ECCV},
pp. \bfpage{297}--\blpage{313}
(\byear{2022})
\end{bchapter}
\endbibitem

\bibitem[\protect\citeauthoryear{Tang et~al.}{2023}]{M3Net_MM23}
\begin{bchapter}
\bauthor{\bsnm{Tang}, \binits{H.}},
\bauthor{\bsnm{Liu}, \binits{J.}},
\bauthor{\bsnm{Yan}, \binits{S.}},
\bauthor{\bsnm{Yan}, \binits{R.}},
\bauthor{\bsnm{Li}, \binits{Z.}},
\bauthor{\bsnm{Tang}, \binits{J.}}:
\bctitle{M{$^3$N}et: Multi-view encoding, matching, and fusion for few-shot
  fine-grained action recognition}.
In: \bbtitle{ACM Multimedia},
pp. \bfpage{1719}--\blpage{1728}
(\byear{2023})
\end{bchapter}
\endbibitem

\bibitem[\protect\citeauthoryear{Guo et~al.}{2024}]{CLIP_M2DF_arxiv24}
\begin{botherref}
\oauthor{\bsnm{Guo}, \binits{F.}},
\oauthor{\bsnm{Wang}, \binits{Y.}},
\oauthor{\bsnm{Qi}, \binits{H.}},
\oauthor{\bsnm{Jin}, \binits{W.}},
\oauthor{\bsnm{Zhu}, \binits{L.}}:
Multi-view distillation based on multi-modal fusion for few-shot action
  recognition ({CLIP-M$^2$DF}).
arXiv preprint arXiv:2401.08345
(2024)
\end{botherref}
\endbibitem

\bibitem[\protect\citeauthoryear{Wang et~al.}{2021}]{TRAPN}
\begin{bchapter}
\bauthor{\bsnm{Wang}, \binits{G.}},
\bauthor{\bsnm{Ye}, \binits{H.}},
\bauthor{\bsnm{Wang}, \binits{X.}},
\bauthor{\bsnm{Ye}, \binits{W.}},
\bauthor{\bsnm{Wang}, \binits{H.}}:
\bctitle{Temporal relation based attentive prototype network for few-shot
  action recognition}.
In: \bbtitle{ACML},
pp. \bfpage{406}--\blpage{421}
(\byear{2021})
\end{bchapter}
\endbibitem

\bibitem[\protect\citeauthoryear{Thatipelli et~al.}{2022}]{STRM_2022}
\begin{bchapter}
\bauthor{\bsnm{Thatipelli}, \binits{A.}},
\bauthor{\bsnm{Narayan}, \binits{S.}},
\bauthor{\bsnm{Khan}, \binits{S.}},
\bauthor{\bsnm{Anwer}, \binits{R.M.}},
\bauthor{\bsnm{Khan}, \binits{F.S.}},
\bauthor{\bsnm{Ghanem}, \binits{B.}}:
\bctitle{Spatio-temporal relation modeling for few-shot action recognition}.
In: \bbtitle{CVPR},
pp. \bfpage{19958}--\blpage{19967}
(\byear{2022})
\end{bchapter}
\endbibitem

\bibitem[\protect\citeauthoryear{Zhou et~al.}{2018}]{Zhou_2018_ECCV}
\begin{bchapter}
\bauthor{\bsnm{Zhou}, \binits{B.}},
\bauthor{\bsnm{Andonian}, \binits{A.}},
\bauthor{\bsnm{Oliva}, \binits{A.}},
\bauthor{\bsnm{Torralba}, \binits{A.}}:
\bctitle{Temporal relational reasoning in videos}.
In: \bbtitle{ECCV},
pp. \bfpage{803}--\blpage{818}
(\byear{2018})
\end{bchapter}
\endbibitem

\bibitem[\protect\citeauthoryear{Liu et~al.}{2023}]{MASTAF}
\begin{bchapter}
\bauthor{\bsnm{Liu}, \binits{X.}},
\bauthor{\bsnm{Zhang}, \binits{H.}},
\bauthor{\bsnm{Pirsiavash}, \binits{H.}}:
\bctitle{M{ASTAF}: {A} model-agnostic spatio-temporal attention fusion network
  for few-shot video classification}.
In: \bbtitle{WACV},
pp. \bfpage{2508}--\blpage{2517}
(\byear{2023})
\end{bchapter}
\endbibitem

\bibitem[\protect\citeauthoryear{Tang et~al.}{2024}]{SAFSAR}
\begin{bchapter}
\bauthor{\bsnm{Tang}, \binits{Y.}},
\bauthor{\bsnm{B{\'e}jar}, \binits{B.}},
\bauthor{\bsnm{Vidal}, \binits{R.}}:
\bctitle{Semantic-aware video representation for few-shot action recognition}.
In: \bbtitle{WACV},
pp. \bfpage{6458}--\blpage{6468}
(\byear{2024})
\end{bchapter}
\endbibitem

\bibitem[\protect\citeauthoryear{Wang et~al.}{2023}]{CLIP_FSAR_arXiv23}
\begin{botherref}
\oauthor{\bsnm{Wang}, \binits{X.}},
\oauthor{\bsnm{Zhang}, \binits{S.}},
\oauthor{\bsnm{Cen}, \binits{J.}},
\oauthor{\bsnm{Gao}, \binits{C.}},
\oauthor{\bsnm{Zhang}, \binits{Y.}},
\oauthor{\bsnm{Zhao}, \binits{D.}},
\oauthor{\bsnm{Sang}, \binits{N.}}:
Clip-guided prototype modulating for few-shot action recognition.
Int J Comput Vis.,
1--14
(2023)
\end{botherref}
\endbibitem

\bibitem[\protect\citeauthoryear{Xing et~al.}{2023}]{MACLIP_arXiv23}
\begin{botherref}
\oauthor{\bsnm{Xing}, \binits{J.}},
\oauthor{\bsnm{Wang}, \binits{M.}},
\oauthor{\bsnm{Hou}, \binits{X.}},
\oauthor{\bsnm{Dai}, \binits{G.}},
\oauthor{\bsnm{Wang}, \binits{J.}},
\oauthor{\bsnm{Liu}, \binits{Y.}}:
Multimodal adaptation of clip for few-shot action recognition.
arXiv preprint arXiv:2308.01532
(2023)
\end{botherref}
\endbibitem

\bibitem[\protect\citeauthoryear{Radford et~al.}{2021}]{CLIP_ICML}
\begin{bchapter}
\bauthor{\bsnm{Radford}, \binits{A.}},
\bauthor{\bsnm{Kim}, \binits{J.W.}},
\bauthor{\bsnm{Hallacy}, \binits{C.}},
\bauthor{\bsnm{Ramesh}, \binits{A.}},
\bauthor{\bsnm{Goh}, \binits{G.}},
\bauthor{\bsnm{Agarwal}, \binits{S.}},
\bauthor{\bsnm{Sastry}, \binits{G.}},
\bauthor{\bsnm{Askell}, \binits{A.}},
\bauthor{\bsnm{Mishkin}, \binits{P.}},
\bauthor{\bsnm{Clark}, \binits{J.}}, \betal:
\bctitle{Learning transferable visual models from natural language
  supervision}.
In: \bbtitle{International Conference on Machine Learning},
pp. \bfpage{8748}--\blpage{8763}
(\byear{2021}).
\bcomment{PMLR}
\end{bchapter}
\endbibitem

\bibitem[\protect\citeauthoryear{Wang et~al.}{2023}]{Actionclip_TNNLS}
\begin{botherref}
\oauthor{\bsnm{Wang}, \binits{M.}},
\oauthor{\bsnm{Xing}, \binits{J.}},
\oauthor{\bsnm{Mei}, \binits{J.}},
\oauthor{\bsnm{Liu}, \binits{Y.}},
\oauthor{\bsnm{Jiang}, \binits{Y.}}:
A{ctionCLIP: A}dapting language-image pretrained models for video action
  recognition.
IEEE Trans. Neural Networks Learn. Syst.
(2023)
\end{botherref}
\endbibitem

\bibitem[\protect\citeauthoryear{You et~al.}{2021}]{MACLIP}
\begin{botherref}
\oauthor{\bsnm{You}, \binits{H.}},
\oauthor{\bsnm{Zhou}, \binits{L.}},
\oauthor{\bsnm{Xiao}, \binits{B.}},
\oauthor{\bsnm{Codella}, \binits{N.C.}},
\oauthor{\bsnm{Cheng}, \binits{Y.}},
\oauthor{\bsnm{Xu}, \binits{R.}},
\oauthor{\bsnm{Chang}, \binits{S.-F.}},
\oauthor{\bsnm{Yuan}, \binits{L.}}:
M{A-CLIP}: towards modality-agnostic contrastive language-image pre-training
(2021)
\end{botherref}
\endbibitem

\bibitem[\protect\citeauthoryear{Gao et~al.}{2021}]{TCP_neurips_2021}
\begin{bchapter}
\bauthor{\bsnm{Gao}, \binits{Z.}},
\bauthor{\bsnm{Wang}, \binits{Q.}},
\bauthor{\bsnm{Zhang}, \binits{B.}},
\bauthor{\bsnm{Hu}, \binits{Q.}},
\bauthor{\bsnm{Li}, \binits{P.}}:
\bctitle{Temporal-attentive covariance pooling networks for video recognition}.
In: \bbtitle{NeurIPS}
(\byear{2021})
\end{bchapter}
\endbibitem

\bibitem[\protect\citeauthoryear{Dai et~al.}{2017}]{deform}
\begin{bchapter}
\bauthor{\bsnm{Dai}, \binits{J.}},
\bauthor{\bsnm{Qi}, \binits{H.}},
\bauthor{\bsnm{Xiong}, \binits{Y.}},
\bauthor{\bsnm{Li}, \binits{Y.}},
\bauthor{\bsnm{Zhang}, \binits{G.}},
\bauthor{\bsnm{Hu}, \binits{H.}},
\bauthor{\bsnm{Wei}, \binits{Y.}}:
\bctitle{Deformable convolutional networks}.
In: \bbtitle{ICCV},
pp. \bfpage{764}--\blpage{773}
(\byear{2017})
\end{bchapter}
\endbibitem

\bibitem[\protect\citeauthoryear{Zhao et~al.}{2018}]{Traj_nips18}
\begin{botherref}
\oauthor{\bsnm{Zhao}, \binits{Y.}},
\oauthor{\bsnm{Xiong}, \binits{Y.}},
\oauthor{\bsnm{Lin}, \binits{D.}}:
Trajectory convolution for action recognition.
NuerIPS
\textbf{31}
(2018)
\end{botherref}
\endbibitem

\bibitem[\protect\citeauthoryear{Pennec et~al.}{2006}]{pennec2006riemannian}
\begin{barticle}
\bauthor{\bsnm{Pennec}, \binits{X.}},
\bauthor{\bsnm{Fillard}, \binits{P.}},
\bauthor{\bsnm{Ayache}, \binits{N.}}:
\batitle{A {R}iemannian framework for tensor computing}.
\bjtitle{Int. J. Comput. Vision}
\bvolume{66},
\bfpage{41}--\blpage{66}
(\byear{2006})
\end{barticle}
\endbibitem

\bibitem[\protect\citeauthoryear{Arsigny et~al.}{2005}]{arsigny2005fast}
\begin{bchapter}
\bauthor{\bsnm{Arsigny}, \binits{V.}},
\bauthor{\bsnm{Fillard}, \binits{P.}},
\bauthor{\bsnm{Pennec}, \binits{X.}},
\bauthor{\bsnm{Ayache}, \binits{N.}}:
\bctitle{Fast and simple calculus on tensors in the {L}og-{E}uclidean
  framework}.
In: \bbtitle{International Conference on Medical Image Computing and
  Computer-Assisted Intervention},
pp. \bfpage{115}--\blpage{122}
(\byear{2005})
\end{bchapter}
\endbibitem

\bibitem[\protect\citeauthoryear{Li et~al.}{2017}]{MPNCOV_Li_2017_ICCV}
\begin{bchapter}
\bauthor{\bsnm{Li}, \binits{P.}},
\bauthor{\bsnm{Xie}, \binits{J.}},
\bauthor{\bsnm{Wang}, \binits{Q.}},
\bauthor{\bsnm{Zuo}, \binits{W.}}:
\bctitle{Is second-order information helpful for large-scale visual
  recognition?}
In: \bbtitle{ICCV},
pp. \bfpage{2070}--\blpage{2078}
(\byear{2017})
\end{bchapter}
\endbibitem

\bibitem[\protect\citeauthoryear{Li et~al.}{2018}]{iSQRT_2018_CVPR}
\begin{bchapter}
\bauthor{\bsnm{Li}, \binits{P.}},
\bauthor{\bsnm{Xie}, \binits{J.}},
\bauthor{\bsnm{Wang}, \binits{Q.}},
\bauthor{\bsnm{Gao}, \binits{Z.}}:
\bctitle{Towards faster training of global covariance pooling networks by
  iterative matrix square root normalization}.
In: \bbtitle{CVPR},
pp. \bfpage{947}--\blpage{955}
(\byear{2018})
\end{bchapter}
\endbibitem

\bibitem[\protect\citeauthoryear{Goyal et~al.}{2017}]{ssv2}
\begin{bchapter}
\bauthor{\bsnm{Goyal}, \binits{R.}},
\bauthor{\bsnm{Ebrahimi~Kahou}, \binits{S.}},
\bauthor{\bsnm{Michalski}, \binits{V.}},
\bauthor{\bsnm{Materzynska}, \binits{J.}},
\bauthor{\bsnm{Westphal}, \binits{S.}},
\bauthor{\bsnm{Kim}, \binits{H.}},
\bauthor{\bsnm{Haenel}, \binits{V.}},
\bauthor{\bsnm{Fruend}, \binits{I.}},
\bauthor{\bsnm{Yianilos}, \binits{P.}},
\bauthor{\bsnm{Mueller-Freitag}, \binits{M.}},
\bauthor{\bsnm{Hoppe}, \binits{F.}},
\bauthor{\bsnm{Thurau}, \binits{C.}},
\bauthor{\bsnm{Bax}, \binits{I.}},
\bauthor{\bsnm{Memisevic}, \binits{R.}}:
\bctitle{The ``{S}omething {S}omething" video database for learning and
  evaluating visual common sense}.
In: \bbtitle{ICCV}
(\byear{2017})
\end{bchapter}
\endbibitem

\bibitem[\protect\citeauthoryear{He et~al.}{2016}]{ResNet_He_2016_CVPR}
\begin{bchapter}
\bauthor{\bsnm{He}, \binits{K.}},
\bauthor{\bsnm{Zhang}, \binits{X.}},
\bauthor{\bsnm{Ren}, \binits{S.}},
\bauthor{\bsnm{Sun}, \binits{J.}}:
\bctitle{Deep residual learning for image recognition}.
In: \bbtitle{CVPR}
(\byear{2016})
\end{bchapter}
\endbibitem

\bibitem[\protect\citeauthoryear{Deng et~al.}{2009}]{ImageNet-1K}
\begin{bchapter}
\bauthor{\bsnm{Deng}, \binits{J.}},
\bauthor{\bsnm{Dong}, \binits{W.}},
\bauthor{\bsnm{Socher}, \binits{R.}},
\bauthor{\bsnm{Li}, \binits{L.-J.}},
\bauthor{\bsnm{Li}, \binits{K.}},
\bauthor{\bsnm{Fei-Fei}, \binits{L.}}:
\bctitle{Image{N}et: {A} large-scale hierarchical image database}.
In: \bbtitle{CVPR},
pp. \bfpage{248}--\blpage{255}
(\byear{2009})
\end{bchapter}
\endbibitem

\bibitem[\protect\citeauthoryear{He et~al.}{2022}]{MAE_2022}
\begin{bchapter}
\bauthor{\bsnm{He}, \binits{K.}},
\bauthor{\bsnm{Chen}, \binits{X.}},
\bauthor{\bsnm{Xie}, \binits{S.}},
\bauthor{\bsnm{Li}, \binits{Y.}},
\bauthor{\bsnm{Doll\'ar}, \binits{P.}},
\bauthor{\bsnm{Girshick}, \binits{R.B.}}:
\bctitle{Masked autoencoders are scalable vision learners}.
In: \bbtitle{CVPR},
pp. \bfpage{15979}--\blpage{15988}
(\byear{2022})
\end{bchapter}
\endbibitem

\bibitem[\protect\citeauthoryear{Radford et~al.}{2021}]{CLIP}
\begin{bchapter}
\bauthor{\bsnm{Radford}, \binits{A.}},
\bauthor{\bsnm{Kim}, \binits{J.W.}},
\bauthor{\bsnm{Hallacy}, \binits{C.}},
\bauthor{\bsnm{Ramesh}, \binits{A.}},
\bauthor{\bsnm{Goh}, \binits{G.}},
\bauthor{\bsnm{Agarwal}, \binits{S.}},
\bauthor{\bsnm{Sastry}, \binits{G.}},
\bauthor{\bsnm{Askell}, \binits{A.}},
\bauthor{\bsnm{Mishkin}, \binits{P.}},
\bauthor{\bsnm{Clark}, \binits{J.}}, \betal:
\bctitle{Learning transferable visual models from natural language
  supervision}.
In: \bbtitle{ICML},
pp. \bfpage{8748}--\blpage{8763}
(\byear{2021})
\end{bchapter}
\endbibitem

\bibitem[\protect\citeauthoryear{Huang et~al.}{2022}]{CPM_2022}
\begin{bchapter}
\bauthor{\bsnm{Huang}, \binits{Y.}},
\bauthor{\bsnm{Yang}, \binits{L.}},
\bauthor{\bsnm{Sato}, \binits{Y.}}:
\bctitle{Compound prototype matching for few-shot action recognition}.
In: \bbtitle{ECCV},
pp. \bfpage{351}--\blpage{368}
(\byear{2022})
\end{bchapter}
\endbibitem

\bibitem[\protect\citeauthoryear{}{2024}]{OpenCVDocs}
\begin{botherref}
OpenCV.
\url{https://docs.opencv.org/4.x/index.html}.
Accessed: June 14, 2024
(2024)
\end{botherref}
\endbibitem

\bibitem[\protect\citeauthoryear{Wang et~al.}{2019}]{TSN_PAMI}
\begin{barticle}
\bauthor{\bsnm{Wang}, \binits{L.}},
\bauthor{\bsnm{Xiong}, \binits{Y.}},
\bauthor{\bsnm{Wang}, \binits{Z.}},
\bauthor{\bsnm{Qiao}, \binits{Y.}},
\bauthor{\bsnm{Lin}, \binits{D.}},
\bauthor{\bsnm{Tang}, \binits{X.}},
\bauthor{\bsnm{Gool}, \binits{L.V.}}:
\batitle{Temporal segment networks for action recognition in videos}.
\bjtitle{{IEEE} Trans. Pattern Anal. Mach. Intell.}
\bvolume{41}(\bissue{11}),
\bfpage{2740}--\blpage{2755}
(\byear{2019})
\end{barticle}
\endbibitem

\bibitem[\protect\citeauthoryear{Vinyals et~al.}{2016}]{matchingnet}
\begin{bchapter}
\bauthor{\bsnm{Vinyals}, \binits{O.}},
\bauthor{\bsnm{Blundell}, \binits{C.}},
\bauthor{\bsnm{Lillicrap}, \binits{T.P.}},
\bauthor{\bsnm{Kavukcuoglu}, \binits{K.}},
\bauthor{\bsnm{Wierstra}, \binits{D.}}:
\bctitle{Matching networks for one shot learning}.
In: \bbtitle{NIPS}
(\byear{2016})
\end{bchapter}
\endbibitem

\bibitem[\protect\citeauthoryear{Finn et~al.}{2017}]{maml_2017}
\begin{bchapter}
\bauthor{\bsnm{Finn}, \binits{C.}},
\bauthor{\bsnm{Abbeel}, \binits{P.}},
\bauthor{\bsnm{Levine}, \binits{S.}}:
\bctitle{Model-agnostic meta-learning for fast adaptation of deep networks}.
In: \bbtitle{ICML},
pp. \bfpage{1126}--\blpage{1135}
(\byear{2017})
\end{bchapter}
\endbibitem

\bibitem[\protect\citeauthoryear{Nguyen et~al.}{2022}]{Nguyen_ECCV22}
\begin{bchapter}
\bauthor{\bsnm{Nguyen}, \binits{K.D.}},
\bauthor{\bsnm{Tran}, \binits{Q.-H.}},
\bauthor{\bsnm{Nguyen}, \binits{K.}},
\bauthor{\bsnm{Hua}, \binits{B.-S.}},
\bauthor{\bsnm{Nguyen}, \binits{R.}}:
\bctitle{Inductive and transductive few-shot video classification via
  appearance and temporal alignments}.
In: \bbtitle{ECCV},
pp. \bfpage{471}--\blpage{487}
(\byear{2022})
\end{bchapter}
\endbibitem

\bibitem[\protect\citeauthoryear{Zhu et~al.}{2021}]{PAL_21}
\begin{bchapter}
\bauthor{\bsnm{Zhu}, \binits{X.}},
\bauthor{\bsnm{Toisoul}, \binits{A.}},
\bauthor{\bsnm{Prez-Ra}, \binits{J.-M.}},
\bauthor{\bsnm{Zhang}, \binits{L.}},
\bauthor{\bsnm{Mart{\'i}nez}, \binits{B.}},
\bauthor{\bsnm{Xiang}, \binits{T.}}:
\bctitle{Few-shot action recognition with prototype-centered attentive
  learning}.
In: \bbtitle{BMVC}
(\byear{2021})
\end{bchapter}
\endbibitem

\bibitem[\protect\citeauthoryear{Liu et~al.}{2022}]{Liu2022MultidimensionalPR}
\begin{barticle}
\bauthor{\bsnm{Liu}, \binits{S.}},
\bauthor{\bsnm{Jiang}, \binits{M.}},
\bauthor{\bsnm{Kong}, \binits{J.}}:
\batitle{Multidimensional prototype refactor enhanced network for few-shot
  action recognition}.
\bjtitle{IEEE Trans. Circuits Syst. Video Technol.}
\bvolume{32},
\bfpage{6955}--\blpage{6966}
(\byear{2022})
\end{barticle}
\endbibitem

\bibitem[\protect\citeauthoryear{Zhang et~al.}{2023}]{STA_CT_MM23}
\begin{botherref}
\oauthor{\bsnm{Zhang}, \binits{Y.}},
\oauthor{\bsnm{Fu}, \binits{Y.}},
\oauthor{\bsnm{Ma}, \binits{X.}},
\oauthor{\bsnm{Qi}, \binits{L.}},
\oauthor{\bsnm{Chen}, \binits{J.}},
\oauthor{\bsnm{Wu}, \binits{Z.}},
\oauthor{\bsnm{Jiang}, \binits{Y.-G.}}:
On the importance of spatial relations for few-shot action recognition.
ACM Multimedia
(2023)
\end{botherref}
\endbibitem

\bibitem[\protect\citeauthoryear{Xing et~al.}{2023}]{GgHM_ICCV23}
\begin{bchapter}
\bauthor{\bsnm{Xing}, \binits{J.}},
\bauthor{\bsnm{Wang}, \binits{M.}},
\bauthor{\bsnm{Ruan}, \binits{Y.}},
\bauthor{\bsnm{Chen}, \binits{B.}},
\bauthor{\bsnm{Guo}, \binits{Y.}},
\bauthor{\bsnm{Mu}, \binits{B.}},
\bauthor{\bsnm{Dai}, \binits{G.}},
\bauthor{\bsnm{Wang}, \binits{J.}},
\bauthor{\bsnm{Liu}, \binits{Y.}}:
\bctitle{Boosting few-shot action recognition with graph-guided hybrid
  matching}.
In: \bbtitle{ICCV},
pp. \bfpage{1740}--\blpage{1750}
(\byear{2023})
\end{bchapter}
\endbibitem

\bibitem[\protect\citeauthoryear{Xia et~al.}{2023}]{RFPL_ICCV23}
\begin{bchapter}
\bauthor{\bsnm{Xia}, \binits{H.}},
\bauthor{\bsnm{Li}, \binits{K.}},
\bauthor{\bsnm{Min}, \binits{M.R.}},
\bauthor{\bsnm{Ding}, \binits{Z.}}:
\bctitle{Few-shot video classification via representation fusion and promotion
  learning}.
In: \bbtitle{ICCV},
pp. \bfpage{19311}--\blpage{19320}
(\byear{2023})
\end{bchapter}
\endbibitem

\bibitem[\protect\citeauthoryear{Li et~al.}{2022}]{TAN2N_li2022ta2n}
\begin{bchapter}
\bauthor{\bsnm{Li}, \binits{S.}},
\bauthor{\bsnm{Liu}, \binits{H.}},
\bauthor{\bsnm{Qian}, \binits{R.}},
\bauthor{\bsnm{Li}, \binits{Y.}},
\bauthor{\bsnm{See}, \binits{J.}},
\bauthor{\bsnm{Fei}, \binits{M.}},
\bauthor{\bsnm{Yu}, \binits{X.}},
\bauthor{\bsnm{Lin}, \binits{W.}}:
\bctitle{T{A$^2$N}: Two-stage action alignment network for few-shot action
  recognition}.
In: \bbtitle{AAAI},
vol. \bseriesno{36},
pp. \bfpage{1404}--\blpage{1411}
(\byear{2022})
\end{bchapter}
\endbibitem

\bibitem[\protect\citeauthoryear{{Lin} et~al.}{2015}]{BCNN_lin_2015_ICCV}
\begin{bchapter}
\bauthor{\bsnm{{Lin}}, \binits{T.}},
\bauthor{\bsnm{{RoyChowdhury}}, \binits{A.}},
\bauthor{\bsnm{{Maji}}, \binits{S.}}:
\bctitle{Bilinear {CNN} models for fine-grained visual recognition}.
In: \bbtitle{ICCV},
pp. \bfpage{1449}--\blpage{1457}
(\byear{2015})
\end{bchapter}
\endbibitem

\bibitem[\protect\citeauthoryear{Cui et~al.}{2017}]{KP_Cui_2017_CVPR}
\begin{bchapter}
\bauthor{\bsnm{Cui}, \binits{Y.}},
\bauthor{\bsnm{Zhou}, \binits{F.}},
\bauthor{\bsnm{Wang}, \binits{J.}},
\bauthor{\bsnm{Liu}, \binits{X.}},
\bauthor{\bsnm{Lin}, \binits{Y.}},
\bauthor{\bsnm{Belongie}, \binits{S.}}:
\bctitle{Kernel pooling for convolutional neural networks}.
In: \bbtitle{CVPR}
(\byear{2017})
\end{bchapter}
\endbibitem

\bibitem[\protect\citeauthoryear{Dvornik et~al.}{2021}]{Drop_DTW}
\begin{barticle}
\bauthor{\bsnm{Dvornik}, \binits{M.}},
\bauthor{\bsnm{Hadji}, \binits{I.}},
\bauthor{\bsnm{Derpanis}, \binits{K.G.}},
\bauthor{\bsnm{Garg}, \binits{A.}},
\bauthor{\bsnm{Jepson}, \binits{A.}}:
\batitle{Drop-dtw: Aligning common signal between sequences while dropping
  outliers}.
\bjtitle{NeurIPS}
\bvolume{34},
\bfpage{13782}--\blpage{13793}
(\byear{2021})
\end{barticle}
\endbibitem

\bibitem[\protect\citeauthoryear{Shi et~al.}{2024}]{KP}
\begin{botherref}
\oauthor{\bsnm{Shi}, \binits{Y.}},
\oauthor{\bsnm{Wu}, \binits{X.}},
\oauthor{\bsnm{Lin}, \binits{H.}}:
Knowledge prompting for few-shot action recognition.
TMM
(2024)
\end{botherref}
\endbibitem

\bibitem[\protect\citeauthoryear{Guo et~al.}{2023}]{CLIPCPM2C}
\begin{botherref}
\oauthor{\bsnm{Guo}, \binits{F.}},
\oauthor{\bsnm{Zhu}, \binits{L.}},
\oauthor{\bsnm{Wang}, \binits{Y.}},
\oauthor{\bsnm{Qi}, \binits{H.}}:
Consistency prototype module and motion compensation for few-shot action
  recognition ({CLIP-CPM2C}).
arXiv preprint arXiv:2312.01083
(2023)
\end{botherref}
\endbibitem

\bibitem[\protect\citeauthoryear{Guo et~al.}{2024}]{CLIPM2DF}
\begin{barticle}
\bauthor{\bsnm{Guo}, \binits{F.}},
\bauthor{\bsnm{Wang}, \binits{Y.}},
\bauthor{\bsnm{Qi}, \binits{H.}},
\bauthor{\bsnm{Jin}, \binits{W.}},
\bauthor{\bsnm{Zhu}, \binits{L.}},
\bauthor{\bsnm{Sun}, \binits{J.}}:
\batitle{Multi-view distillation based on multi-modal fusion for few-shot
  action recognition {(CLIP-MDMF)}}.
\bjtitle{Knowledge-Based Systems}
\bvolume{304},
\bfpage{112539}
(\byear{2024})
\end{barticle}
\endbibitem

\bibitem[\protect\citeauthoryear{Wang et~al.}{2024}]{CLIPFSAR}
\begin{barticle}
\bauthor{\bsnm{Wang}, \binits{X.}},
\bauthor{\bsnm{Zhang}, \binits{S.}},
\bauthor{\bsnm{Cen}, \binits{J.}},
\bauthor{\bsnm{Gao}, \binits{C.}},
\bauthor{\bsnm{Zhang}, \binits{Y.}},
\bauthor{\bsnm{Zhao}, \binits{D.}},
\bauthor{\bsnm{Sang}, \binits{N.}}:
\batitle{Clip-guided prototype modulating for few-shot action recognition}.
\bjtitle{International Journal of Computer Vision}
\bvolume{132}(\bissue{6}),
\bfpage{1899}--\blpage{1912}
(\byear{2024})
\end{barticle}
\endbibitem

\end{thebibliography}



\begin{thebibliography}{6}
\ifx \bisbn   \undefined \def \bisbn  #1{ISBN #1}\fi
\ifx \binits  \undefined \def \binits#1{#1}\fi
\ifx \bauthor  \undefined \def \bauthor#1{#1}\fi
\ifx \batitle  \undefined \def \batitle#1{#1}\fi
\ifx \bjtitle  \undefined \def \bjtitle#1{#1}\fi
\ifx \bvolume  \undefined \def \bvolume#1{\textbf{#1}}\fi
\ifx \byear  \undefined \def \byear#1{#1}\fi
\ifx \bissue  \undefined \def \bissue#1{#1}\fi
\ifx \bfpage  \undefined \def \bfpage#1{#1}\fi
\ifx \blpage  \undefined \def \blpage #1{#1}\fi
\ifx \burl  \undefined \def \burl#1{\textsf{#1}}\fi
\ifx \doiurl  \undefined \def \doiurl#1{\url{https://doi.org/#1}}\fi
\ifx \betal  \undefined \def \betal{\textit{et al.}}\fi
\ifx \binstitute  \undefined \def \binstitute#1{#1}\fi
\ifx \binstitutionaled  \undefined \def \binstitutionaled#1{#1}\fi
\ifx \bctitle  \undefined \def \bctitle#1{#1}\fi
\ifx \beditor  \undefined \def \beditor#1{#1}\fi
\ifx \bpublisher  \undefined \def \bpublisher#1{#1}\fi
\ifx \bbtitle  \undefined \def \bbtitle#1{#1}\fi
\ifx \bedition  \undefined \def \bedition#1{#1}\fi
\ifx \bseriesno  \undefined \def \bseriesno#1{#1}\fi
\ifx \blocation  \undefined \def \blocation#1{#1}\fi
\ifx \bsertitle  \undefined \def \bsertitle#1{#1}\fi
\ifx \bsnm \undefined \def \bsnm#1{#1}\fi
\ifx \bsuffix \undefined \def \bsuffix#1{#1}\fi
\ifx \bparticle \undefined \def \bparticle#1{#1}\fi
\ifx \barticle \undefined \def \barticle#1{#1}\fi
\bibcommenthead
\ifx \bconfdate \undefined \def \bconfdate #1{#1}\fi
\ifx \botherref \undefined \def \botherref #1{#1}\fi
\ifx \url \undefined \def \url#1{\textsf{#1}}\fi
\ifx \bchapter \undefined \def \bchapter#1{#1}\fi
\ifx \bbook \undefined \def \bbook#1{#1}\fi
\ifx \bcomment \undefined \def \bcomment#1{#1}\fi
\ifx \oauthor \undefined \def \oauthor#1{#1}\fi
\ifx \citeauthoryear \undefined \def \citeauthoryear#1{#1}\fi
\ifx \endbibitem  \undefined \def \endbibitem {}\fi
\ifx \bconflocation  \undefined \def \bconflocation#1{#1}\fi
\ifx \arxivurl  \undefined \def \arxivurl#1{\textsf{#1}}\fi
\csname PreBibitemsHook\endcsname

\bibitem[\protect\citeauthoryear{Rubner et~al.}{2000}]{EMD_2000}
\begin{barticle}
\bauthor{\bsnm{Rubner}, \binits{Y.}},
\bauthor{\bsnm{Tomasi}, \binits{C.}},
\bauthor{\bsnm{Guibas}, \binits{L.J.}}:
\batitle{The earth mover's distance as a metric for image retrieval}.
\bjtitle{Int. J. Comput. Vision}
\bvolume{40}(\bissue{2}),
\bfpage{99}--\blpage{121}
(\byear{2000})
\end{barticle}
\endbibitem

\bibitem[\protect\citeauthoryear{Li et~al.}{2013}]{SR_EMD}
\begin{bchapter}
\bauthor{\bsnm{Li}, \binits{P.}},
\bauthor{\bsnm{Wang}, \binits{Q.}},
\bauthor{\bsnm{Zhang}, \binits{L.}}:
\bctitle{A novel earth mover's distance methodology for imagematching with
  gaussian mixture models}.
In: \bbtitle{ICCV}
(\byear{2013})
\end{bchapter}
\endbibitem

\bibitem[\protect\citeauthoryear{Christian et~al.}{2011}]{quadratic_sim}
\begin{bchapter}
\bauthor{\bsnm{Christian}, \binits{B.}},
\bauthor{\bsnm{Anca}, \binits{M.I.}},
\bauthor{\bsnm{Steffen}, \binits{K.}},
\bauthor{\bsnm{Thomas}, \binits{S.}}:
\bctitle{Modeling image similarity by gaussian mixture models and the signature
  quadratic form distance}.
In: \bbtitle{ICCV},
pp. \bfpage{1754}--\blpage{1761}
(\byear{2011})
\end{bchapter}
\endbibitem

\bibitem[\protect\citeauthoryear{Pan et~al.}{2022}]{ST_adapter}
\begin{barticle}
\bauthor{\bsnm{Pan}, \binits{J.}},
\bauthor{\bsnm{Lin}, \binits{Z.}},
\bauthor{\bsnm{Zhu}, \binits{X.}},
\bauthor{\bsnm{Shao}, \binits{J.}},
\bauthor{\bsnm{Li}, \binits{H.}}:
\batitle{S{T}-adapter: Parameter-efficient image-to-video transfer learning}.
\bjtitle{NeurIPS}
\bvolume{35},
\bfpage{26462}--\blpage{26477}
(\byear{2022})
\end{barticle}
\endbibitem

\bibitem[\protect\citeauthoryear{Wang et~al.}{2024}]{CLIPFSAR}
\begin{barticle}
\bauthor{\bsnm{Wang}, \binits{X.}},
\bauthor{\bsnm{Zhang}, \binits{S.}},
\bauthor{\bsnm{Cen}, \binits{J.}},
\bauthor{\bsnm{Gao}, \binits{C.}},
\bauthor{\bsnm{Zhang}, \binits{Y.}},
\bauthor{\bsnm{Zhao}, \binits{D.}},
\bauthor{\bsnm{Sang}, \binits{N.}}:
\batitle{Clip-guided prototype modulating for few-shot action recognition}.
\bjtitle{International Journal of Computer Vision}
\bvolume{132}(\bissue{6}),
\bfpage{1899}--\blpage{1912}
(\byear{2024})
\end{barticle}
\endbibitem

\bibitem[\protect\citeauthoryear{Radford et~al.}{2021}]{CLIP}
\begin{bchapter}
\bauthor{\bsnm{Radford}, \binits{A.}},
\bauthor{\bsnm{Kim}, \binits{J.W.}},
\bauthor{\bsnm{Hallacy}, \binits{C.}},
\bauthor{\bsnm{Ramesh}, \binits{A.}},
\bauthor{\bsnm{Goh}, \binits{G.}},
\bauthor{\bsnm{Agarwal}, \binits{S.}},
\bauthor{\bsnm{Sastry}, \binits{G.}},
\bauthor{\bsnm{Askell}, \binits{A.}},
\bauthor{\bsnm{Mishkin}, \binits{P.}},
\bauthor{\bsnm{Clark}, \binits{J.}}, \betal:
\bctitle{Learning transferable visual models from natural language
  supervision}.
In: \bbtitle{ICML},
pp. \bfpage{8748}--\blpage{8763}
(\byear{2021})
\end{bchapter}
\endbibitem

\end{thebibliography}

\putbib                
\end{bibunit}

\clearpage
\onecolumn

\begin{bibunit}[sn-mathphys-num]


\begingroup
\makeatletter
\renewcommand{\bibnumfmt}[1]{[S#1]}

\renewcommand{\citenumfont}[1]{\textcolor{red}{S}#1}

\makeatother

\begin{appendices}

\newcommand{\ouracc}[1]{\textbf{#1}}

\section*{Appendix}

\section{Temporal alignment paradigms}
We illustrate various temporal alignment paradigms in Figure~\ref{fig:align_counterparts}, as referenced in Table 3(e). Specifically, figure (a) depicts a simple alignment paradigm, matching representations with cosine similarity at the same timestamp in a single-scale setting; figure (b) allows for alignment across different timestamps by using A$^2$ module within the single scale. In a multi-scale setting, \revision{(c) and (d) employ a fixed alignment approach by computing cosine similarity in a (c) point-to-point matching (p-p) or (d) cross matching (cr.) manner. Specifically, they compute the similarity between query and support features at (c) corresponding timestamps within each scale or (d) cross various timestamps or scales.} While figure (e) introduces an adaptive alignment strategy (A$^2$ module) that permits alignment across various timestamps or even across scales, as indicated by the red lines in the figure.

\begin{figure}[H]
  \centering
  \includegraphics[width=0.55\textwidth]{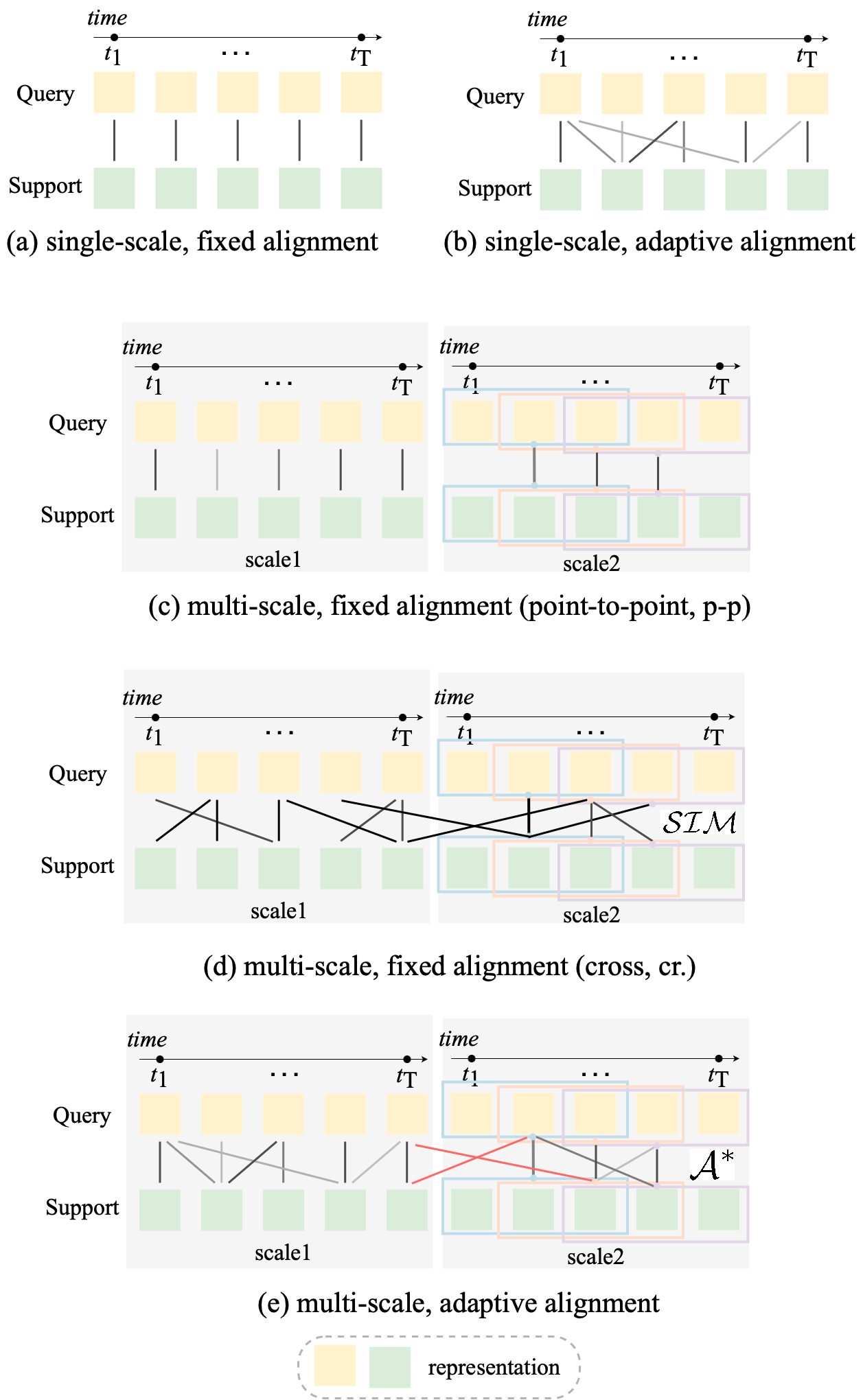}
  \caption{Various temporal alignment paradigms in Table 3(e).}
  \label{fig:align_counterparts}
\end{figure}


\section{Discussion on \texorpdfstring{$\mathcal{SIM}$}{SIM} matrix
and \texorpdfstring{$\mathcal{A}^*$}{A*}}

\newcommand{\colorl}{\textcolor{red}{l}}

\mnrevision{In this section, we will discuss  regarding why we need a linear programming formulation to obtain the optimal transportation matrix $\mathcal{A}^*$. Below, we provide additional theoretical and qualitative analysis to clarify why we use $\mathcal{A}^*$ rather than $\mathcal{SIM}$.}

\mnrevision{Notably, although the optimal transportation matrix $\mathcal{A}^{*}$ originates from the similarity matrix $\mathcal{SIM}$, it further refines each alignment weight $\mathcal{A}_{l,l'}$ by considering the whole video clip, whereas $\mathcal{SIM}$ only accounts for visual similarity between two separate frames.}

\mnrevision{More concretely, $\mathcal{A}_{\colorl,l'}$ is influenced by the importance of the $\colorl$-th query frame $\mathbf{Q}^{[\colorl]}$ relative to the whole support video clip $\mathbf{S}$. If $\mathbf{Q}^{[\colorl]}$ has a strong correlation with the whole support video $\mathbf{S}$, it receives a larger mass $\mu_{\colorl}$, indicating that this query frame is likely more discriminative with respect to the support video. Consequently, the alignment weights $\mathcal{A}_{\colorl,:}$ tend to be enlarged. Conversely, if $\mathbf{Q}^{[\colorl]}$ has a weaker correlation with the support video, its alignment weights $\mathcal{A}_{\colorl,:}$ are reduced. We adaptively adjust the alignment based on how each query frame relates to the support video (or vice versa using $\gamma_{l'}$). By incorporating this global context, $\mathcal{A}^*$ produces an optimal overall alignment strategy that minimizes the alignment cost.}

\mnrevision{The idea of $\mathcal{A}^{*}$ stems from the optimal transportation theory~\cite{EMD_2000}, also known as the Earth Mover’s Distance (EMD).  This algorithm seeks to find the minimal ``cost" of transforming one distribution into another and has proven more effective~\cite{SR_EMD} on handling misalignment than simple similarity-based strategy such as quadratic minimization~\cite{quadratic_sim}.}

\begin{figure}[!ht]
  \centering
  \begin{subfigure}[b]{0.45\linewidth}
    \centering
    \includegraphics[width=\linewidth]{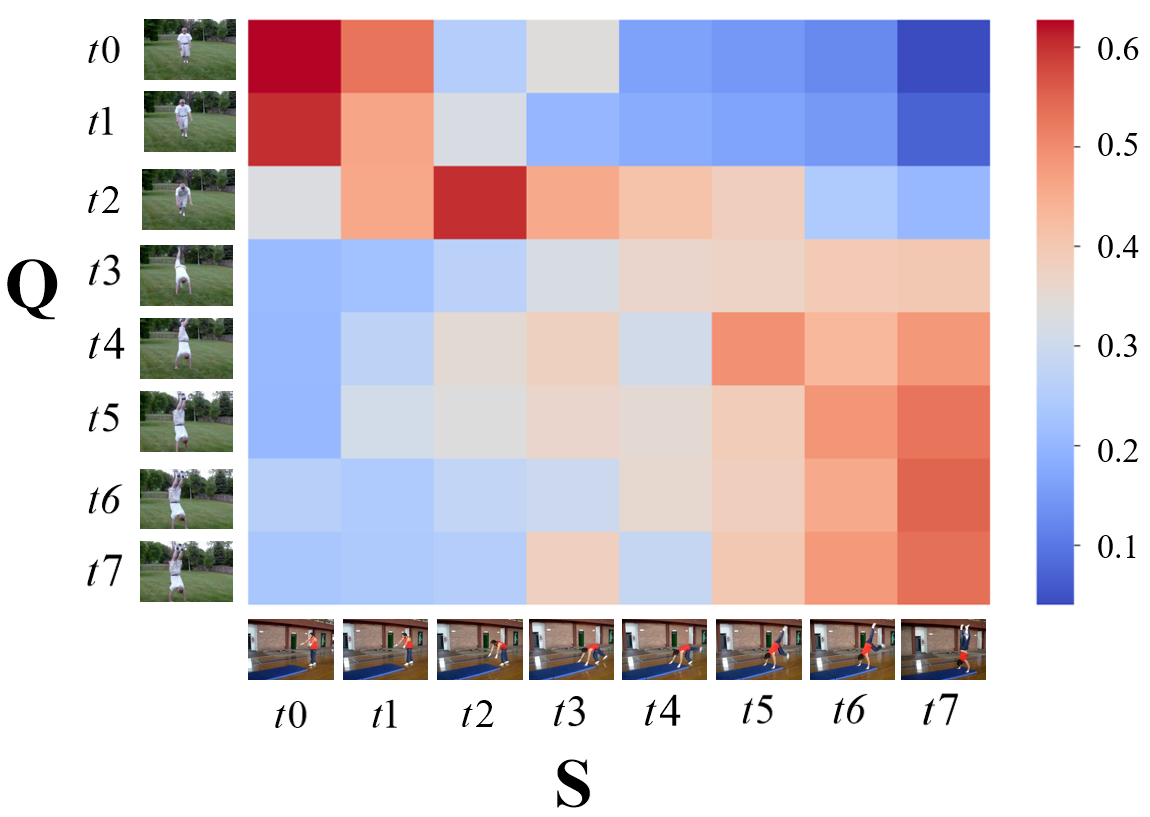}
    \caption{ $\mathcal{SIM}$}
    \label{fig:vis_sim}
  \end{subfigure}
  \begin{subfigure}[b]{0.45\linewidth}
    \centering
    \includegraphics[width=\linewidth]{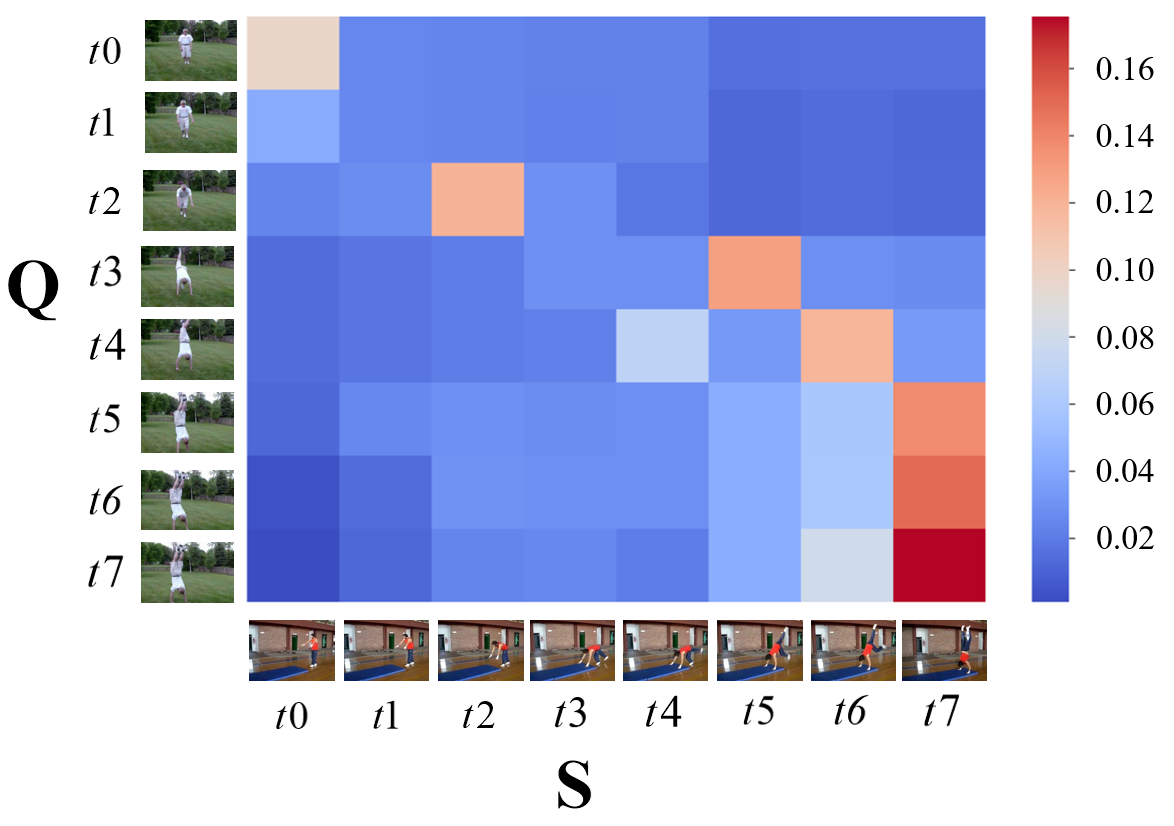}
    \caption{$\mathcal{A}^*$}
    \label{fig:vis_A_star}
  \end{subfigure}
  \caption{The comparison of temporal alignment results of video clips ($\mathbf{Q}$ and $\mathbf{S}$ as shown in Fig.~\ref{fig:videos_for_sim_matrix}) for similarity matrix $\mathcal{SIM}$ and our optimal alignment matrix $\mathcal{A}^*$.}
  \label{fig:vis_alignment_matrix}
\end{figure}
\begin{figure}[h]
  \centering
  \includegraphics[width=0.85\linewidth]{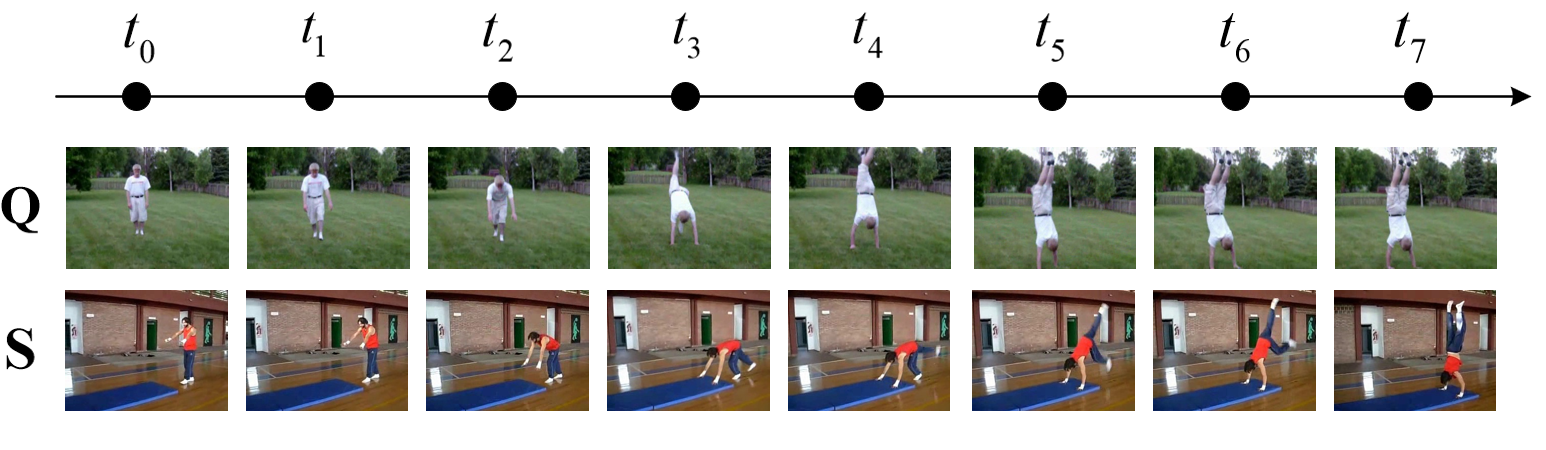}
  \caption{The video clips used in Figure~\ref{fig:vis_alignment_matrix}.}
  \label{fig:videos_for_sim_matrix}
\end{figure}

\newcommand{\acc}[2]{#1$/$#2\%}

Furthermore, our empirical study, as shown in Table~\ref{tab:sim_score}, demonstrates that removing $\mathcal{A}^*$ and relying solely on the $\mathcal{SIM}$ matrix for computing prediction scores leads to a substantial performance drop of \acc{14.3}{16.8}. This significant decrease underscores the necessity of the $\mathcal{A}^*$ matrix in considering the global relationship for temporal alignment problem.

\revision{In conclusion, computing $\mathcal{A}^*$ is important for both adaptive alignment by the global relationship and making accurate predictions.}

\begin{table}[!htb]
\centering
\tablestyle{14pt}{1.8}
\renewcommand\arraystretch{1.8}
\caption{\revision{Evaluation on various relation metric for computing similarity scores in A$^2$ module. Reported 5-way accuracy (\%) on SSV2-Full dataset using ResNet-50 network. The superior performances are noted by the \ouracc{boldface}. $f_{cos}$: cosine similarity function. $\mathbf{Q}^{[l]}$ and $\mathbf{S}^{[l']}$ indicate the $l$-th and $l'$-th timestamp descriptor in sequence of $\mathbf{Q}$ and $\mathbf{S}$ respectively, and $L$ is their temporal length.}}\label{tab:sim_score}
\begin{tabular}{lccc}
\hline
Temporal relation metric & \makecell{Prediction score \\ $sim\left(\mathcal{Q},\mathcal{S}\right)$} & 1-shot & 5-shot \\
\hline
$\mathcal{SIM}$ & $\sum_{l,l'}f_{cos}(\mathbf{Q}^{[l]}, \mathbf{S}^{[l']}) \cdot \frac{1}{L^2}$ & 42.4 & 57.3 \\
$\mathcal{A}^*$ (ours) & $\sum_{l,l'}f_{cos}(\mathbf{Q}^{[l]}, \mathbf{S}^{[l']})\mathcal{A}^{*}_{l,l'}$ & \ouracc{56.7} & \ouracc{74.1} \\
\hline
\end{tabular}
\end{table}

\section{CLIP-A$^2$M$^2$ network}

\revision{Our M$^2$ module and A$^2$ module are model-agnostic approaches designed for the few-shot action recognition (FSAR) problem, which can be seamlessly integrated into popular architectures without requiring modifications. We build our A$^2$M$^2$-Net upon the CLIP ViT-B network, resulting in the CLIP-A$^2$M$^2$ variant, as illustrated in Figure~\ref{fig:clip_A2M2}.}

\newcommand{\todo}[1]{{#1}}

\begin{figure}[htb!]
  \centering
  \includegraphics[width=\linewidth]{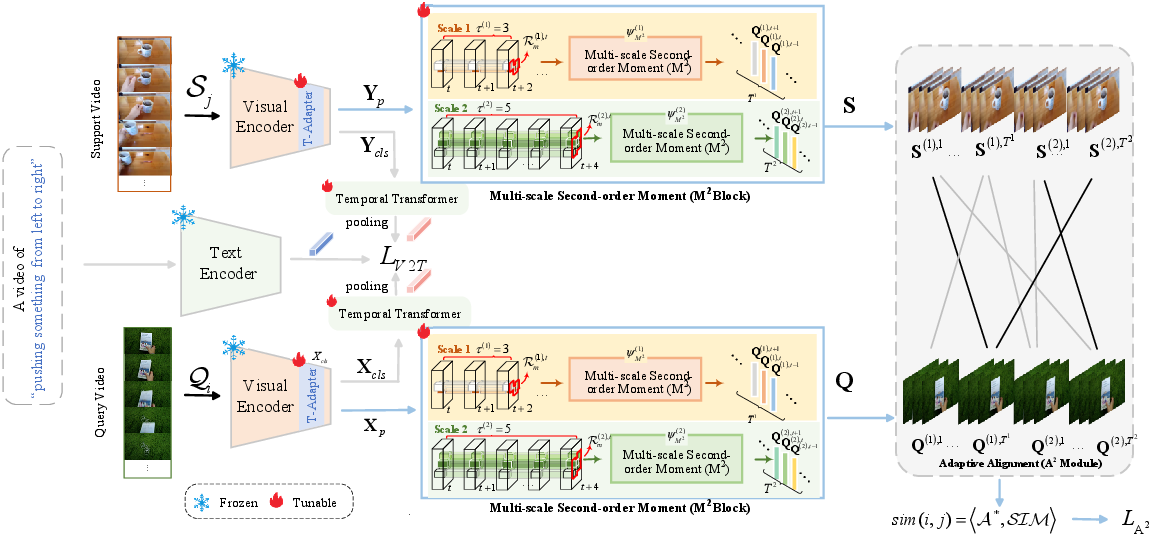}
  \caption{Overview of CLIP-A$^2$M$^2$ using CLIP backbone of ViT-B.}
  \label{fig:clip_A2M2}
\end{figure}

\revision{Specifically, our CLIP-A$^2$M$^2$-Net integrates the M$^2$ block and A$^2$ module sequentially after the CLIP visual encoder while inheriting the contrastive learning mechanism to leverage multi-modality knowledge. First, the video clips of support and query videos ($\mathcal{Q}$ and $\mathcal{S}$) are sent into the visual encoder separately. Then, the generated visual tokens ($\mathbf{X}_p$ and $\mathbf{Y}_p$) and class tokens ($\mathbf{X}_{cls}$ and $\mathbf{Y}_{cls}$) are respectively utilized for visual modality comparison and multi-modality contrastive learning. In the visual branch, the patch tokens are processed by our M$^2$ module, whose outputs are subsequently aligned using the A$^2$ module. The resulting similarity scores from the A$^2$ module are used to compute the visual loss $L_{A^2}$ via a cross-entropy loss. Notably, above operation is consistent with our implementation based on ResNet, with the primary difference being that the M$^2$ module receives transformer-style features (patch tokens) instead of convolution features. For parameter-efficient fine-tuning, the pretrained visual and text encoders are frozen, and we alternately tune the lightweight temporal adapters~\cite{ST_adapter} (T-adapter). These T-adapters are embedded into the last two transformer layers of the visual encoder, positioned just before the multi-head self-attention layers within each transformer block~\cite{ST_adapter}. Each T-adapter consists of a two-layer temporal MLP. In addition, for the multi-modality branch, we incorporate a lightweight temporal transformer following a temporal \todo{average} pooling layer for $\mathbf{X}_{cls}$ and $\mathbf{Y}_{cls}$, aiming to develop a temporal-aware video representation, inspired by~\cite{CLIPFSAR}. The output contrastive loss $L_{V2T}$~\cite{CLIP} is ultimately combined with $L_{A^2}$ for optimization.}

\section{Failure cases analysis}

\mnrevision{In this section, we further expand the failure case illustrations across multiple datasets in Figures~\ref{fig:vis_failure_ssv2} (on the SSv2-Full dataset) and~\ref{fig:vis_failure_others} (on the HMDB-51 and K-100 datasets) to provide a more comprehensive evaluation of our method. Specifically, we label each pair of negative and positive alignment examples as (a) and (b), respectively, preceded by a pair index, e.g., (1a) and (1b). In each pair, the query video $\mathcal{Q}_i$ and its positive support video $\mathcal{S}_{pos}$ share the same ground-truth category, but $\mathcal{Q}_i$ is incorrectly classified into the category of a negative support video $\mathcal{S}_{neg}$. }

\mnrevision{From these figures, we observe that certain misclassifications arise from high appearance similarity between the query and negative support videos (e.g., Figure~\ref{fig:vis_failure_ssv2}(1) and (2)), while in some cases the positive support exhibits markedly different visual characteristics. Additionally, a few failure cases result from ambiguous category instance. For example, in  Figure~\ref{fig:vis_failure_others}(1), the negative support video is labeled as \textit{Kick} rather than \textit{Kick ball}, yet it clearly involves kicking a ball, which causes confusion when classifying the query video labeled \textit{Kick ball}. Furthermore, actions involving multiple objects can increase intra-class diversity, making it challenging to distinguish closely related classes, as seen in the \textit{Tap dancing} vs. \textit{Dancing Charleston} example in Figure~\ref{fig:vis_failure_others}(2).}

\newcommand{\viswidth}[0]{0.9}

\begin{figure}
  \centering

  \captionsetup[subfigure]{labelformat=empty}
  
  \begin{subfigure}[b]{\viswidth\linewidth}
    \centering
    \includegraphics[width=\linewidth]{vis_q_s_neg.png}
    \caption{(1a) $\mathcal{Q}_i:$[\textit{Pouring sth into sth until it overflows}] $\sim \mathcal{S}_{neg}$:[\textit{Pouring sth into sth}]}
    \label{fig:vis_neg}
  \end{subfigure}
  \hspace{0.05\linewidth}
  \begin{subfigure}[b]{\viswidth\linewidth}
    \centering
    \includegraphics[width=\linewidth]{vis_q_s_pos.png}
    \caption{(1b) $\mathcal{Q}_i~\&~\mathcal{S}_{pos}:$[\textit{Pouring sth into sth until it overflows}]}
    \label{fig:vis_pos}
  \end{subfigure}
    \begin{subfigure}[b]{\viswidth\linewidth}
    \centering
    \includegraphics[width=\linewidth]{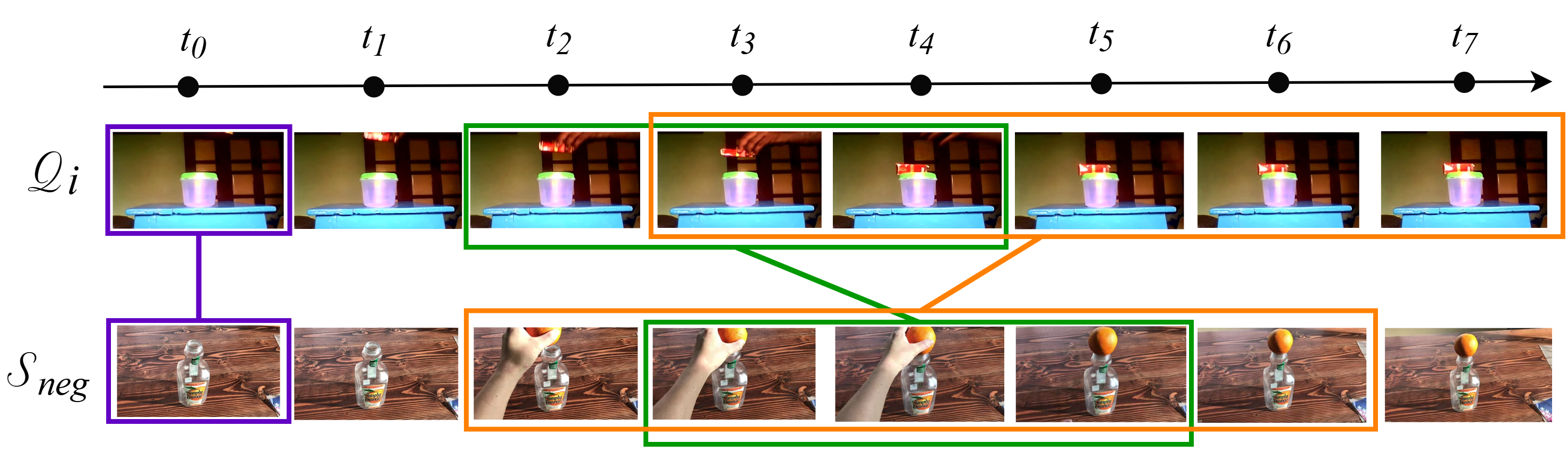}
    \caption{(2a) $\mathcal{Q}_i:$[\textit{Dropping sth onto sth}] $\sim \mathcal{S}_{neg}$:[\textit{Failing to put sth into sth because sth does not fit}]}
    \label{fig:vis_neg_hmdb_kick}
  \end{subfigure}
   \begin{subfigure}[b]{\viswidth\linewidth}
    \centering
    \includegraphics[width=\linewidth]{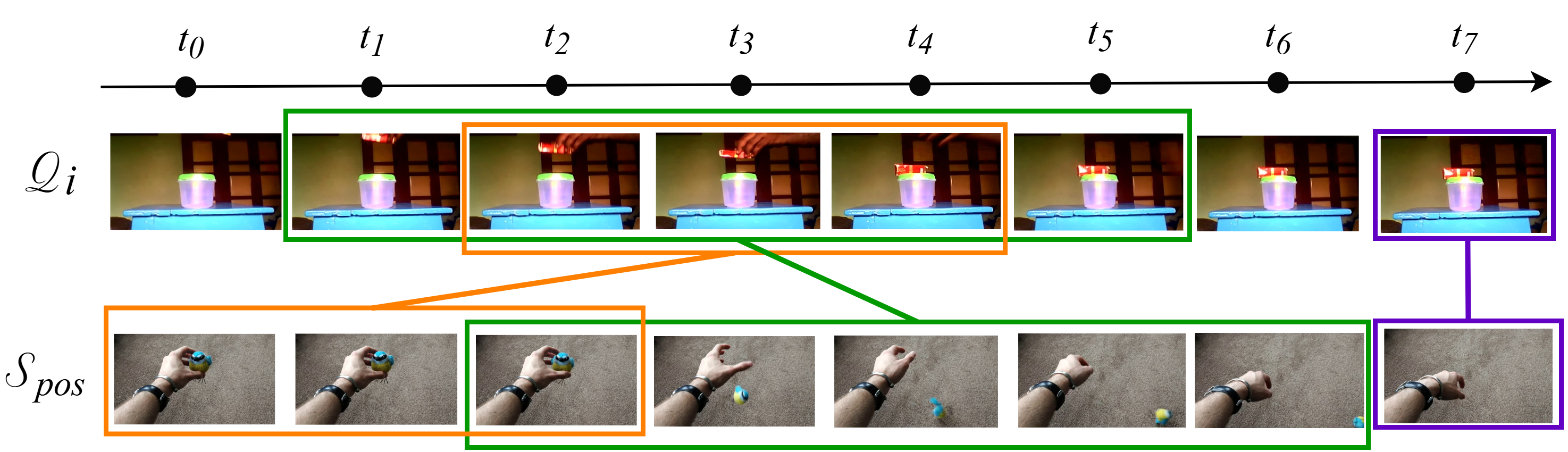}
    \caption{(2b) $\mathcal{Q}_i~\&~\mathcal{S}_{pos}:$[\textit{Dropping sth onto sth}]}
    \label{fig:vis_pos_hmdb_kick}
  \end{subfigure}
  \caption{\mnrevision{Failure examples of A$^2$M$^2$-Net alignment results on SSv2-Full dataset. (1) and (2) each show a pair of failure alignments, with (a) indicating the negative and (b) the positive alignment. $\mathcal{Q}_i$: Query video, $\mathcal{S}_{pos}$: Positive support video, $\mathcal{S}_{neg}$: Negative support video. Each pair displays the top-3 alignments based on matching scores. Boxes of the same color highlight the aligned candidates.}}
  \label{fig:vis_failure_ssv2}
\end{figure}

\begin{figure}
  \centering

    \captionsetup[subfigure]{labelformat=empty}

    \begin{subfigure}[b]{\viswidth\linewidth}
    \centering
    \includegraphics[width=\linewidth]{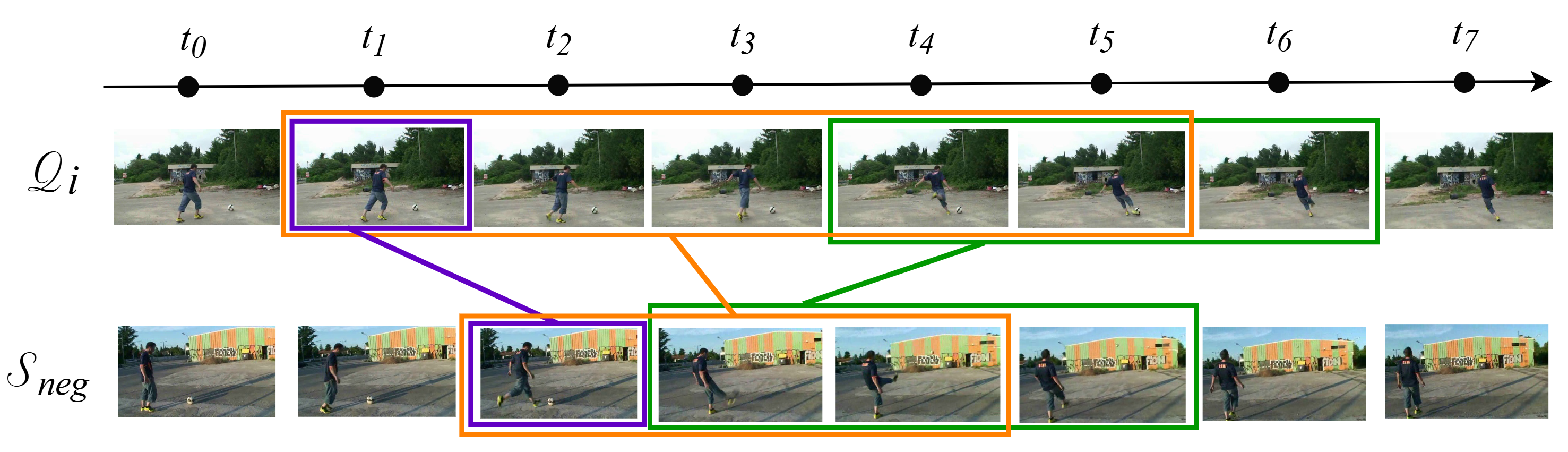}
    \caption{(1a) $\mathcal{Q}_i:$[\textit{Kick ball}] $\sim \mathcal{S}_{neg}$:[\textit{Kick}]}
    \label{fig:vis_neg_hmdb_kick}
  \end{subfigure}
  
  \begin{subfigure}[b]{\viswidth\linewidth}
    \centering
    \includegraphics[width=\linewidth]{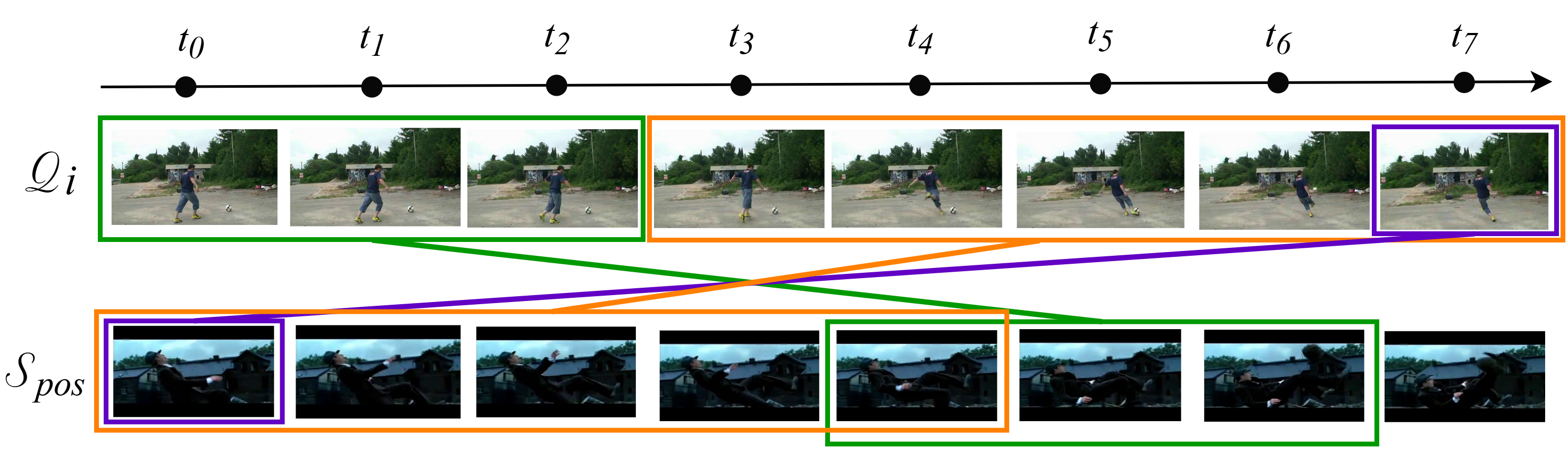}
    \caption{(1b) $\mathcal{Q}_i~\&~\mathcal{S}_{pos}:$[\textit{Kick ball}]}
    \label{fig:vis_pos_hmdb_kick}
  \end{subfigure}

\begin{subfigure}[b]{\viswidth\linewidth}
    \centering
    \includegraphics[width=\linewidth]{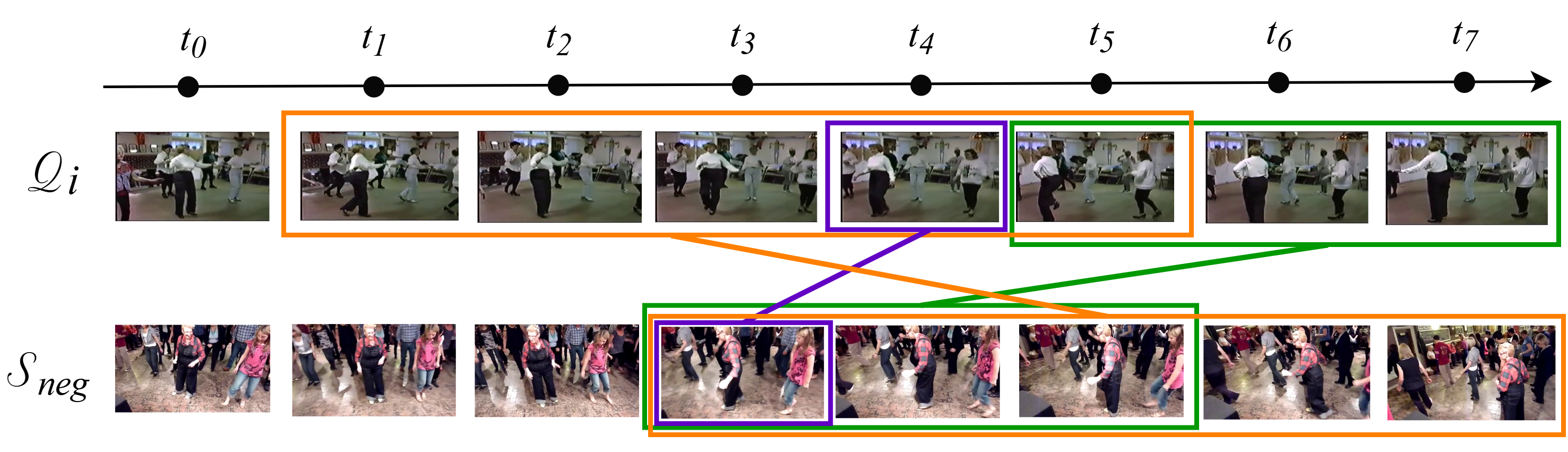}
    \caption{(2a) $\mathcal{Q}_i:$[\textit{Tap dancing}] $\sim \mathcal{S}_{neg}$:[\textit{Dancing Charleston}]}
    \label{fig:vis_neg}
  \end{subfigure}
     \begin{subfigure}[b]{\viswidth\linewidth}
    \centering
    \includegraphics[width=\linewidth]{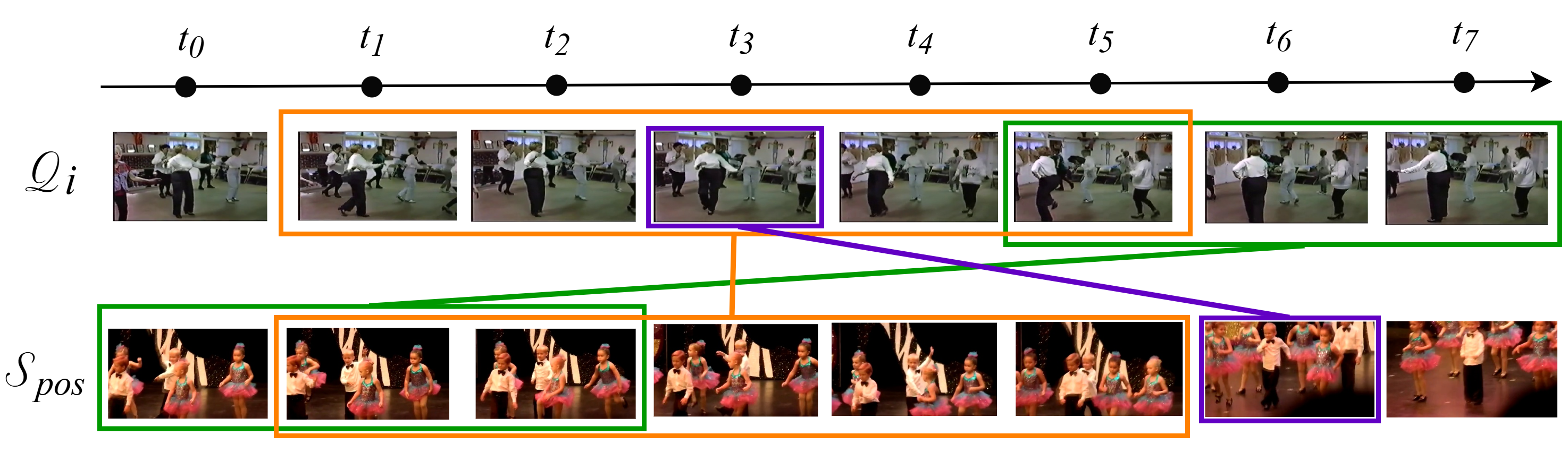}
    \caption{(2b) $\mathcal{Q}_i~\&~\mathcal{S}_{pos}:$[\textit{Tap dancing}]}
    \label{fig:vis_pos_hmdb_kick}
  \end{subfigure}
      \caption{\mnrevision{Failure examples of A$^2$M$^2$-Net alignment results on (1) HMDB-51 and (2) K-100 datasets. (1) and (2) each show a pair of failure alignments, with (a) indicating the negative and (b) the positive alignment. $\mathcal{Q}_i$: Query video, $\mathcal{S}_{pos}$: Positive support video, $\mathcal{S}_{neg}$: Negative support video. Each pair displays the top-3 alignments based on matching scores. Boxes of the same color highlight the aligned candidates.}}

  \label{fig:vis_failure_others}
\end{figure}

\end{appendices}

\defaultbibliography{ref_fsl}  
\putbib
\endgroup

\end{bibunit}

\end{document}